\documentclass{article}

\usepackage{PRIMEarxiv}

\usepackage[utf8]{inputenc} 
\usepackage[T1]{fontenc}    
\usepackage{hyperref}       
\usepackage{url}            
\usepackage{booktabs}       
\usepackage{amsfonts}       
\usepackage{nicefrac}       
\usepackage{microtype}      
\usepackage{lipsum}
\usepackage{fancyhdr}       
\usepackage{graphicx}       
\graphicspath{{media/}}     
\usepackage{amsmath}
\usepackage{multirow}
\usepackage{wrapfig}
\usepackage{placeins}
\usepackage{needspace}
\usepackage[table]{xcolor}
\usepackage{tabularx}
\usepackage{array}
\usepackage{caption}

\usepackage{xcolor}
\usepackage{tcolorbox}
\tcbuselibrary{breakable}
\definecolor{caseBackground}{HTML}{F3F6F8}
\definecolor{caseBorder}{HTML}{D8E1E7}
\newtcolorbox{llmcase}{
  breakable,
  colback=caseBackground,
  colframe=caseBorder,
  boxrule=0.3pt,
  arc=1pt,
  left=7pt, right=7pt, top=6pt, bottom=6pt,
  before skip=8pt, after skip=8pt,
  fontupper=\small,
  before upper={\setlength{\parindent}{0pt}\setlength{\parskip}{4pt}}
}

\usepackage{float}
\usepackage{listings}
\newfloat{listing}{htbp}{lop}
\floatname{listing}{Listing}
\providecolor{codebeige}{RGB}{249,246,237}

\floatstyle{ruled}
\restylefloat{listing}
\floatstyle{plain}

\definecolor{smacBg}{HTML}{F1F1EB}
\definecolor{smacInk}{HTML}{24292F}
\definecolor{smacKeyword}{HTML}{B01459}
\definecolor{smacFunction}{HTML}{6639A6}
\definecolor{smacVariable}{HTML}{0550AE}
\definecolor{smacString}{HTML}{9A4813}
\definecolor{smacComment}{HTML}{426648}

\lstdefinestyle{smacpolicy}{
    language=Python,
    backgroundcolor=\color{smacBg},
    basicstyle=\ttfamily\footnotesize\upshape\color{smacInk},
    keywordstyle=\color{smacKeyword},
    stringstyle=\color{smacString},
    commentstyle=\color{smacComment},
    emph=[1]{act,get,items,min,abs,hypot},
    emphstyle=[1]{\color{smacFunction}},
    emph=[2]{context,allies,stalker,tx,ty,ranged,preferred,close,targets,opts,contact,target,remaining,ds,assigned,md,cd,kite,best,action,dx,dy,px,py,d,score,out},
    emphstyle=[2]{\color{smacVariable}},
    numbers=none,
    frame=none,
    xleftmargin=0pt,
    xrightmargin=0pt,
    aboveskip=0pt,
    belowskip=0pt,
    breaklines=false,
    columns=fullflexible,
    keepspaces=true,
    showstringspaces=false
}

\title{Do We Still Need Reinforcement Learning \\When We Can Read the World Model?
}

\author{
    \textbf{Mingchao Zhang}$^{1,*}$ \
    \textbf{Miduo Cui}$^{1,*}$ \
    \textbf{Kangze He}$^{1,*}$ \
    \textbf{Yuanjun Li}$^{1,*}$ \
    \textbf{Dapeng Li}$^{3,4}$ \
    \textbf{Ce Zhang}$^{2}$ \
    \textbf{Chuanchen Luo}$^{1}$ \\[0.4em]
    \textbf{Xue Yan}$^{5}$ \
    \textbf{Mingzhe Xing}$^{6}$ \
    \textbf{Zhumin Chen}$^{1}$ \
    \textbf{Hangyu Mao}$^{6}$ \
    \textbf{Bin Zhang}$^{2}$ \
    \textbf{Zhiwei Xu}$^{1,\dagger}$ \\[0.8em]
    $^{1}$ Shandong University \quad
    $^{2}$ Institute of Automation, Chinese Academy of Sciences \\
    $^{3}$ Li Auto Inc. \quad
    $^{4}$ Tsinghua University \\
    $^{5}$ Nanyang Technological University \quad
    $^{6}$ Institute of Microelectronics, Chinese Academy of Sciences \\[0.5em]
    $^{*}$ Equal contribution. \quad
    $^{\dagger}$ Corresponding author:
    \texttt{zhiwei\_xu@sdu.edu.cn}
}

\begin{document}
\maketitle

\newcommand{\impr}[2]{%
  \textbf{#1}%
  \hspace{0.15em}%
  \raisebox{0.6ex}{\scriptsize\textbf{$\uparrow$#2}}%
}

\newcommand{\decr}[2]{%
  #1%
  \hspace{0.15em}%
  \raisebox{0.6ex}{\scriptsize$\downarrow$#2}%
}

\vspace{-10pt}
\begin{abstract}
Reinforcement learning has achieved remarkable success in sequential decision-making, but learning effective policies often requires substantial interaction experience.
Model-based reinforcement learning alleviates this dependence by leveraging world models, which are typically used to support planning or policy optimization.
In this paper, we investigate an alternative paradigm in which autonomous LLM-based coding agents directly synthesize control policies by reasoning over readable world models.
By examining environment source code, these agents translate environmental knowledge into executable decision rules and iteratively refine their policies using online feedback or offline data, without task-specific RL training.
Experiments across diverse decision-making benchmarks show that synthesized policies match or outperform conventional RL methods while requiring substantially less experience.
Further analyses reveal interpretable decision rules and how they evolve through feedback-driven refinement.
Comparisons with policies generated without source code access and experiments under challenging conditions further demonstrate the benefits of readable world models.
These findings suggest that direct policy synthesis can reduce reliance on trial and error, challenging the necessity of task-specific RL training.
\end{abstract}



\begin{flushright}
\begin{minipage}{0.85\columnwidth}
\raggedleft
\small
\textit{The great end of life is not knowledge, but action.}\\[0.3em]
--- Thomas Henry Huxley
\end{minipage}
\end{flushright}

\vspace{0.5em}

\section{Introduction}

\begin{wrapfigure}{r}{0.54\textwidth}
\vspace{-.2in}
  \begin{center}
    \includegraphics[width=0.53\textwidth]{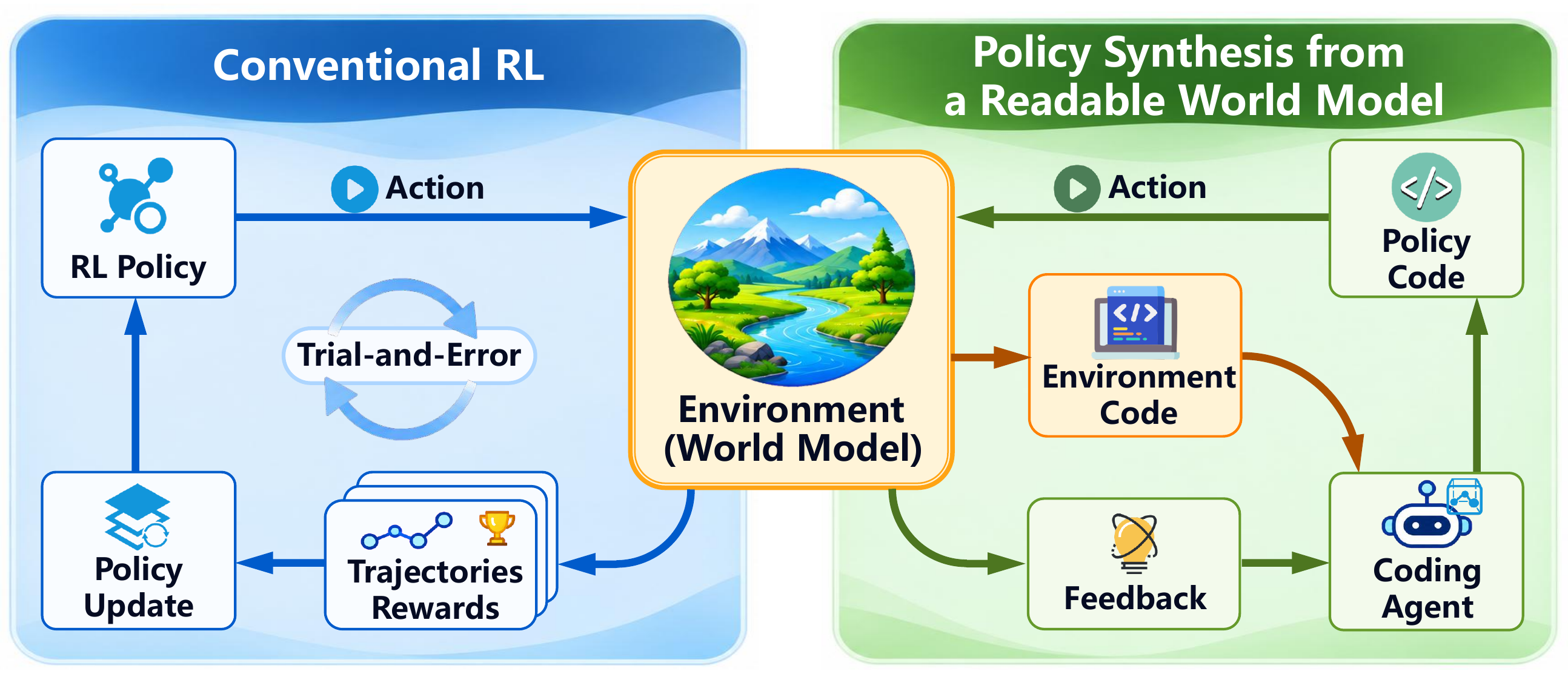}
    \vspace{-.1in}
  \caption{Comparison of conventional RL and policy synthesis from readable world models.}
  \label{fig:intro}
  \end{center}
  \vspace{-.2in}
\end{wrapfigure}
Reinforcement learning (RL)~\cite{sutton1998reinforcement} has emerged as a powerful framework for sequential decision-making.
It has demonstrated remarkable capabilities in single-agent control~\cite{van2016deep} and multi-agent coordination~\cite{sunehag2017value, lowe2017multi}.
Offline RL has further extended its applicability by enabling policy learning from previously collected experience~\cite{levine2020offline}.
Despite these advances, the effectiveness of RL remains heavily dependent on interaction data.
Online RL often requires extensive trial and error to acquire sufficient experience.
In contrast, offline RL avoids additional environment interaction but remains constrained by the quality and coverage of existing data.
To improve data efficiency, model-based RL~\cite{m2023model} leverages \emph{world models}~\cite{ha2018world} to predict how environments respond to actions.
By supporting planning or generating simulated experience, these models can reduce the need for real environment interactions~\cite{sutton1991dyna,feinberg2018model}.
However, prediction-based planning and policy optimization are not the only ways to exploit knowledge of the environment.
When the environment implementation is accessible, its source code provides an explicit description of the underlying dynamics and reward mechanisms.
Such knowledge opens up an alternative possibility: directly constructing executable control policies from a \emph{readable world model}, without task-specific RL training.
This raises a fundamental question: \textbf{Can access to a readable world model make reinforcement learning unnecessary for effective decision-making?}

Answering this question requires bridging the gap between understanding an environment and deriving an effective control policy.
Although a known world model can support planning, it does not directly specify how an agent should act to maximize cumulative rewards.
Traditional approaches typically address this problem through explicit planning or task-specific optimization~\cite{oh2017value, schrittwieser2020mastering}.
Recent advances in large language models (LLMs)~\cite{minaee2024large}, particularly their code understanding~\cite{jiang2026survey} and reasoning capabilities~\cite{huang2023towards}, offer a different possibility.
By examining the environment implementation, LLMs can reason about how actions influence future states.
They can further use this knowledge to identify effective decision rules and translate them into executable control policies.
Unlike conventional RL methods, this process does not necessarily require learning policy parameters through repeated interaction.
Instead, environment interactions can primarily serve to \emph{evaluate} and \emph{refine} the generated policies.
This suggests a different paradigm for sequential decision-making, in which control policies are directly synthesized from world models rather than acquired through task-specific RL training.

To investigate this possibility, we study whether autonomous LLM-based coding agents, such as \emph{Codex} (OpenAI)~\cite{openai2025codex} and \emph{Claude Code} (Anthropic)~\cite{anthropic2025claudecode}, can synthesize effective control policies directly from environment source code.
In this setting, the agent is given access to the environment implementation, which serves as an explicit and readable world model.
By analyzing this implementation, the agent seeks to understand the underlying environment mechanisms and derive suitable decision rules.
These rules are then translated into \emph{executable programs} that directly determine the agent's actions.
The generated policies can be evaluated through environment interaction and iteratively refined based on execution feedback, as shown in Figure~\ref{fig:intro}.
Importantly, this process does not require task-specific RL training, as policies are constructed through code reasoning rather than conventional policy optimization.
We conduct extensive experiments across diverse sequential decision-making benchmarks to examine the effectiveness of this approach.
Our evaluation focuses on both the performance of the synthesized policies and the environment interactions required to obtain them.\looseness=-1

Experimental results demonstrate that policies synthesized from readable world models achieve performance comparable to or even surpassing that of established RL methods across diverse sequential decision-making tasks.
More importantly, they attain such performance with substantially fewer environment interactions than conventional RL training.
In some tasks, effective policies can be synthesized directly from readable world models without any trial and error.
In others, a small number of feedback-driven refinements can further improve policy performance.
Beyond these aggregate results, we further investigate how the synthesized policies make decisions by examining their underlying code.
This analysis provides insights into the structure and interpretability of the generated strategies, revealing how environmental knowledge is translated into concrete decision rules.
To understand the contribution of readable world models, we further compare these strategies with code-based policies generated without access to environment source code.
This comparison allows us to assess how explicit environmental knowledge influences policy performance and interaction efficiency.
These analyses suggest that access to a readable world model can fundamentally change the role of environment interaction.
Rather than learning policies primarily from interaction experience, agents can construct effective strategies by reasoning over explicit environmental knowledge.
Subsequent refinement can be guided by execution feedback in online settings or by existing datasets in offline settings.
Taken together, these experimental results and further analyses suggest that \emph{direct policy synthesis from readable world models} offers a promising alternative to task-specific RL training.


\section{Related Work}

\subsection{World Models for Decision-Making}

World models provide representations of environment dynamics that can support planning and control.
When the environment dynamics are known, optimal decision-making can be formulated through dynamic programming or planning over Markov decision processes~\cite{Brandt2018DynamicP,Puterman1994MarkovDP}.
Early model-based RL methods, such as Dyna~\cite{Sutton1990IntegratedAF}, integrated model learning with planning to improve policy learning from experience.
AlphaZero~\cite{Silver2018AGR} further demonstrated the effectiveness of planning with known game rules, while still relying on reinforcement learning to optimize its policy.
Subsequent research has focused on learning world models when the environment dynamics are unknown.
PILCO~\cite{Deisenroth2011PILCOAM} and PETS~\cite{PETS} employ probabilistic dynamics models to enable sample-efficient control.
PlaNet~\cite{PlaNet} extends model-based planning to visual environments by learning compact latent dynamics.
Similarly, TD-MPC~\cite{TD-MPC} and TD-MPC2~\cite{TD-MPC2} leverage learned latent models for efficient trajectory optimization.
Another line of research exploits world models to generate simulated experience for policy learning.
Early work on recurrent world models~\cite{RWM} demonstrated that policies could be trained within learned simulated environments.
SimPLe~\cite{SimPLe} and the Dreamer family~\cite{Dreamer,DreamerV2,DreamerV3} further advanced this direction by enabling policy learning from imagined trajectories.
MBPO~\cite{MBPO} incorporates short model-generated rollouts into policy optimization, while MOReL~\cite{MOReL}, MOPO~\cite{MOPO}, and COMBO~\cite{COMBO} extend model-based learning to offline settings.
Meanwhile, MuZero~\cite{MuZero} and EfficientZero~\cite{EfficientZero} combine learned models with search to improve decision-making.

Beyond neural representations, recent research has explored explicit and executable world models.
WorldCoder~\cite{WorldCoder}, for example, employs language models to construct programmatic world models from interaction experience and subsequently performs planning over them.
These advances highlight the potential of representing environmental knowledge in a form that supports explicit reasoning.
However, when the environment implementation itself is accessible, the underlying dynamics need not be inferred solely from interaction data.
Instead of using a world model primarily for simulation or policy optimization, an agent may directly reason over its source code to construct executable control policies.
This perspective motivates our investigation into whether readable world models can support effective decision-making without task-specific RL training.

\subsection{LLMs for Decision-Making}

Large language models (LLMs) have increasingly been explored as a means of solving sequential decision-making problems through reasoning and planning.
Early studies demonstrated that pretrained language models could generate actionable plans from task descriptions~\cite{zeroshotplanner} or select feasible actions by leveraging pretrained skills~\cite{doasican}.
Subsequent approaches, such as ReAct~\cite{React} and AdaPlanner~\cite{adaplanner}, incorporated environmental feedback to support adaptive decision-making.
Reflexion~\cite{Reflexion} further demonstrated that language agents could improve their behavior through verbal feedback without updating model parameters.
Other studies combined LLMs with symbolic planning, either by translating task descriptions into formal planning problems~\cite{LLMP} or by constructing explicit domain models~\cite{GuanVSK23}.
Together, these advances establish the potential of LLMs to support decision-making without task-specific policy training.

Beyond generating action sequences, recent research has explored the use of LLMs to synthesize executable decision programs.
Code as Policies~\cite{Liang2022CodeAP} and ProgPrompt~\cite{ProgPrompt} demonstrated how generated programs could represent robot control policies and task plans.
VoxPoser~\cite{VoxPoser} further leveraged code generation to construct spatial value maps for model-based control.
Similarly, Voyager~\cite{Voyager} developed reusable executable skills through iterative interaction with the environment.
In symbolic planning, LLM-generated programs have also been shown to generalize across different task instances~\cite{Silver2023GeneralizedPI}.
These studies suggest that executable code can provide a reusable representation of decision-making strategies.\looseness=-1

Another line of research has employed LLMs to automate components of conventional RL policy optimization.
Language to Rewards~\cite{LanguagetoRewards} and Text2Reward~\cite{Text2Reward} used LLMs to generate reward functions that guide subsequent control optimization.
Eureka~\cite{Eureka} and DrEureka~\cite{DrEureka} further explored automated reward design, while retaining reinforcement learning as the primary mechanism for acquiring control policies.
More closely related to direct policy synthesis, Code to Control~\cite{CodetoControl} generated executable controllers whose parameters were optimized using environment feedback.
Recent work on generalized task and motion planning~\cite{CodingAgents} also demonstrated that coding agents could synthesize reusable decision programs and investigated the benefits of environment source code access.

Building on these advances, our work investigates whether readable world models can substantially reduce the need for task-specific RL training across diverse sequential decision-making benchmarks.
Rather than focusing on a particular policy synthesis algorithm, we systematically examine the performance and interaction efficiency of generated control programs.
We further analyze their underlying decision logic and investigate how access to environment source code contributes to effective policy synthesis.


\section{Policy Synthesis from Readable World Models}

This section introduces the setting for synthesizing control policies from readable world models.
The sequential decision-making problem is first formulated, with environment source code serving as an explicit representation of the world model.
The policy development process is then described, focusing on how autonomous coding agents construct and iteratively refine executable policies.
Finally, we present the evaluation protocol for assessing policy performance and interaction efficiency.
Appendices~\ref{app:ab-environment-access} and~\ref{app:online-prompts} provide details on environment access and online task prompts, respectively.\looseness=-1

\subsection{Problem Formulation}

We consider a sequential decision-making environment $\mathcal{E}$ involving $N$ agents, where $N=1$ corresponds to the single-agent setting.
At each time step $t$, the environment has an underlying state $s_t$, from which each agent $i$ receives an observation $o_t^i$.
In partially observable settings, the observation $o_t^i$ may not fully reveal the underlying state $s_t$.
Each agent selects an action $a_t^i$ based on its interaction history $h_t^i$.
The environment then transitions to the next state $s_{t+1}$ and provides a reward $r_t$.
The objective is to find a joint policy $\boldsymbol{\pi}=\{\pi_i\}_{i=1}^{N}$ that maximizes the expected cumulative reward, defined as $J_{\mathcal{E}}(\boldsymbol{\pi})=\mathbb{E}[\sum_{t=0}^{T-1}\gamma^t r_t]$, where $\gamma$ denotes the discount factor and $T$ is the episode horizon.

Conventional RL methods acquire policies by optimizing them using interaction experience, either collected online or provided through offline datasets.
In contrast, we consider a setting where environmental mechanisms can be directly accessed and understood through readable representations.
Motivated by this perspective, we formally define the concept of a readable world model as follows.

\noindent\textbf{Definition 1 (Readable World Model).}\quad
A \emph{readable world model} is an accessible and interpretable representation of an environment's underlying mechanisms that can be directly examined and reasoned over by coding agents.
In this work, the environment source code $\mathcal{C}_{\mathcal{E}}$ serves as a readable world model.
It encodes the state transition dynamics $P_{\mathcal{E}}(s_{t+1}\mid s_t,\boldsymbol{a}_t)$ and the reward function $R_{\mathcal{E}}(s_t,\boldsymbol{a}_t,s_{t+1})$, where $\boldsymbol{a}_t=(a_t^1,\ldots,a_t^N)$ denotes the joint action.
The implementation also specifies the observation and termination mechanisms.

Given the readable world model $\mathcal{C}_{\mathcal{E}}$, the goal is to synthesize an executable program $c$ that implements a control policy $\boldsymbol{\pi}_c$.
A pretrained coding agent constructs this program by reasoning over the environment implementation.
The generated policy can be evaluated and iteratively refined through environment interaction.
Unlike conventional RL training, this process does not require task-specific policy parameter optimization.
Our objective is to investigate whether such synthesized policies can achieve competitive performance with substantially fewer environment interactions.

\begin{figure}[t]
    \centering
    \includegraphics[width=\textwidth]{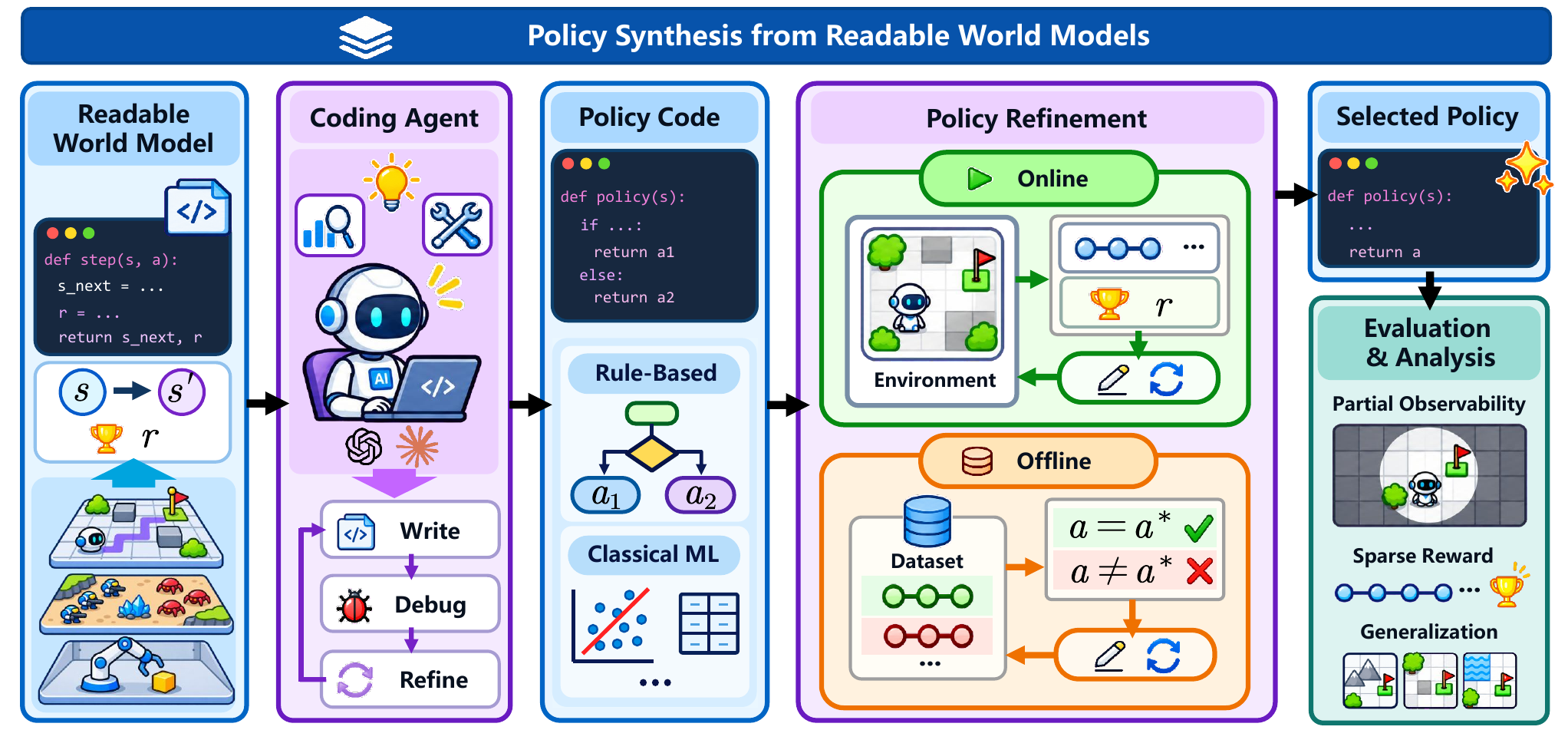}
    \vspace{-0.2 in}
    \caption{Overview of policy synthesis from readable world models. An autonomous coding agent constructs executable policies by reasoning over environment source code. The policies are iteratively refined using online feedback or offline data. The resulting policies are then independently evaluated and further analyzed under various settings.}
    \label{fig:framework}  
\end{figure}

\subsection{Policy Construction and Refinement}

We consider an autonomous LLM-based coding agent, such as Codex or Claude Code, that independently manages the entire policy development process.
Given the environment source code $\mathcal{C}_{\mathcal{E}}$, the agent first explores the implementation to understand the mechanisms governing environment behavior.
In particular, it examines how actions influence future states and how the resulting transitions affect rewards.
Based on this understanding, the agent derives decision rules aimed at maximizing long-term returns.
These rules are then translated into an executable program $c$ that implements the control policy $\boldsymbol{\pi}_c$.
Importantly, the entire process is driven by the coding agent without human intervention, including subsequent policy refinement.

Policy synthesis is \textbf{NOT} necessarily completed in a single attempt.
Starting from an initial program $c^{(0)}$, the agent can iteratively revise its decision logic based on available feedback.
This process produces a sequence of candidate programs $\{c^{(k)}\}_{k=0}^{K}$, where each revision aims to address weaknesses identified in earlier versions.
The candidate with the best assessed performance during development is selected as the final policy $c^*$.
The feedback available for policy refinement depends on whether the agent operates in an online or offline setting:

\vspace{-0.2em}
\textbf{Online.}\quad
In online settings, the agent can execute a candidate policy in the environment to collect interaction trajectories.
By examining the resulting states and rewards, it can identify situations where the current policy makes ineffective decisions.
These observations then guide modifications to the policy code, allowing the agent to progressively improve its behavior.
Each execution contributes to the interaction cost of policy development, which is explicitly accounted for in our evaluation.

\vspace{-0.2em}
\textbf{Offline.}\quad
In offline settings, the agent instead relies on a fixed dataset $\mathcal{D}_{\mathrm{off}}$ containing previously collected interaction trajectories.
The availability of such data also allows the agent to construct policies using statistical machine learning (ML) methods, such as decision trees~\cite{Ke2017LightGBMAH} or regression models~\cite{Tibshirani1996RegressionSA}.
To assess a candidate policy, the agent can compare its predicted actions with those recorded in the dataset.
For example, agreement with actions from high-return trajectories may reveal useful behavioral patterns, while disagreement with actions from low-return trajectories may highlight alternative decisions.
Such comparisons provide useful diagnostic signals, although they do not directly establish the expected return of the generated policy.
Based on these observations, the agent can identify potential weaknesses and refine its decision logic without collecting additional environment interactions.

To track the performance of synthesized policies, we independently evaluate selected intermediate versions $c^{(k)}$ and the final version $c^*$.
These evaluations follow the same protocol as the RL baselines, using held-out random seeds to estimate policy performance.
Unlike the feedback used for policy refinement, the evaluation results are withheld from the coding agent and do not influence subsequent code revisions.
Consequently, the corresponding evaluation interactions are excluded from the policy development interaction budget.
During evaluation, each synthesized program operates through the prescribed observation-action interface without further LLM inference or policy updates.
This separation ensures that evaluation measures policy performance without providing additional information for policy development.\looseness=-1

\subsection{Evaluation Protocol}

We evaluate the effectiveness of synthesized policies from two primary perspectives: \emph{policy performance} and \emph{interaction efficiency}.
For each benchmark, policy performance is measured using its standard evaluation metrics, such as episode return or task success rate.
At selected development checkpoints, the synthesized program $c^{(k)}$ is independently evaluated using held-out random seeds.
The same evaluation protocol is applied to conventional RL baselines to ensure consistent performance comparisons.
These evaluations allow us to track how policy quality evolves throughout the development process.

Interaction efficiency measures the amount of experience required to obtain a policy with a given level of performance.
For online policy synthesis, interaction efficiency is measured by the cumulative number of environment transitions consumed during policy development.
This includes all interactions whose feedback is accessible to the coding agent, regardless of whether they lead to successful policy improvements.
Independent evaluation interactions are excluded, as their results are not available for policy refinement.
For offline policy synthesis, the experience budget is determined by the size of the provided dataset $\mathcal{D}_{\mathrm{off}}$, without additional environment interaction during development.
We compare synthesized policies with RL baselines under matched experience budgets and examine the amount of experience needed to achieve comparable performance.

Beyond aggregate performance, we analyze the decision logic embedded in the generated programs.
By examining representative policies and their revisions, we investigate how environmental knowledge is translated into executable decision rules.
To assess the contribution of readable world models, we further compare policies synthesized with and without access to environment source code.
Both settings employ the same coding agent and comparable development budgets, while differing in the availability of the environment implementation.
This comparison helps isolate the benefits of explicit environmental knowledge from the general policy synthesis capabilities of coding agents.
Finally, we conduct additional analyses under specialized scenarios and experimental settings to gain a broader understanding of the capabilities and limitations of this policy synthesis paradigm.


\section{Policy Performance and Sample Efficiency}
\label{sec:performance}

\subsection{Experimental Setup}
\label{sec:experimental-setup}

\noindent\textbf{Benchmarks.}
We first evaluate the overall effectiveness of policy synthesis from readable world models across diverse sequential decision-making benchmarks.
The online experiments include the \emph{StarCraft Multi-Agent Challenge} (SMAC)~\cite{SMAC} and \emph{Multi-Agent Particle Environments} (MPE)~\cite{lowe2017multi} for multi-agent coordination, along with \emph{MiniGrid}~\cite{MinigridMiniworld23} and \emph{Box2D}~\cite{Parberry2013IntroductionTG} for single-agent decision-making and control.
The offline experiments employ \emph{Datasets for Deep Data-Driven Reinforcement Learning} (D4RL)~\cite{Fu2020D4RLDF} to investigate policy synthesis from fixed experience datasets.
\emph{BabyAI}~\cite{chevalier2018babyai} is further included to examine instruction-following tasks under different demonstration budgets.
Table~\ref{tab:benchmark-summary} summarizes the benchmark configurations and evaluation metrics.


\noindent\textbf{Policy Synthesis Setup.}
All policy synthesis experiments employ Codex powered by \texttt{gpt-6-astra} with \texttt{max} reasoning effort.
For clarity, we refer to \textbf{P}olicy \textbf{S}ynthesis with \textbf{R}eadable \textbf{W}orld \textbf{M}odels as \textbf{PS-RWM} throughout the experiments.
Each experimental configuration is evaluated through three independent runs with different random seeds, covering both online and offline settings.
In each run, the coding agent autonomously examines the environment implementation, constructs executable control policies, and refines them using available development feedback.
During policy construction and refinement, PS-RWM can leverage additional environmental knowledge obtained from the readable world model.
However, during policy execution and independent evaluation, synthesized policies receive the \textbf{SAME} observations as the corresponding RL baselines, without access to privileged environment information.
In multi-agent environments, each agent receives its own observation and selects its action independently.
For SMAC, each agent additionally receives its legal-action mask and previous action, consistent with the corresponding RL baselines.
This ensures that the comparison focuses on differences in policy development rather than differences in the information available during decision-making.

\begin{wraptable}{r}{0.55\linewidth}
\centering
\vspace{-5pt}
\caption{Summary of evaluation benchmarks.}
\label{tab:benchmark-summary}

\footnotesize
\setlength{\tabcolsep}{3.5pt}
\renewcommand{\arraystretch}{1.16}

\resizebox{\linewidth}{!}{%
\begin{tabular}{@{}lllll@{}}
\toprule[0.8pt]
\textbf{Setting} &
\textbf{Benchmark} &
\textbf{Action} &
\textbf{Task} &
\textbf{Metric} \\
\midrule[0.4pt]

\multirow{4}{*}{\textit{Online}}
  & SMAC
  & Discrete
  & Multi-agent combat
  & Win (\%) \\

  & MPE
  & Continuous
  & Multi-agent coordination
  & Return \\

  & MiniGrid
  & Discrete
  & Single-agent navigation
  & Success (\%) \\

  & Box2D
  & Continuous
  & Single-agent control
  & Return \\

\midrule[0.3pt]

\multirow{2}{*}{\textit{Offline}}
  & D4RL
  & Continuous
  & Offline control
  & Return \\

  & BabyAI
  & Discrete
  & Instruction following
  & Success (\%) \\

\bottomrule[0.8pt]
\end{tabular}%
}

\vspace{-5pt}
\end{wraptable}
\noindent\textbf{Baselines.}
For online benchmarks, we compare PS-RWM with representative model-free and model-based RL methods.
The model-free baselines include QMIX~\cite{QMIX} for SMAC, MAPPO~\cite{MAPPO} for MPE, DQN~\cite{dqn} for MiniGrid, and PPO~\cite{PPO} for Box2D.
The model-based baselines include MBVD~\cite{Xu2022MBVD} for SMAC, SeqWM~\cite{Zhao2026SeqWM} for MPE, and DreamerV3~\cite{DreamerV3} for both MiniGrid and Box2D.
QMIX and MBVD have a training budget of approximately 2 million environment steps. The remaining online baselines use a 10-million-step training budget, with comparisons for SeqWM on \texttt{Spread (5 agents)} using the first 8 million steps.
For offline benchmarks, we consider Conservative Q-Learning (CQL)~\cite{CQL}, Implicit Q-Learning (IQL)~\cite{IQL}, Bootstrapping Error Accumulation Reduction (BEAR)~\cite{BEAR}, and TD3+BC~\cite{TDBC} as representative offline RL baselines.
Behavior Cloning (BC) and Decision Transformer (DT)~\cite{DT} are further included to represent imitation learning and sequence modeling approaches, respectively.
Implementation details and evaluation schedules for the core online baselines are provided in Appendix~\ref{app:online-baselines}.

\noindent\textbf{Evaluation Settings.}
Policy selection and performance evaluation follow different protocols depending on the experimental setting.
In online benchmarks, the final PS-RWM policy is selected based on the highest score achieved in a complete 100-episode development validation within the first 1,600 attempted episodes.
This budget covers preliminary tests, failed attempts, and validation episodes.
Offline experiments adopt a different procedure, where policy construction relies on previously collected data \textbf{WITHOUT} additional environment interaction.
On D4RL, three policies are independently constructed for each task and subsequently evaluated over 100 episodes using normalized scores.
Policy synthesis is further evaluated on BabyAI using demonstration budgets of 0, 64, and 256 episodes, without additional environment interaction during development.
The resulting policies are evaluated on 100 held-out random seeds, with success rate as the performance metric.
Throughout offline policy development, independent evaluation feedback remains inaccessible to the coding agent.
All results are summarized using the mean and standard deviation across independent runs where applicable.\looseness=-1

\subsection{Overall Policy Performance}
\label{sec:overall-performance}

\begin{figure}[t]
    \centering
    \includegraphics[width=\linewidth]{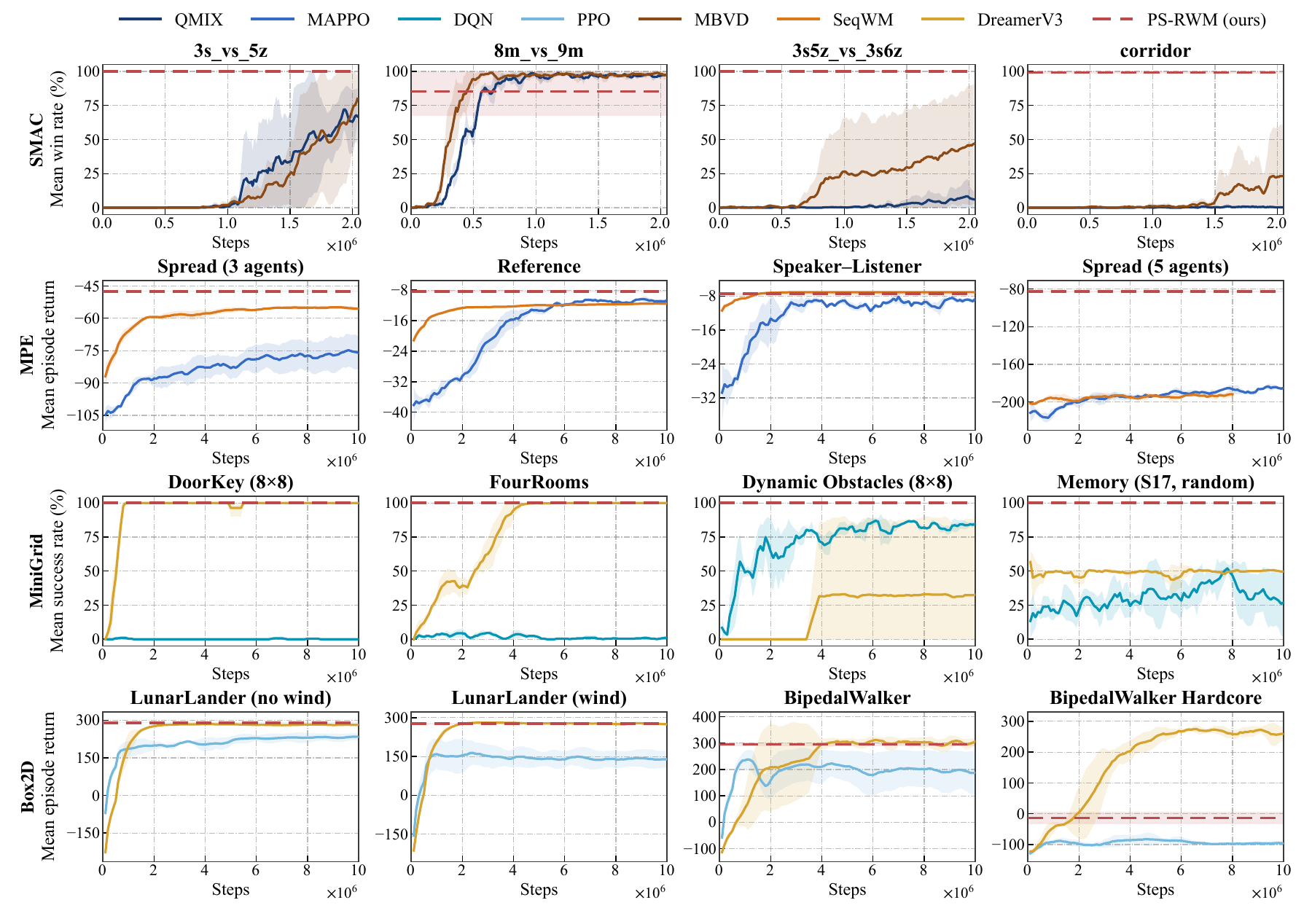}
    \vspace{-20 pt}
    \caption{Performance comparison between PS-RWM and RL baselines across 16 online tasks.
Solid curves show RL performance, with shading indicating standard deviation across three runs.
Box2D curves report training returns.}
    \label{fig:sec4-performance}
\end{figure}

We first investigate whether policies synthesized from readable world models can achieve competitive performance without task-specific RL training.
Figure~\ref{fig:sec4-performance} presents the performance of PS-RWM alongside the learning curves of conventional RL methods.
PS-RWM outperforms the corresponding model-free RL baselines in 15 of the 16 online scenarios.
In SMAC, it achieves mean win rates of at least 99.3\% on three maps, while QMIX and MBVD achieve higher scores on \texttt{8m\_vs\_9m}.
In MPE, PS-RWM outperforms MAPPO across all four tasks and surpasses SeqWM in three, with SeqWM achieving a higher return on \texttt{Speaker--Listener}.
In single-agent benchmarks, PS-RWM achieves 100\% success across all four MiniGrid tasks in each independent run.
It also outperforms PPO across all Box2D tasks and surpasses DreamerV3 on both \texttt{LunarLander} variants.
However, DreamerV3 maintains an advantage on the more challenging \texttt{BipedalWalker} tasks.
Overall, these results demonstrate the strong competitiveness of synthesized policies across diverse online decision-making environments.

The offline experiments further examine whether effective policies can be constructed directly from fixed experience datasets.
As shown in Table~\ref{tab:d4rl-comparison}, PS-RWM achieves the highest average normalized scores on Maze2D, AntMaze, and Kitchen.
The advantages are particularly pronounced in Maze2D and Kitchen, where PS-RWM outperforms all reported baselines across every evaluated task.
On AntMaze, the generated policies also demonstrate competitive performance, although the advantages vary across datasets.
In contrast, PS-RWM remains less competitive on MuJoCo locomotion and Adroit manipulation, particularly in tasks requiring precise continuous control.
These results indicate that direct policy synthesis can achieve strong performance in various offline decision-making problems, although its effectiveness remains task-dependent.

Beyond D4RL, BabyAI extends the offline evaluation to instruction-following tasks under varying demonstration budgets.
As reported in Table~\ref{tab:babyai-demonstration-efficiency}(a), PS-RWM achieves an average success rate of 99.28\%, substantially outperforming all six learning-based baselines.
Remarkably, PS-RWM maintains a 100\% success rate across all six tasks even without access to demonstration data, as shown in Table~\ref{tab:babyai-demonstration-efficiency}(b).
Comparable performance is observed with 64 demonstrations, while the average success rate decreases slightly to 98.83\% with 256 demonstrations.
These findings suggest that readable world models can provide sufficient environmental knowledge to construct highly effective policies even when interaction experience is limited or unavailable during development.

\begin{table*}[t]
\centering
\caption{
Performance comparison on D4RL benchmarks.
Higher normalized scores indicate better performance.
PS-RWM results are reported as mean $\pm$ sample standard deviation over three independent policy-construction runs.
}
\label{tab:d4rl-comparison}
\begingroup
\setlength{\tabcolsep}{2.5pt}
\renewcommand{\arraystretch}{1.10}
\footnotesize
\begin{tabular}{@{}cl*{6}{c}>{\columncolor{black!5}}c@{}}
\toprule
\multicolumn{2}{c}{\textbf{Scenarios}} & \textbf{CQL} & \textbf{IQL} & \textbf{BEAR} & \textbf{TD3+BC} & \textbf{BC} & \textbf{DT} & \textbf{PS-RWM} \\
\midrule

\multirow{4}{*}{\textbf{\shortstack{Maze2D}}} & \texttt{maze2d-umaze}
& $94.7$ & $42.1$ & $65.7$ & $14.8$
& $16.4 \pm 4.7$ & $60.3 \pm 7.7$
& $154.0 \pm 7.6$ \\

& \texttt{maze2d-medium}
& $41.8$ & $34.9$ & $25.0$ & $62.1$
& $20.1 \pm 8.4$ & $37.0 \pm 8.3$
& $157.5 \pm 18.3$ \\

& \texttt{maze2d-large}
& $49.6$ & $61.7$ & $81.0$ & $88.6$
& $10.3 \pm 8.6$ & $25.4 \pm 4.9$
& $219.2 \pm 21.4$ \\

\cmidrule(l){2-9}
& \textbf{Average (3 tasks)}
& $62.0$ & $46.2$ & $57.2$ & $55.2$
& $15.6$ & $40.9$
& \impr{$\bold{176.9}$}{$114.9$} \\
\midrule

\multirow{7}{*}{\textbf{\shortstack{AntMaze}}} & \texttt{antmaze-umaze}
& $74.0$ & $87.5$ & $73.0$ & $78.6$
& $82.3 \pm 3.2$ & $67.0 \pm 1.7$
& $83.0 \pm 15.6$ \\

& \texttt{antmaze-umaze-diverse}
& $84.0$ & $62.2$ & $61.0$ & $71.4$
& $81.7 \pm 9.5$ & $63.0 \pm 4.6$
& $88.7 \pm 9.9$ \\

& \texttt{antmaze-medium-play}
& $61.2$ & $71.2$ & $0.0$ & $10.6$
& $0.0 \pm 0.0$ & $0.0\pm 0.0$
& $77.0 \pm 11.3$ \\

& \texttt{antmaze-medium-diverse}
& $53.7$ & $70.0$ & $8.0$ & $3.0$
& $2.0 \pm 1.0$ & $2.7 \pm 1.2$
& $51.0 \pm 23.3$ \\

& \texttt{antmaze-large-play}
& $15.8$ & $39.6$ & $0.0$ & $0.2$
& $0.0\pm 0.0$ & $0.0 \pm 0.0$
& $54.0 \pm 37.7$ \\

& \texttt{antmaze-large-diverse}
& $14.9$ & $47.5$ & $0.0$ & $0.0$
& $0.0 \pm 0.0$ & $1.0 \pm 1.0$
& $76.3 \pm 27.2$ \\

\cmidrule(l){2-9}
& \textbf{Average (6 tasks)}
& $50.6$ & $63.0$ & $23.7$ & $27.3$
& $27.7$ & $22.3$
& \impr{$\bold{71.7}$}{$8.7$} \\
\midrule

\multirow{3}{*}{\textbf{\shortstack{Kitchen}}} & \texttt{kitchen-complete}
& $43.8$ & $62.5$ & $0.0$ & $0.0$
& $48.3 \pm 3.8$ & $40.3 \pm 5.3$
& $84.25 \pm 8.1$ \\

& \texttt{kitchen-partial}
& $49.8$ & $46.3$ & $13.1$ & $0.0$
& $52.5 \pm 12.5$ & $59.9 \pm 4.2$
& $96.2 \pm 2.8$ \\

\cmidrule(l){2-9}
& \textbf{Average (2 tasks)}
& $46.8$ & $54.4$ & $6.6$ & $0.0$
& $50.4$ & $50.1$
& \impr{$\bold{90.2}$}{$35.8$} \\
\midrule

\multirow{10}{*}{\textbf{\shortstack{MuJoCo\\Locomotion}}} & \texttt{halfcheetah-medium}
& $49.2$ & $47.4$ & $41.7$ & $48.4$
& $40.5 \pm 0.1$ & $40.0 \pm 0.1$
& $42.4 \pm 0.3$ \\

& \texttt{halfcheetah-medium-replay}
& $45.5$ & $44.2$ & $38.6$ & $44.6$
& $35.8 \pm 0.7$ & $35.0 \pm 1.0$
& $36.1 \pm 4.0$ \\

& \texttt{halfcheetah-medium-expert}
& $91.6$ & $86.7$ & $53.4$ & $90.7$
& $86.2 \pm 9.4$ & $91.7 \pm 0.3$
& $80.3 \pm 6.2$ \\

& \texttt{hopper-medium}
& $69.4$ & $66.3$ & $52.1$ & $59.3$
& $59.9 \pm 1.5$ & $63.6 \pm 2.6$
& $54.3 \pm 1.9$ \\

& \texttt{hopper-medium-replay}
& $95.0$ & $94.7$ & $33.7$ & $60.9$
& $48.0 \pm 28.2$ & $78.7 \pm 0.3$
& $25.5 \pm 17.8$ \\

& \texttt{hopper-medium-expert}
& $105.4$ & $91.5$ & $96.3$ & $98.0$
& $67.5 \pm 13.1$ & $109.8 \pm 0.5$
& $105.6 \pm 2.4$ \\

& \texttt{walker2d-medium}
& $83.0$ & $78.3$ & $59.1$ & $83.7$
& $78.8 \pm 1.1$ & $78.1 \pm 1.5$
& $70.5 \pm 1.9$ \\

& \texttt{walker2d-medium-replay}
& $77.2$ & $73.9$ & $19.2$ & $81.8$
& $57.5 \pm 3.3$ & $71.5 \pm 1.7$
& $17.4 \pm 17.6$ \\

& \texttt{walker2d-medium-expert}
& $108.8$ & $109.6$ & $40.1$ & $110.1$
& $108.7 \pm 0.3$ & $108.9 \pm 0.1$
& $107.1 \pm 0.5$ \\

\cmidrule(l){2-9}
& \textbf{Average (9 tasks)}
& $80.6$ & $77.0$ & $48.2$ & $75.3$
& $64.8$ & $75.3$
& \decr{$59.9$}{$20.7$} \\
\midrule

\multirow{7}{*}{\textbf{\shortstack{Adroit}}} & \texttt{pen-human}
& $37.5$ & $71.5$ & $-1.0$ & $-3.9$
& $109.9 \pm 7.6$ & $83.6 \pm 7.2$
& $70.5 \pm 9.0$ \\

& \texttt{hammer-human}
& $4.4$ & $1.4$ & $0.3$ & $1.0$
& $2.5 \pm 0.8$ & $3.0 \pm 0.8$
& $0.9 \pm 0.4$ \\

& \texttt{door-human}
& $9.9$ & $4.3$ & $-0.3$ & $-0.3$
& $14.5 \pm 1.9$ & $22.7 \pm 4.4$
& $0.8 \pm 0.7$ \\

& \texttt{pen-cloned}
& $39.2$ & $37.3$ & $26.5$ & $5.1$
& $77.8 \pm 12.3$ & $60.3 \pm 8.7$
& $30.0 \pm 3.3$ \\

& \texttt{hammer-cloned}
& $2.1$ & $2.1$ & $0.3$ & $0.3$
& $2.6 \pm 2.4$ & $1.6 \pm 0.4$
& $1.1 \pm 0.6$ \\

& \texttt{door-cloned}
& $0.4$ & $1.6$ & $-0.1$ & $-0.3$
& $1.6 \pm 1.0$ & $14.3 \pm 1.7$
& $0.9 \pm 1.0$ \\

\cmidrule(l){2-9}
& \textbf{Average (6 tasks)}
& $15.6$ & $19.7$ & $4.3$ & $0.3$
& $34.8$ & $30.9$
& \decr{$17.4$}{$17.4$} \\

\bottomrule
\end{tabular}
\vspace{-15 pt}
\endgroup
\end{table*}


\begin{table*}[t]
\centering
\caption{
Performance comparison and demonstration efficiency on BabyAI.
Success rates are reported in percentages.
$M$ denotes the number of complete demonstration episodes available to PS-RWM during policy construction.
}
\label{tab:babyai-demonstration-efficiency}
\begingroup
\setlength{\tabcolsep}{3pt}
\renewcommand{\arraystretch}{1.10}
\footnotesize
\begin{tabular}{@{}l*{6}{c}>{\columncolor{black!5}}c@{\hspace{18pt}}*{3}{c}@{}}
\toprule
\multicolumn{8}{c@{\hspace{12pt}}}{\textbf{(a) Full Dataset}}
& \multicolumn{3}{c}{\textbf{(b) PS-RWM Data Efficiency}} \\
\cmidrule(lr){1-8}
\cmidrule(l){9-11}

\textbf{Task}
& \textbf{IQL}
& \textbf{CQL}
& \textbf{BEAR}
& \textbf{TD3+BC}
& \textbf{BC}
& \textbf{DT}
& \textbf{PS-RWM}
& $\boldsymbol{M=0}$
& $\boldsymbol{M=64}$
& $\boldsymbol{M=256}$ \\

\midrule

\texttt{GoToRedBall}
& $93.00$ & $96.33$ & $90.33$ & $91.67$
& $94.33$ & $98.67$ & $\mathbf{100.00}$
& $100.00$ & $100.00$ & $100.00$ \\

\texttt{GoToLocal}
& $74.00$ & $76.33$ & $73.00$ & $75.67$
& $73.67$ & $80.33$ & $\mathbf{100.00}$
& $100.00$ & $100.00$ & $100.00$ \\

\texttt{GoToSeq}
& $53.33$ & $55.00$ & $56.67$ & $41.00$
& $52.00$ & $70.33$ & $\mathbf{100.00}$
& $100.00$ & $100.00$ & $95.00$ \\

\texttt{PutNextLocal}
& $9.33$ & $19.00$ & $14.33$ & $13.67$
& $11.67$ & $12.67$ & $\mathbf{100.00}$
& $100.00$ & $100.00$ & $100.00$ \\

\texttt{UnlockLocalDist}
& $58.00$ & $50.67$ & $35.00$ & $38.33$
& $53.67$ & $84.67$ & $\mathbf{100.00}$
& $100.00$ & $100.00$ & $100.00$ \\

\texttt{BossLevel}
& $32.33$ & $23.33$ & $23.33$ & $30.67$
& $30.67$ & $45.33$ & $\mathbf{95.67}$
& $100.00$ & $100.00$ & $98.00$ \\

\midrule

\textbf{Average}
& $53.33$ & $53.44$ & $48.78$ & $48.50$
& $52.67$ & $65.33$
& \impr{$\mathbf{99.28}$}{$33.95$}
& $100.00$ & $100.00$ & $98.83$ \\

\bottomrule
\end{tabular}
\endgroup
\end{table*}

\subsection{Sample Efficiency}
\label{sec:sample-efficiency}

We further investigate how much experience is required to construct effective control policies in both online and offline settings.
In online settings, a performance target is defined for each task to enable a consistent comparison of the environment steps required by different methods.
Specifically, the targets are set to an 80\% win rate for SMAC and an 80\% success rate for MiniGrid.
For MPE and Box2D, the target is the lowest final-test mean return among the three model-free RL runs.
The interaction cost of PS-RWM includes all development steps up to the first 100-episode validation that reaches the target, including unsuccessful attempts and earlier policy versions.
For online RL baselines, the required steps are determined by the earliest periodic or final checkpoint evaluation that reaches the target.
Since Box2D reports training returns, smoothed training returns are used instead to determine target attainment.
RL evaluation interactions and model-generated rollouts are accounted for separately.
In offline settings, data efficiency is assessed by varying the amount of previously collected experience available for policy construction.
The performance of the resulting policies is then evaluated under different data budgets, without collecting additional environment interactions during development.

\vspace{-0.2em}
\textbf{Online.}\quad
Table~\ref{tab:sec4-steps-to-target} summarizes the online sample efficiency results under the predefined performance targets.
PS-RWM successfully reaches the target in all three independent runs on 15 of the 16 scenarios.
The advantage is particularly pronounced in MiniGrid, where PS-RWM requires only 1,789 environment steps on \texttt{DoorKey}, compared with 400,000 steps for DreamerV3.
This represents a reduction of more than 200 times in environment interactions.
Similar improvements are observed in other tasks, although their magnitude varies.
For instance, PS-RWM reaches the target on \texttt{BipedalWalker} after 529,773 steps, compared with 633,333 steps for PPO.
Overall, these results suggest that reasoning over readable world models can substantially reduce the experience required to obtain competitive policies.

\vspace{-0.2em}
\textbf{Offline.}\quad
The offline analysis further examines the data efficiency of PS-RWM using four nested subsets of \texttt{halfcheetah-medium-v2}, ranging from 1\% to 100\% of the original dataset.
As shown in Table~\ref{tab:halfcheetah-data-efficiency}, PS-RWM can construct an effective controller using only 1\% of the data, corresponding to 10 trajectories and 10,000 transitions.
With this limited experience, the coding agent constructs a policy based on Mat\'ern kernel ridge regression to predict control actions from observed states.
The resulting policy achieves a D4RL normalized score of 41.03, comparable to the average score of 40.886 obtained by the demonstration trajectories.
Notably, this policy is constructed without additional environment interaction or task-specific RL training.
These results suggest that PS-RWM can extract useful control strategies from limited offline experience by combining readable world models with conventional statistical learning methods.

The above results demonstrate that readable world models can substantially reduce the experience required for effective policy development.
PS-RWM achieves competitive performance with fewer environment interactions in online settings and constructs effective policies from limited data in offline settings.
To better understand these findings, we next examine the decision-making mechanisms and evolution of the synthesized policies.

\begin{table*}[t]
\centering
\caption{
Environment steps required to reach predefined performance targets.
Results indicate the mean steps to first attainment among successful runs.
Parentheses report the number of successful runs out of three.
Bold indicates the fewest steps, while a dash indicates that the target was never reached.
}
\label{tab:sec4-steps-to-target}
\begingroup
\setlength{\tabcolsep}{5pt}
\renewcommand{\arraystretch}{1.10}
\footnotesize
\begin{tabular}{@{}llrrr>{\columncolor{black!5}}r@{}}
\toprule
\textbf{Benchmark}
& \textbf{Scenario}
& \textbf{Target}
& \textbf{Model-Free RL}
& \textbf{Model-Based RL}
& \textbf{PS-RWM} \\
\midrule

\multirow{5}{*}{\textbf{SMAC}}
& &
& \multicolumn{1}{c}{{QMIX}}
& \multicolumn{1}{c}{{MBVD}}
& \\
\cmidrule(lr){4-4}\cmidrule(lr){5-5}

& \texttt{3s\_vs\_5z}
& $80$
& $1{,}617{,}569$ $(3/3)$
& $1{,}821{,}847$ $(3/3)$
& $\mathbf{19{,}409}$ $(3/3)$ \\

& \texttt{8m\_vs\_9m}
& $80$
& $490{,}910$ $(3/3)$
& $320{,}452$ $(3/3)$
& $\mathbf{27{,}413}$ $(2/3)$ \\

& \texttt{3s5z\_vs\_3s6z}
& $80$
& \textemdash{} $(0/3)$
& $972{,}874$ $(1/3)$
& $\mathbf{70{,}711}$ $(3/3)$ \\

& \texttt{corridor}
& $80$
& \textemdash{} $(0/3)$
& $2{,}002{,}383$ $(1/3)$
& $\mathbf{47{,}744}$ $(3/3)$ \\

\midrule

\multirow{5}{*}{\textbf{MPE}}
& &
& \multicolumn{1}{c}{{MAPPO}}
& \multicolumn{1}{c}{{SeqWM}}
& \\
\cmidrule(lr){4-4}\cmidrule(lr){5-5}

& \texttt{Spread (3 agents)}
& $-81.78$
& $2{,}900{,}000$ $(3/3)$
& $166{,}667$ $(3/3)$
& $\mathbf{2{,}500}$ $(3/3)$ \\

& \texttt{Reference}
& $-10.34$
& $6{,}066{,}667$ $(3/3)$
& $10{,}000{,}000$ $(3/3)$
& $\mathbf{2{,}583}$ $(3/3)$ \\

& \texttt{Speaker--Listener}
& $-8.53$
& $2{,}533{,}333$ $(3/3)$
& $533{,}333$ $(3/3)$
& $\mathbf{9{,}667}$ $(3/3)$ \\

& \texttt{Spread (5 agents)}
& $-188.63$
& $3{,}066{,}667$ $(3/3)$
& $2{,}433{,}333$ $(3/3)$
& $\mathbf{3{,}158}$ $(3/3)$ \\

\midrule

\multirow{5}{*}{\textbf{MiniGrid}}
& &
& \multicolumn{1}{c}{{DQN}}
& \multicolumn{1}{c}{{DreamerV3}}
& \\
\cmidrule(lr){4-4}\cmidrule(lr){5-5}

& \texttt{DoorKey ($8\times8$)}
& $80$
& \textemdash{} $(0/3)$
& $400{,}000$ $(3/3)$
& $\mathbf{1{,}789}$ $(3/3)$ \\

& \texttt{FourRooms}
& $80$
& \textemdash{} $(0/3)$
& $3{,}100{,}000$ $(3/3)$
& $\mathbf{3{,}651}$ $(3/3)$ \\

& \texttt{Dynamic Obstacles ($8\times8$)}
& $80$
& $533{,}333$ $(3/3)$
& $3{,}500{,}000$ $(1/3)$
& $\mathbf{1{,}643}$ $(3/3)$ \\

& \texttt{Memory (S17, random)}
& $80$
& \textemdash{} $(0/3)$
& \textemdash{} $(0/3)$
& $\mathbf{1{,}221}$ $(3/3)$ \\

\midrule

\multirow{5}{*}{\textbf{Box2D}}
& &
& \multicolumn{1}{c}{{PPO}}
& \multicolumn{1}{c}{{DreamerV3}}
& \\
\cmidrule(lr){4-4}\cmidrule(lr){5-5}

& \texttt{LunarLander (no wind)}
& $210.72$
& $2{,}133{,}333$ $(3/3)$
& $1{,}100{,}000$ $(3/3)$
& $\mathbf{20{,}316}$ $(3/3)$ \\

& \texttt{LunarLander (wind)}
& $120.27$
& $2{,}566{,}667$ $(3/3)$
& $700{,}000$ $(3/3)$
& $\mathbf{50{,}630}$ $(3/3)$ \\

& \texttt{BipedalWalker}
& $170.72$
& $633{,}333$ $(3/3)$
& $2{,}166{,}667$ $(3/3)$
& $\mathbf{529{,}773}$ $(3/3)$ \\

& \texttt{BipedalWalker Hardcore}
& $-99.03$
& $666{,}667$ $(3/3)$
& $566{,}667$ $(3/3)$
& $\mathbf{59{,}193}$ $(3/3)$ \\

\bottomrule
\end{tabular}
\endgroup
\end{table*}

\begin{table*}[t]
\centering
\begingroup
\setlength{\tabcolsep}{4pt}
\renewcommand{\arraystretch}{1.12}
\footnotesize

\begin{minipage}[t]{0.4\textwidth}
\vspace{0pt}
\centering

\captionsetup{
    justification=raggedright,
    singlelinecheck=false,
    margin={23pt,0pt},
}
\caption{
Offline data efficiency on \texttt{halfcheetah-medium-v2}.
}
\label{tab:halfcheetah-data-efficiency}

\begin{tabular}{@{}lrrrr@{}}
\toprule
& \multicolumn{4}{c}{\textbf{Offline Dataset Fraction}} \\
\cmidrule(lr){2-5}

\textbf{Method}
& \textbf{1\%}
& \textbf{10\%}
& \textbf{30\%}
& \textbf{100\%} \\
\midrule

BC
& $25.5$ & $42.1$ & $42.5$ & $42.5$ \\

TD3+BC
& $24.9$ & $47.9$ & $48.3$ & $48.4$ \\

IQL
& $25.3$ & $46.1$ & $47.0$ & $47.4$ \\

CQL
& $40.9$ & $46.7$ & $48.8$ & $49.1$ \\

BEAR
& $18.7$ & $42.6$ & $42.9$ & $42.8$ \\

DT
& $37.1$ & $41.6$ & $42.1$ & $42.3$ \\

\midrule
\rowcolor{black!5}
\textbf{PS-RWM}
& $41.0$ & $42.5$ & $41.9$ & $42.7$ \\

\bottomrule
\end{tabular}

\par\vspace{4pt}
\raggedright

\end{minipage}
\hfill
\begin{minipage}[t]{0.53\textwidth}
\vspace{0pt}
\centering

\includegraphics[width=\linewidth]{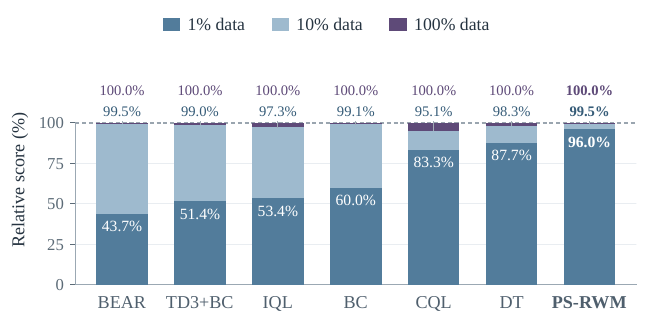}

\captionof{figure}{
Performance trends under different offline dataset fractions on \texttt{halfcheetah-medium-v2}.
}
\label{fig:halfcheetah-data-efficiency}
\end{minipage}

\endgroup
\end{table*}

\section{Understanding Synthesized Policies}
\label{sec:Understanding Synthesized Policies}

\subsection{Decision-Making Mechanisms}
\label{sec:Decision-Making Mechanisms}

While the preceding results demonstrate the effectiveness of PS-RWM, aggregate performance metrics provide limited insight into how the synthesized policies make decisions.
Understanding their decision-making behavior is essential for revealing the strategies underlying their performance.
To this end, we examine representative policy behaviors in both online and offline settings.
The analysis focuses on coordination among agents in multi-agent tasks and the use of environment dynamics in continuous control.

\begin{wraptable}{r}{0.33\linewidth}
\centering
\vspace{-8pt}
\captionsetup{
    font=footnotesize,
    justification=raggedright,
    singlelinecheck=false,
    skip=5pt
}
\caption{
Allied death rates and zero-death episode rates on SMAC (\%).
}
\label{tab:smac-unit-survival}

\footnotesize
\begingroup
\setlength{\tabcolsep}{3pt}
\renewcommand{\arraystretch}{1.12}

\begin{tabular*}{\linewidth}{
@{}l@{\extracolsep{\fill}}r>{\columncolor{black!5}}r@{}
}
\toprule
\textbf{Scenario}
& \textbf{QMIX}
& \textbf{PS-RWM} \\
\midrule

\multicolumn{3}{@{}l}{
\textbf{Allied Death Rate ($\downarrow$)}
} \\

\texttt{3s\_vs\_5z}
& $51.78$ & $\mathbf{8.56}$ \\

\texttt{8m\_vs\_9m}
& $86.29$ & $\mathbf{53.83}$ \\

\texttt{3s5z\_vs\_3s6z}
& $99.79$ & $\mathbf{39.08}$ \\

\texttt{corridor}
& $99.94$ & $\mathbf{37.72}$ \\

\cmidrule(l){1-3}
\textbf{Average}
& $84.45$ & $\mathbf{34.80}$ \\

\midrule

\multicolumn{3}{@{}l}{
\textbf{Zero-Death Episode Rate ($\uparrow$)}
} \\

\texttt{3s\_vs\_5z}
& $13.67$ & $\mathbf{74.67}$ \\

\texttt{8m\_vs\_9m}
& $0.00$ & $\mathbf{1.67}$ \\

\texttt{3s5z\_vs\_3s6z}
& $0.00$ & $\mathbf{0.67}$ \\

\texttt{corridor}
& $0.00$ & $\mathbf{3.33}$ \\

\cmidrule(l){1-3}
\textbf{Average}
& $3.42$ & $\mathbf{20.08}$ \\

\bottomrule
\end{tabular*}
\endgroup
\vspace{-8pt}
\end{wraptable}
\vspace{-0.2em}
\textbf{Online.}\quad
To examine multi-agent decision-making behavior, we compare PS-RWM with QMIX across four SMAC scenarios.
Table~\ref{tab:smac-unit-survival} reports the allied death rate and the proportion of episodes without allied casualties.
Across the four scenarios, PS-RWM achieves a substantially lower average death rate than QMIX (34.80\% vs. 84.45\%).
It also achieves higher zero-death episode rates in every scenario.
The largest difference occurs on \texttt{3s\_vs\_5z}, where 74.67\% of PS-RWM episodes end without allied casualties, compared with only 13.67\% for QMIX.
These results indicate that synthesized policies are more effective at preserving allied units during combat.

\begin{figure}[t]
    \centering
    \includegraphics[width=\linewidth]{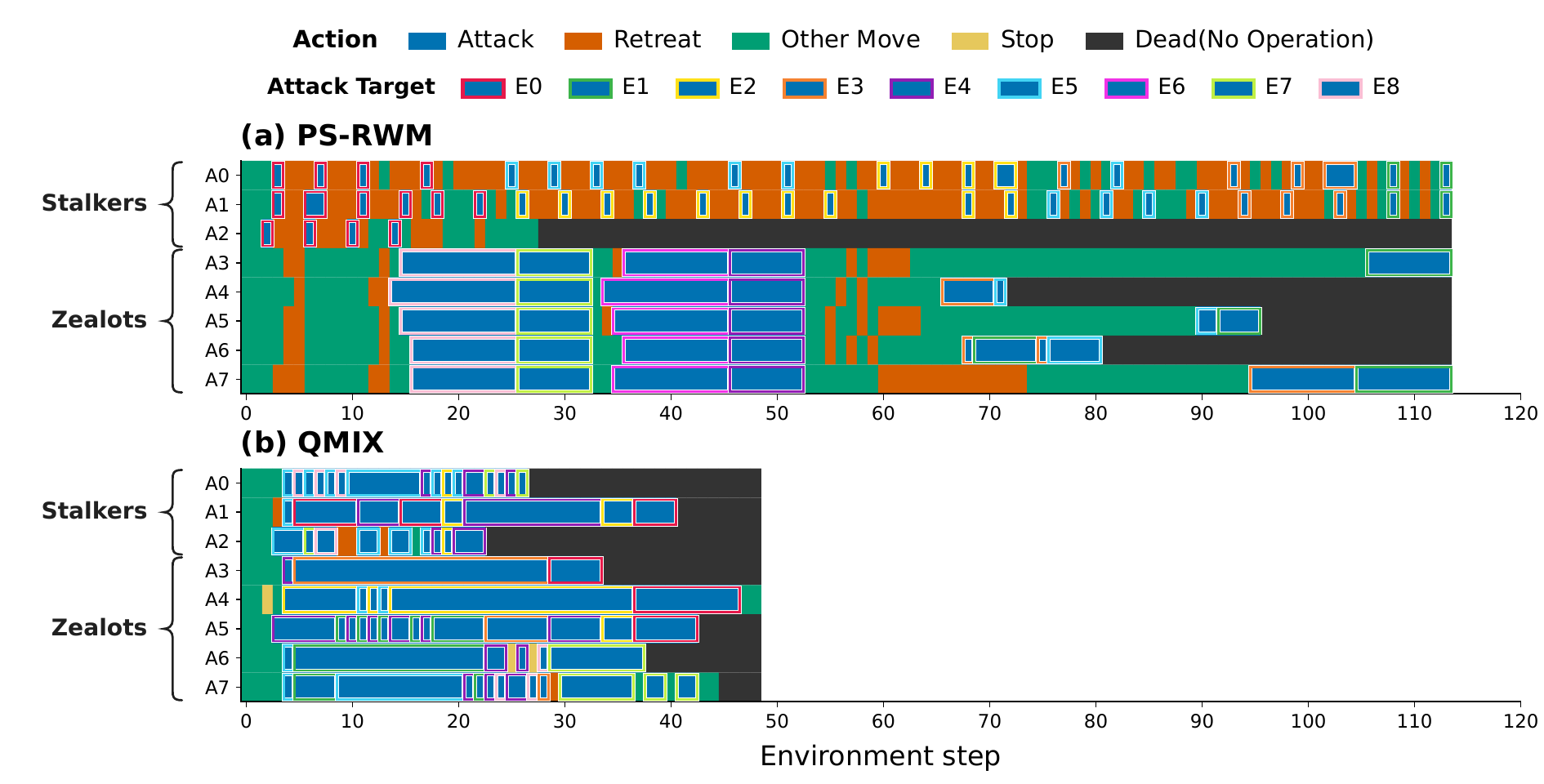}
    \caption{Joint action traces on \texttt{3s5z\_vs\_3s6z} from representative episodes of (a) PS-RWM and (b) QMIX.
Rows represent allied units, and columns represent environment steps.
Cell colors indicate actions, while colored outlines identify attack targets.
Retreat refers to movement away from the nearest living enemy.}
    \label{fig:smac-mixed-action-traces}
\end{figure}

\begin{figure}[t]
    \centering
    \includegraphics[width=0.9\linewidth]{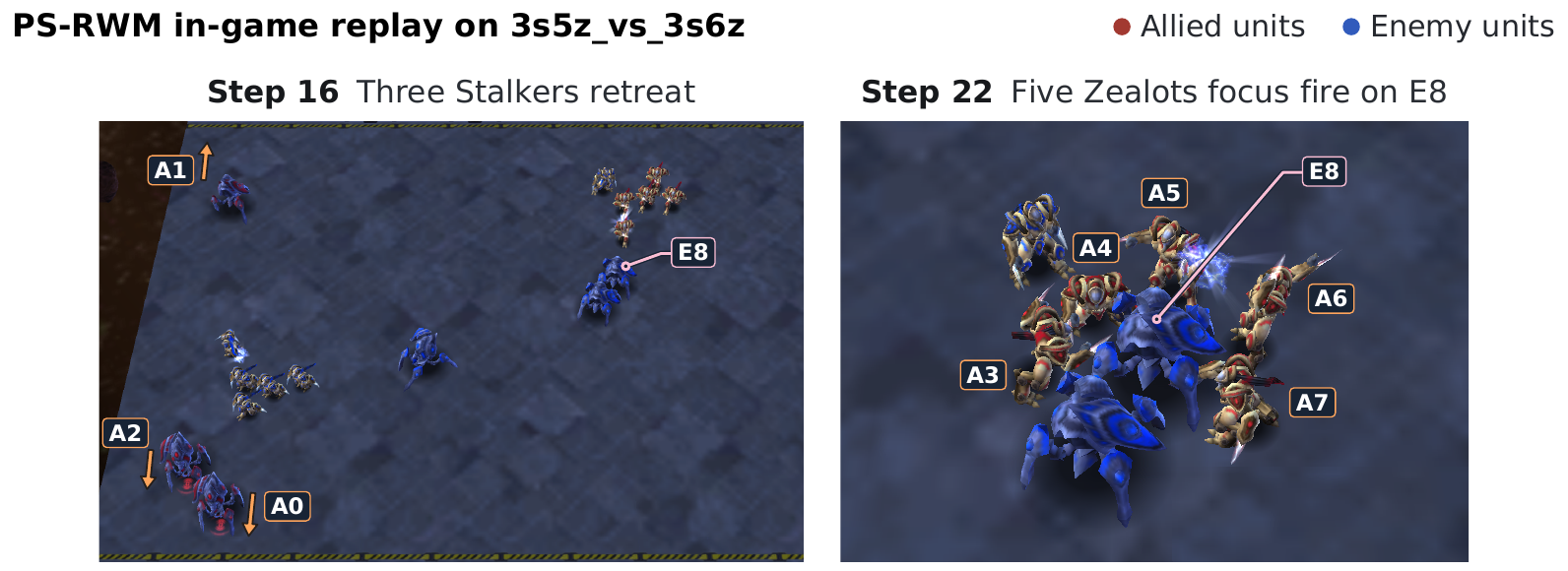}
    \caption{Illustration of PS-RWM's coordinated behavior on \texttt{3s5z\_vs\_3s6z}.
At step 16 (left), three allied Stalkers retreat, as indicated by the orange arrows.
At step 22 (right), all five allied Zealots (A3--A7) focus fire on enemy Stalker E8.\looseness=-1}
    \label{fig:smac-mixed-gameplay}
\end{figure}

To further understand this behavior, Figure~\ref{fig:smac-mixed-action-traces} compares the action sequences of both policies on \texttt{3s5z\_vs\_3s6z}.
PS-RWM exhibits two coordinated combat strategies: \emph{kiting} and \emph{focus fire}.
The allied Stalkers repeatedly retreat between attacks, while all five Zealots concentrate their attacks on enemy Stalker \texttt{E8} during steps 16-25 before switching together to \texttt{E7}.
In contrast, QMIX shows less coordinated behavior in the illustrated episode, with limited retreating and more dispersed target selection.
Figure~\ref{fig:smac-mixed-gameplay} further visualizes these strategies in action, showing the Stalkers retreating while the Zealots jointly attack the same target.
These observations illustrate how the synthesized policy coordinates individual actions to improve team-level behavior.
The underlying decision rules are further examined in Section~\ref{sec:policy-interpretability}.

\vspace{-0.2em}
\textbf{Offline.}\quad
The offline analysis further examines how synthesized policies account for environment dynamics in continuous control.
We consider \texttt{maze2d-large}, where a point mass is controlled through acceleration and receives rewards for remaining inside a goal region.
Since the episode continues after reaching the goal, excessive momentum may cause the agent to leave the rewarding region.
An effective policy therefore needs to regulate its velocity to remain near the goal.
To examine this behavior, we evaluate PS-RWM on recorded observations where the agent is inside the goal region but moving away from its center.
PS-RWM selects actions that oppose the current velocity in 99.87\% of these states, compared with only 28.64\% of the recorded dataset actions.
This difference suggests that the synthesized policy accounts for momentum and favors actions that help maintain the agent within the rewarding region.

These findings reveal that synthesized policies can exhibit meaningful decision-making behaviors across different environments.
The coordinated combat strategies in SMAC and momentum-aware control in Maze2D suggest that PS-RWM can translate environmental knowledge into effective actions.
The following subsections further examine how these behaviors are reflected in the generated policy code.

\subsection{Interpretability of Synthesized Policies}
\label{sec:policy-interpretability}

Unlike neural policies, the executable policies synthesized by PS-RWM expose explicit decision rules that can be directly inspected. 
We examine representative policies from both online and offline tasks to understand how these rules produce effective behaviors.

\vspace{-0.2em}
\textbf{Online.}\quad
Listing~\ref{lst:smac-mixed-policy} presents the policy rules for Stalker kiting and Zealot focus fire in \texttt{3s5z\_vs\_3s6z}.
Each allied unit selects its action based on its local observation, legal-action mask, and previous action.
The listing is abridged, with the complete policy and additional examples provided in Appendix~\ref{app:synthesized-policies}.

For Stalkers, the policy triggers retreat when the nearest visible enemy Zealot is within $2.8$ map units.
This threshold increases to $5.7$ units for fewer than four steps after the last recorded attack.
During retreat, the unit moves away from visible enemies, with closer enemies receiving greater weight in determining the escape direction.
When retreat is no longer necessary, the Stalker resumes attacking if a target is within range.
These rules enable Stalkers to alternate between attacking and retreating based on enemy proximity and recent actions.
For Zealots, the policy prioritizes visible enemy Stalkers based on their distance, health, and shields.
Each Zealot attacks its preferred target when in range and moves toward it otherwise.
This shared targeting rule allows multiple Zealots to independently converge on the same enemy, producing the focus-fire behavior observed in Figure~\ref{fig:smac-mixed-action-traces}(a).

\vspace{-0.2em}
\textbf{Offline.}\quad
Listing~\ref{lst:maze2d-goal-braking} illustrates how the synthesized controller approaches the final waypoint in the offline \texttt{maze2d-large} task.
The controller first computes a desired velocity toward the goal center from the displacement \texttt{(dx, dy)}.
It then determines a bounded action based on the difference between this desired velocity and the current velocity \texttt{v}.
When the agent overshoots the goal center, the desired velocity reverses direction, causing the controller to brake rather than continue moving away.
This feedback mechanism explains the goal-region behavior observed in Section~\ref{sec:Decision-Making Mechanisms}.

These examples show that PS-RWM synthesizes interpretable decision rules rather than opaque action mappings.
The resulting policies exhibit distinct mechanisms for multi-agent coordination and continuous control, which can be directly traced to their executable implementations.

\begin{listing}[t]
\caption{Local decision rules for Stalker kiting and Zealot focus fire in \texttt{3s5z\_vs\_3s6z}.}
\label{lst:smac-mixed-policy}
\begin{lstlisting}[style=smacpolicy,breaklines=true,breakatwhitespace=false]
def make_policy(agent_id):  # ... Initialize this unit's memory.
    def act(context):
        obs = context['observation']
        avail = context['avail_actions']
        # ... Update the local step counter and read visible unit features.
        # ... Handle opening moves and search when no enemy is visible.
        if stalker:
            if size == 136:  # Local observation size on this map.
                melee = [r for _,r in enemies if r[6] > .5]  # Enemy Zealots.
                danger = min((r[1]*9 for r in melee), default=9.)
                kite = danger < 2.8 or (tick-last_shot < 4 and danger < 5.7)
            if kite:
                vx = vy = 0.
                for _, r in enemies:
                    w = 1./max(.1, r[1]*9)**2  # Weight nearby enemies more.
                    vx -= r[2]*9*w  # Steer away from the enemy's x offset.
                    vy -= r[3]*9*w  # Steer away from its y offset.
                return move(vx, vy - (.35 if tick < 25 else 0), True)
        if size == 136 and not stalker:  # Allied Zealot.
            ranged = [(i,r) for i,r in enemies if r[7] > .5]  # Enemy Stalkers.
            if ranged:
                # Rank by health fraction, shield fraction, and distance.
                enemy = min(ranged, key=lambda e:e[1][4]*.2+e[1][5]*.1+e[1][1]*2)
                if avail[6+enemy[0]]:return 6+enemy[0]  # Attack if in range.
                return move(enemy[1][2]*9, enemy[1][3]*9)  # Approach otherwise.
        # ... Handle early movement and score other legal targets.
        if targets:
            target = min(targets,key=target_score)[0]
            if stalker and tick-last_shot >= 4:
                last_shot = tick  # Update this Stalker's attack timer.
            return 6+target
        return move(nearest[2],nearest[3])
    return act
\end{lstlisting}
\end{listing}


\begin{listing}[t]
\caption{Velocity feedback at the final waypoint in \texttt{maze2d-large}.}
\label{lst:maze2d-goal-braking}
\begin{lstlisting}[style=smacpolicy]
speed = min(cfg["max_speed"], cfg["goal_position_gain"] * d)
s = speed / max(d, 1e-12)
vd = (dx * s, dy * s)
gain = cfg["velocity_gain"]
actions = [clip(gain * (vd[j] - v[j]) + cfg["drag_compensation"] * v[j]) for j in range(2)]
\end{lstlisting}
\end{listing}

\subsection{Policy Evolution through Feedback}
\label{sec:policy-evolution}

In this subsection, we examine how PS-RWM uses feedback during development to revise the decision rules of synthesized policies.
By comparing initial and revised policy versions, we investigate how these revisions affect decision-making behavior in both online and offline tasks.

\begin{figure}[!htbp]
    \centering
    \includegraphics[width=\linewidth]{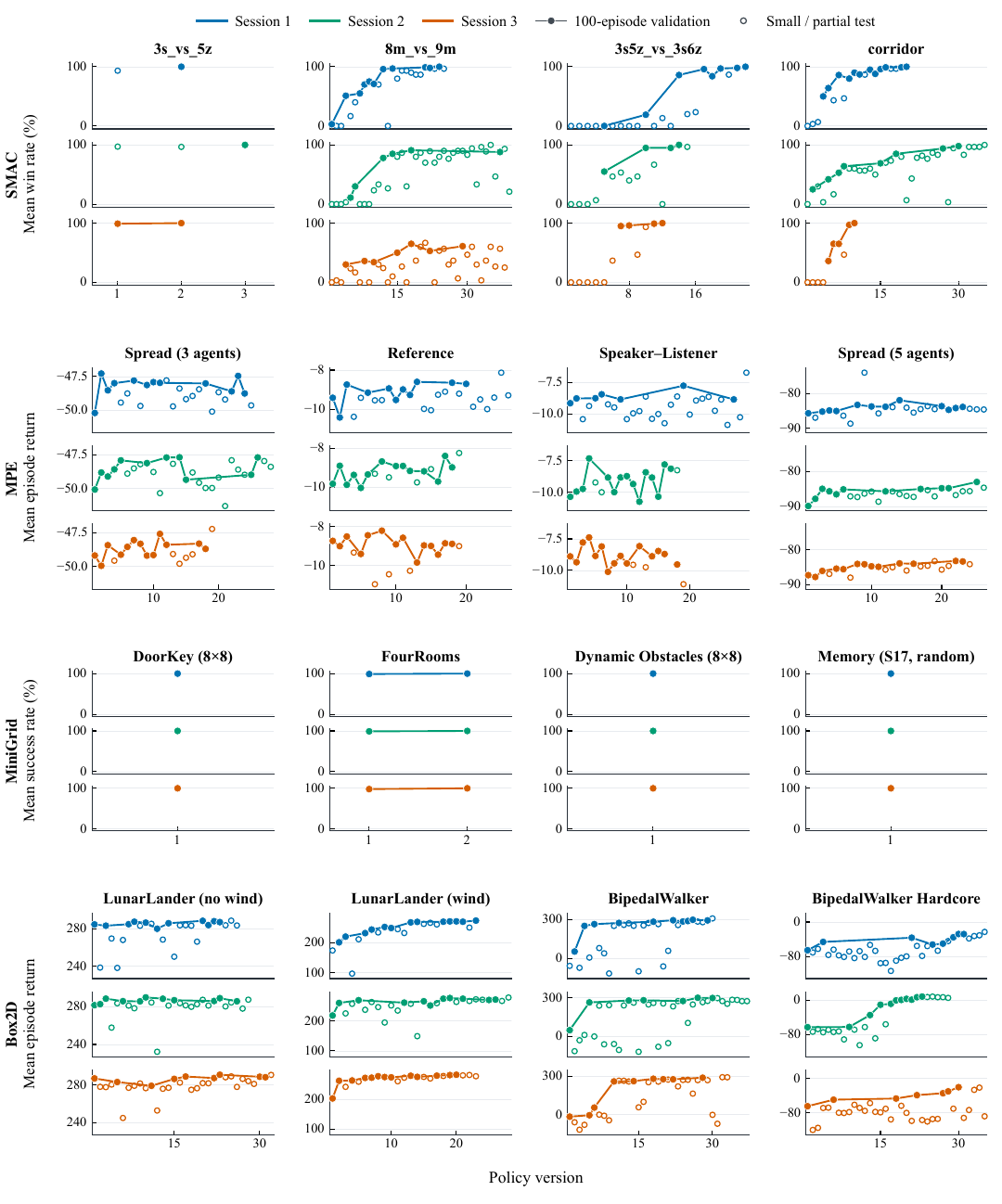}
    \caption{Policy development across 16 online scenarios.
    For each scenario, the three vertically stacked panels represent independent development sessions.
The horizontal axis indicates the policy version, while the vertical axis shows the mean evaluation performance.
Filled markers connected by lines indicate evaluations over 100 episodes, while hollow markers represent smaller or partial tests conducted during development.}
    \label{fig:policy-development-public}
\end{figure}

\begin{figure}[h]
    \centering
    \includegraphics[width=\linewidth]{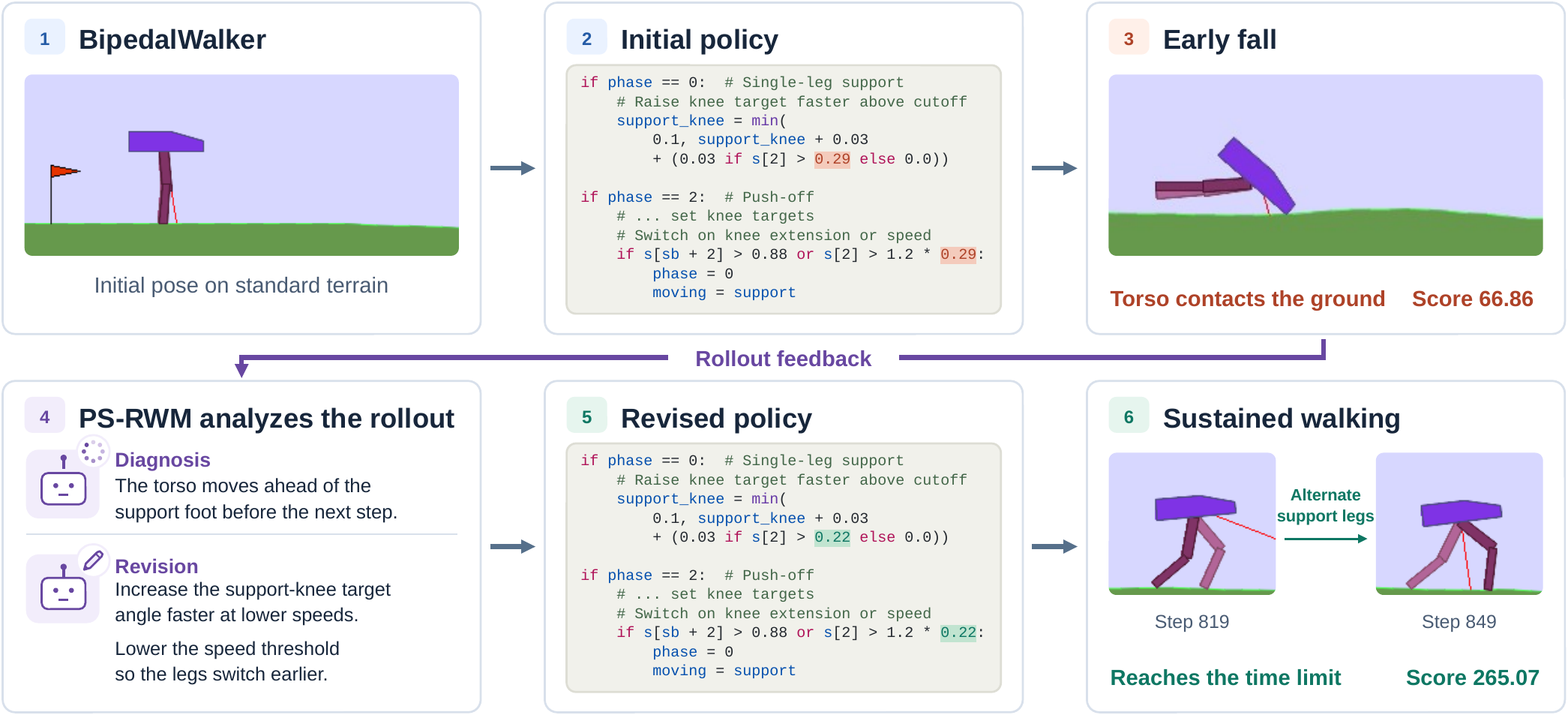}
    \caption{Feedback-driven policy refinement on \texttt{BipedalWalker}.}
    \label{fig:walker-policy-evolution}
\end{figure}

\vspace{-0.2em}
\textbf{Online.}\quad
We analyze successive policy versions synthesized by PS-RWM for \texttt{BipedalWalker-v3}.
In this Box2D environment, a two-legged robot controls its hip and knee joints to walk forward without falling.
Figure~\ref{fig:walker-policy-evolution} illustrates how feedback from early failures leads to revised control rules and improved walking behavior.
In the initial rollout, the torso moves ahead of the support foot before the next step, eventually causing the robot to fall.
After examining these failures, PS-RWM reduces the forward-speed threshold (\texttt{s[2]}) from $0.29$ to $0.22$ in two decision rules.
During single-leg support, this change allows the support-knee target angle to increase more rapidly at lower speeds.
During push-off, it also permits an earlier leg switch when the transition is triggered by forward speed.
With these revised rules, the walker alternates support legs and remains upright until the $1{,}600$-step time limit in the illustrated rollout.
Figure~\ref{fig:policy-development-public} provides additional evaluations of frozen policy versions across all four environment families, with three development sessions per family.

\vspace{-0.2em}
\textbf{Offline.}\quad
Unlike online refinement, offline refinement relies on recorded trajectories rather than feedback from executing candidate policies.
For example, in \texttt{maze2d-umaze-v1}, PS-RWM examines recorded positions and velocities to refine how the controller negotiates a sharp turn.

\begin{figure}[htbp]
    \centering
    \includegraphics[width=\linewidth]{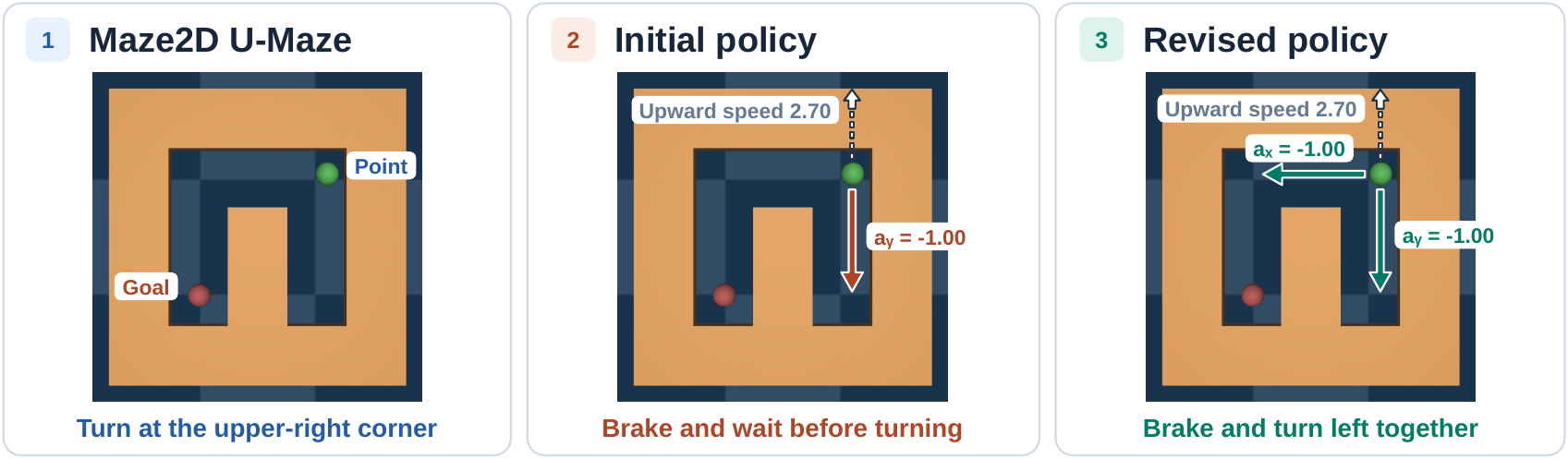}
    \caption{Offline policy refinement at the upper-right turn in \texttt{maze2d-umaze-v1}. All three panels show the same recorded state. Dashed arrows indicate upward velocity, and solid arrows show braking and turning commands.}
    \label{fig:maze2d-umaze-policy-evolution}
\end{figure}

Figure~\ref{fig:maze2d-umaze-policy-evolution} compares the initial and revised policies at the same recorded state.
As shown in the left panel, the agent is at the upper-right corner of the maze, moving upward with a vertical velocity of $2.70$.
The initial policy applies vertical braking ($a_y=-1.00$) and delays turning until the vertical speed falls below $1.10$.
The revised policy instead considers whether the agent has reached the turn and whether its estimated stopping interval remains within the corridor boundaries.
It then initiates a leftward command ($a_x=-1.00$) while continuing to brake vertically.
In the subsequent evaluation from this state, the revised policy reaches the goal region in $99$ steps, compared with $167$ for the initial policy.
The corresponding returns increase from $134$ to $202$.

These analyses show how PS-RWM's synthesized policies make decisions and improve through feedback.
Their decision-making behaviors can be traced to explicit policy rules, providing insight into multi-agent coordination and continuous control.
The evolution of these rules further reveals how PS-RWM incorporates feedback from online rollouts or offline trajectories to refine its policies.


\FloatBarrier

\section{The Role of Readable World Models}
\label{sec:world-model-access}

The preceding experiments demonstrate the performance of PS-RWM when readable world models are available.
This section investigates how access to environment implementations affects policy synthesis.
We compare PS-RWM with and without such access under matched experimental conditions, covering both online and offline tasks.

\subsection{With vs. Without Readable World Models}
\label{sec:world-model-access-comparison}

\vspace{-0.2em}
\textbf{Online.}\quad
We first compare policy synthesis with and without access to readable world models across 16 online scenarios.
Both conditions use the same underlying model, reasoning effort, task instructions, and environment versions.
With readable world models, PS-RWM can inspect the environment implementation during development.
Without such access, it relies on task descriptions and public interface documentation.
Both conditions use the same observation and action interfaces and can refine policies through environment feedback.
Both conditions follow the evaluation protocol in Section~\ref{sec:experimental-setup}.

Table~\ref{tab:world-model-access} shows that access to readable world models improves the mean development-validation score in eight scenarios, while the condition without access performs better in three and ties in five.
Both conditions achieve 100\% success rates on all four MiniGrid tasks and \texttt{3s\_vs\_5z}.
Access to readable world models provides higher scores on the remaining three SMAC scenarios, while differences across MPE tasks remain below 0.50.
For Box2D, the condition without access performs slightly better on both \texttt{LunarLander} variants, whereas access improves performance on both \texttt{BipedalWalker} variants.
The largest difference occurs on \texttt{BipedalWalker-v3}.
The condition without access also exhibits greater variation across development sessions.

\begin{table*}[h]
\centering
\caption{
Performance comparison with and without access to readable world models across 16 online scenarios.
Results are reported as mean $\pm$ sample standard deviation over three runs.
}
\label{tab:world-model-access}
\begingroup
\setlength{\tabcolsep}{4pt}
\renewcommand{\arraystretch}{1.12}
\footnotesize

\begin{tabular}{@{}llrr>{\columncolor{black!5}}r@{}}
\toprule
\multirow{2}{*}{\textbf{Benchmark}}
& \multirow{2}{*}{\textbf{Scenario}}
& \multirow{2}{*}{\textbf{Model-Free RL}}
& \multicolumn{2}{c}{\textbf{PS-RWM}} \\
\cmidrule(lr){4-5}
&
&
& \textbf{Without RWM}
& \textbf{With RWM} \\
\midrule

\multirow{5}{*}{\textbf{SMAC}}
& \texttt{3s\_vs\_5z}
& $69.79 \pm 28.36$
& $\boldsymbol{100.00 \pm 0.00}$
& $\boldsymbol{100.00 \pm 0.00}$ \\

& \texttt{8m\_vs\_9m}
& $\boldsymbol{94.79 \pm 6.51}$
& $79.00 \pm 7.00$
& $85.33 \pm 18.18$ \\

& \texttt{3s5z\_vs\_3s6z}
& $6.25 \pm 5.41$
& $93.00 \pm 8.19$
& $\boldsymbol{100.00 \pm 0.00}$ \\

& \texttt{corridor}
& $0.00 \pm 0.00$
& $94.00 \pm 7.94$
& $\boldsymbol{99.33 \pm 1.15}$ \\

\cmidrule(l){2-5}
& \textbf{Average}
& $42.71 \pm 5.32$
& $91.50 \pm 4.18$
& $\boldsymbol{96.17 \pm 4.47}$ \\

\midrule

\multirow{5}{*}{\textbf{MPE}}
& \texttt{Spread (3 agents)}
& $-74.10 \pm 7.43$
& $\boldsymbol{-47.26 \pm 0.47}$
& $-47.52 \pm 0.21$ \\

& \texttt{Reference}
& $-9.94 \pm 0.35$
& $-8.58 \pm 0.09$
& $\boldsymbol{-8.40 \pm 0.19}$ \\

& \texttt{Speaker--Listener}
& $-8.28 \pm 0.24$
& $-7.58 \pm 0.11$
& $\boldsymbol{-7.48 \pm 0.23}$ \\

& \texttt{Spread (5 agents)}
& $-186.36 \pm 2.39$
& $-83.21 \pm 0.68$
& $\boldsymbol{-82.71 \pm 0.67}$ \\

\cmidrule(l){2-5}
& \textbf{Average}
& $-69.67 \pm 1.21$
& $-36.66 \pm 0.33$
& $\boldsymbol{-36.53 \pm 0.11}$ \\

\midrule

\multirow{5}{*}{\textbf{MiniGrid}}
& \texttt{DoorKey ($8\times8$)}
& $0.00 \pm 0.00$
& $\boldsymbol{100.00 \pm 0.00}$
& $\boldsymbol{100.00 \pm 0.00}$ \\

& \texttt{FourRooms}
& $1.33 \pm 1.15$
& $\boldsymbol{100.00 \pm 0.00}$
& $\boldsymbol{100.00 \pm 0.00}$ \\

& \texttt{Dynamic Obstacles ($8\times8$)}
& $90.67 \pm 4.16$
& $\boldsymbol{100.00 \pm 0.00}$
& $\boldsymbol{100.00 \pm 0.00}$ \\

& \texttt{Memory (S17, random)}
& $22.00 \pm 19.08$
& $\boldsymbol{100.00 \pm 0.00}$
& $\boldsymbol{100.00 \pm 0.00}$ \\

\cmidrule(l){2-5}
& \textbf{Average}
& $28.50 \pm 4.27$
& $\boldsymbol{100.00 \pm 0.00}$
& $\boldsymbol{100.00 \pm 0.00}$ \\

\midrule

\multirow{5}{*}{\textbf{Box2D}}
& \texttt{LunarLander (no wind)}
& $228.44 \pm 15.70$
& $\boldsymbol{291.38 \pm 2.19}$
& $289.60 \pm 1.29$ \\

& \texttt{LunarLander (wind)}
& $148.37 \pm 29.76$
& $\boldsymbol{280.83 \pm 2.30}$
& $277.57 \pm 3.74$ \\

& \texttt{BipedalWalker}
& $208.81 \pm 62.35$
& $236.03 \pm 79.24$
& $\boldsymbol{295.62 \pm 7.63}$ \\

& \texttt{BipedalWalker Hardcore}
& $-95.94 \pm 5.13$
& $-28.22 \pm 28.59$
& $\boldsymbol{-13.38 \pm 19.17}$ \\

\cmidrule(l){2-5}
& \textbf{Average}
& $122.42 \pm 17.62$
& $195.01 \pm 23.64$
& $\boldsymbol{212.35 \pm 5.71}$ \\

\bottomrule
\end{tabular}

\par\vspace{3pt}
\begin{minipage}{\linewidth}
\scriptsize
\end{minipage}

\endgroup
\end{table*}

\vspace{-0.2em}
\textbf{Offline.}\quad
We further examine the role of readable world models on the offline Kitchen benchmark.
Both conditions receive the same full offline dataset, four task goals, and public policy interface.
With readable world models, PS-RWM additionally accesses the environment implementation and implementation notes, including object coordinates and task completion thresholds.
Without such access, policy synthesis relies only on the provided dataset and task information.
Three non-neural policies are constructed under each condition without environment interaction.
Each frozen policy is evaluated on the same 100 seeds using public observations and within-episode history.
We report the D4RL normalized score and full completion rate, which measure the average percentage of completed tasks and the percentage of episodes completing all four tasks, respectively.

\begin{table*}[h]
\centering
\caption{
Performance comparison with and without access to readable world models on offline Kitchen tasks.
Results are reported as mean $\pm$ sample standard deviation over three independently constructed policies.
Full completion denotes the percentage of episodes in which all four tasks are completed.
}
\label{tab:kitchen-offline-source-access}
\begingroup
\setlength{\tabcolsep}{6pt}
\renewcommand{\arraystretch}{1.12}
\footnotesize

\begin{tabular}{@{}l r>{\columncolor{black!5}}r r>{\columncolor{black!5}}r@{}}
\toprule
& \multicolumn{2}{c}{\textbf{D4RL Normalized Score}}
& \multicolumn{2}{c}{\textbf{Full Completion (\%)}} \\
\cmidrule(lr){2-3}
\cmidrule(l){4-5}

\textbf{Dataset}
& \textbf{Without RWM}
& \textbf{With RWM}
& \textbf{Without RWM}
& \textbf{With RWM} \\
\midrule

\texttt{kitchen-complete}
& $60.75 \pm 31.95$
& $\boldsymbol{84.25 \pm 8.11}$
& $41.00 \pm 37.99$
& $\boldsymbol{63.33 \pm 17.04}$ \\

\texttt{kitchen-partial}
& $40.92 \pm 12.88$
& $\boldsymbol{96.17 \pm 2.84}$
& $3.00 \pm 5.20$
& $\boldsymbol{92.33 \pm 5.69}$ \\

\bottomrule
\end{tabular}
\endgroup
\end{table*}


As shown in Table~\ref{tab:kitchen-offline-source-access}, access to readable world models substantially improves both normalized scores and full completion rates on offline Kitchen tasks.
The improvement is particularly pronounced on \texttt{kitchen-partial}, where successful demonstrations are relatively scarce.
These results suggest that readable world models can provide valuable guidance for policy synthesis when offline data offer limited examples of successful task completion.
Dataset sizes are provided in Appendix~\ref{app:offline-datasets}.


Overall, the results from both online and offline tasks demonstrate the value of readable world models for policy synthesis, although their benefits vary across scenarios.
The improvements are particularly evident in challenging control tasks and offline settings with limited successful demonstrations, highlighting the value of environment knowledge beyond task descriptions and interaction data.

\subsection{How World Model Access Changes Policy Synthesis}
\label{sec:world-model-access-mechanisms}

\begin{wrapfigure}{r}{0.56\textwidth}
  \vspace{-15pt}
  \centering
  \includegraphics[width=\linewidth]{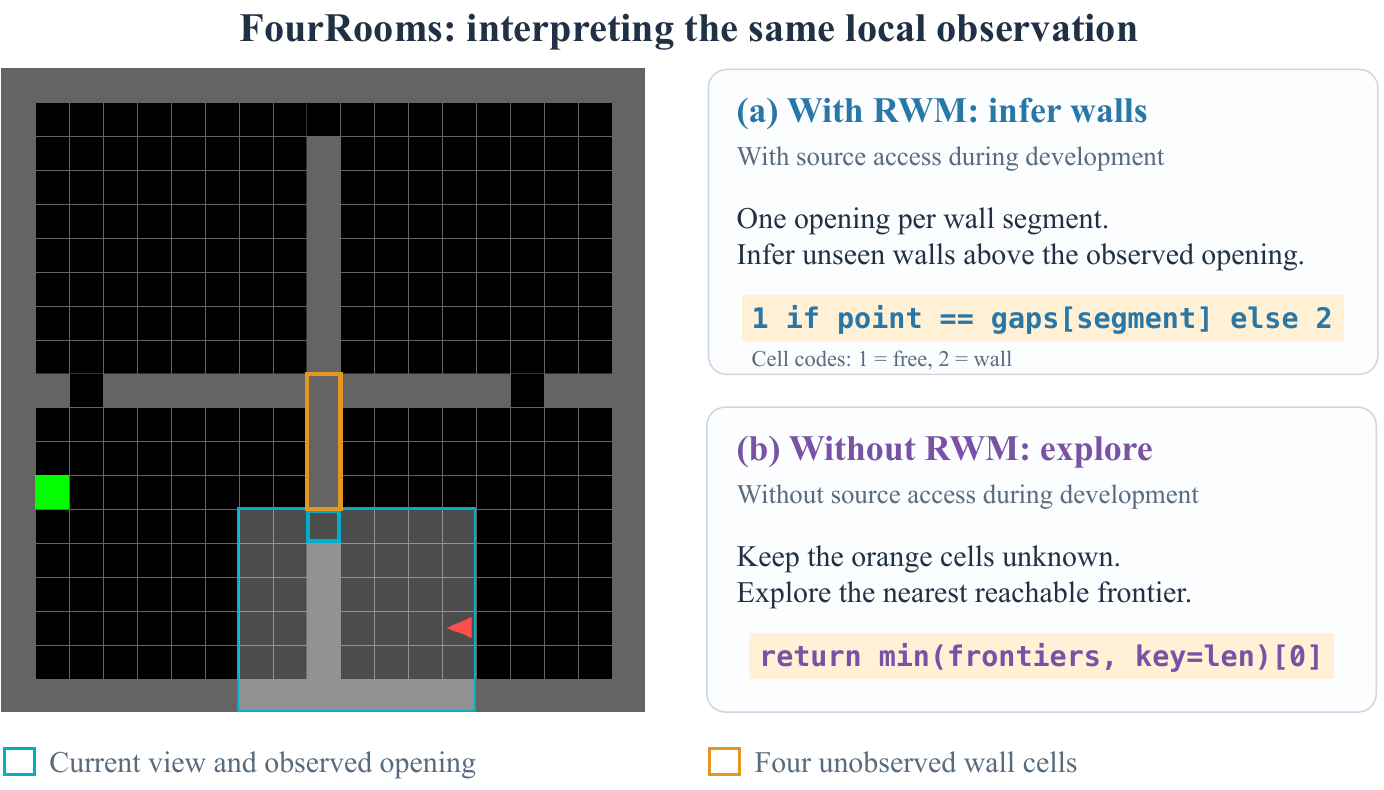}
  \captionsetup{font=footnotesize,skip=3pt}
  \caption{\texttt{FourRooms} navigation with and without readable world model access.
PS-RWM infers unobserved walls from environment rules, while policies without source access retain them as unknown.}
  \label{fig:fourrooms-source-access}
  \vspace{-8pt}
\end{wrapfigure}
Beyond performance comparisons, we further investigate how access to readable world models influences the policy synthesis process.
Two representative examples illustrate how explicit environmental knowledge can guide decision-making and improve the utilization of offline experience.

\vspace{-0.2em}
\textbf{Online.}\quad
In MiniGrid \texttt{FourRooms}, PS-RWM incorporates the environment geometry and doorway rules into its decision logic.
As illustrated in Figure~\ref{fig:fourrooms-source-access}, knowing that each internal wall contains a single opening allows the policy to infer walls that have not yet been observed.
In contrast, policies synthesized without readable world model access construct their maps from observations and explore unknown regions.
Although both settings achieve 100\% success, PS-RWM requires only 28.4 actions per episode on average, compared with 34.4 without source access.
This suggests that readable world models can improve decision efficiency by providing structural knowledge beyond immediate observations. Further details are provided in Appendix~\ref{app:fourrooms-construction}.

\vspace{-0.2em}
\textbf{Offline.}\quad
A different benefit emerges in Kitchen, where readable world models help extract task-specific information from offline trajectories.
Although recorded rewards indicate task progress, they do not directly identify which objects have reached their target states.
By examining the task completion rules, PS-RWM can distinguish the outcomes of different action sequences and organize the data accordingly.
In two recorded segments with the same reward at step 60, the next objects to reach their reference configurations are the bottom burner and the kettle, respectively.
PS-RWM uses these events to label action sequences by object, keeping actions associated with the unscored burner out of the kettle's reference set.
Additional details are provided in Appendix~\ref{app:kitchen-readable-model}.

These examples demonstrate how readable world models provide useful knowledge beyond direct interaction experience.
Such knowledge supports inference about unobserved environmental structure and enables more informative use of offline data.


\section{Policy Synthesis under Challenging Conditions}

Readable world models provide explicit knowledge of environmental mechanisms, which may offer advantages under challenging decision-making conditions.
Sparse rewards provide limited feedback for discovering effective behaviors.
Partial observability restricts the information available during policy execution.
Task generalization further requires synthesized policies to remain effective under changing task configurations.
This section investigates whether PS-RWM can address these challenges by leveraging environmental knowledge rather than relying primarily on trial and error.

\subsection{Sparse-Reward Environments}
\label{sec:sparse-rewards}
Sparse rewards pose a major challenge to RL, as infrequent feedback makes effective behaviors difficult to discover through trial and error~\cite{AndrychowiczCRS17,EcoffetHLSC21}.
To examine whether readable world models can alleviate this challenge, we investigate policy synthesis under sparse rewards in both online and offline settings.
For a fair comparison, the source code provided to the coding agent contains only the sparse-reward implementation, with all dense-reward shaping terms removed.

\textbf{Online.}\quad
We compare PS-RWM with conventional RL methods under dense and sparse reward settings across SMAC, MiniGrid, and Box2D, as reported in Table~\ref{tab:sec7-sparse_reward}.
For each reward setting, RL baselines are trained separately, while PS-RWM independently constructs policies from the corresponding readable world models.
A dense-reward variant is additionally introduced for \texttt{DoorKey}, which originally provides only terminal rewards.
The results show that conventional RL methods suffer substantial performance degradation under sparse rewards.
In particular, QMIX and DQN fail to achieve successful episodes, while PPO exhibits a significant decline in landing performance.
In contrast, PS-RWM maintains strong performance across these benchmarks, with nearly unchanged success or win rates on SMAC and MiniGrid.
Similar results are observed in Box2D, where PS-RWM achieves nearly perfect landing success despite the removal of intermediate rewards.
Notably, the lower numerical returns under sparse rewards primarily reflect the removal of reward shaping rather than degraded control performance.

\begin{table}[!htbp]
\centering
\caption{Performance under dense and sparse rewards across SMAC (win rate, \%), MiniGrid (success rate, \%), and Box2D (episode return), reported as mean $\pm$ standard deviation over three seeds. RL baselines are QMIX (SMAC), DQN (MiniGrid), and PPO (Box2D). Both RL baselines and PS-RWM are trained or synthesized separately for each reward setting. Bold indicates the higher score between RL and PS-RWM within each reward setting.}
\label{tab:sec7-sparse_reward}
\begingroup
\footnotesize
\setlength{\tabcolsep}{4pt}
\renewcommand{\arraystretch}{1.10}
\begin{tabular}{@{}llrr>{\columncolor{black!5}}r>{\columncolor{black!5}}r@{}}
\toprule
\multirow{2}{*}{\textbf{Benchmark}} & \multirow{2}{*}{\textbf{Scenario}} & \multicolumn{2}{c}{\textbf{RL baseline}}  & \multicolumn{2}{c}{\textbf{PS-RWM}}                                      \\ \cmidrule(lr){3-4} \cmidrule(l){5-6} 
 &  & \multicolumn{1}{c}{\textbf{Dense}}  & \multicolumn{1}{c}{\textbf{Sparse}} 
   & \multicolumn{1}{c}{\textbf{Dense}}   & \multicolumn{1}{c}{\textbf{Sparse}} \\
 \midrule
\multirow{4}{*}{\textbf{SMAC}}  
& \texttt{3s5z\_vs\_3s6z}            
& $6.25 \pm 5.41$             & $0.00 \pm 0.00$    & $\boldsymbol{100.00 \pm 0.00}$ & $\boldsymbol{100.00 \pm 0.00}$ \\
& \texttt{3s\_vs\_5z}                
& $69.79 \pm 28.36$             & $0.00 \pm 0.00$    & $\boldsymbol{100.00 \pm 0.00}$ & $\boldsymbol{100.00 \pm 0.00}$ \\
& \texttt{8m\_vs\_9m}                
& $\boldsymbol{94.79 \pm 6.51}$ & $0.00 \pm 0.00$    & $85.33 \pm 18.18$              & $\boldsymbol{55.67 \pm  1.53}$              \\
& \texttt{corridor}                  
& $0.00 \pm 0.00$               & $0.00 \pm 0.00$    & $\boldsymbol{99.33 \pm 1.15}$  & $\boldsymbol{100.00 \pm 0.00}$  \\
\midrule 
\textbf{MiniGrid}               
& \texttt{DoorKey ($8\times8$)}             
& $11.67 \pm 5.77$              & $0.00 \pm 0.00$    & $\boldsymbol{100.00 \pm 0.00}$  &$\boldsymbol{100.00 \pm 0.00}$  \\
\midrule 
\multirow{2}{*}{\textbf{Box2D}} 
& \texttt{LunarLander (no wind)}     
& $228.44 \pm 15.70$            & $-11.33 \pm 16.29$ & $\boldsymbol{289.60 \pm 1.29}$ & $\boldsymbol{100.00 \pm 0.00}$  \\
                       
& \texttt{LunarLander (wind)}
& $148.37 \pm 29.76$            & $-18.67 \pm 11.02$ & $\boldsymbol{277.57 \pm 3.74}$   & $\boldsymbol{97.33 \pm 1.15}$                              \\ \bottomrule
\end{tabular}
\endgroup
\end{table}

\vspace{-0.2em}
\textbf{Offline.}\quad
The offline analysis further examines whether useful control strategies can be derived from datasets containing limited reward information.
As shown in Table~\ref{tab:d4rl-comparison}, PS-RWM outperforms the reported baselines across all three Maze2D layouts, where rewards are obtained only within the goal region.
In \texttt{maze2d-large}, positive rewards account for less than 1\% of recorded transitions in a representative construction.
Nevertheless, the agent can exploit the recorded movements and environment dynamics to construct navigation and velocity-control strategies.
A similar phenomenon appears in AntMaze, where trajectories without positive returns can still provide useful information about locomotion.
Such experience allows the coding agent to derive gait-related control rules, although performance remains less consistent across different AntMaze datasets.

These results demonstrate that readable world models can substantially reduce the dependence of policy synthesis on dense reward feedback.
Across both online and offline settings, PS-RWM can construct effective policies by reasoning over environmental mechanisms and extracting useful information from experience, even when rewards are sparse.
This highlights the potential of direct policy synthesis for decision-making problems where reward-driven learning is particularly challenging.


\subsection{Partial Observability}

Partial observability poses a fundamental challenge to sequential decision-making, as individual observations may not fully reveal the underlying environment state~\cite{KaelblingLC98}.
This challenge becomes more pronounced in multi-agent environments, where agents must coordinate their actions based on limited local information~\cite{abs-1111-0062}.
To investigate this issue, we examine the performance of PS-RWM under varying observation scopes in both online and offline settings.

\begin{table*}[t]
\centering
\caption{
Performance comparison under different observation settings on MiniGrid and SMAC.
Results are reported as mean $\pm$ standard deviation over three seeds.
PS-RWM policies synthesized under the default observation settings are directly reused, with one-round interaction-free revisions evaluated under altered settings.
}
\label{tab:sec7-partial_obs}
\begingroup
\setlength{\tabcolsep}{4pt}
\renewcommand{\arraystretch}{1.12}
\footnotesize

\begin{tabular}{@{}llcr>{\columncolor{black!5}}r>{\columncolor{black!5}}r@{}}
\toprule

\multirow{2}{*}{\textbf{Benchmark}}
& \multirow{2}{*}{\textbf{Scenario}}
& \multirow{2}{*}{\textbf{Observation}}
& \multirow{2}{*}{\textbf{RL Baseline}}
& \multicolumn{2}{c}{\textbf{PS-RWM}} \\
\cmidrule(lr){5-6}

& & &
& \textbf{Original Policy}
& \textbf{One-Round Revision} \\

\midrule

\multirow{13}{*}{\textbf{MiniGrid}}
& &
&
& \multicolumn{1}{c}{}
& \multicolumn{1}{c}{} \\
\noalign{\vskip-10pt}

& \multirow{3}{*}{\texttt{DoorKey ($8\times8$)}}
& Full
& $0.00 \pm 0.00$
& $0.00 \pm 0.00$
& $\boldsymbol{100.00 \pm 0.00}$ \\

& & $7\times7$
& $0.00 \pm 0.00$
& $\boldsymbol{100.00 \pm 0.00}$
& --- \\

& & $5\times5$
& $0.00 \pm 0.00$
& $0.00 \pm 0.00$
& $\boldsymbol{100.00 \pm 0.00}$ \\

\cmidrule(l){2-6}

& \multirow{3}{*}{\texttt{FourRooms}}
& Full
& $2.67 \pm 2.31$
& $11.33 \pm 0.58$
& $\boldsymbol{100.00 \pm 0.00}$ \\

& & $7\times7$
& $1.33 \pm 1.15$
& $\boldsymbol{100.00 \pm 0.00}$
& --- \\

& & $5\times5$
& $3.33 \pm 3.06$
& $1.00 \pm 1.00$
& $\boldsymbol{98.67 \pm 1.53}$ \\

\cmidrule(l){2-6}

& \multirow{3}{*}{\texttt{Dynamic Obstacles ($8\times8$)}}
& Full
& $4.67 \pm 8.08$
& $1.33 \pm 2.31$
& $\boldsymbol{100.00 \pm 0.00}$ \\

& & $7\times7$
& $90.67 \pm 4.16$
& $\boldsymbol{100.00 \pm 0.00}$
& --- \\

& & $5\times5$
& $87.33 \pm 3.06$
& $0.00 \pm 0.00$
& $\boldsymbol{100.00 \pm 0.00}$ \\

\cmidrule(l){2-6}

& \multirow{3}{*}{\texttt{Memory (S17, random)}}
& Full
& $40.00 \pm 24.25$
& $0.00 \pm 0.00$
& $\boldsymbol{100.00 \pm 0.00}$ \\

& & $7\times7$
& $22.00 \pm 19.08$
& $\boldsymbol{100.00 \pm 0.00}$
& --- \\

& & $5\times5$
& $49.33 \pm 5.03$
& $0.00 \pm 0.00$
& $\boldsymbol{100.00 \pm 0.00}$ \\

\midrule

\multirow{10}{*}{\textbf{SMAC}}
& &
&
& \multicolumn{1}{c}{}
& \multicolumn{1}{c}{} \\
\noalign{\vskip-10pt}

& \multirow{3}{*}{\texttt{3s\_vs\_5z}}
& \texttt{15s}
& $66.00 \pm 43.35$
& $\boldsymbol{100.00 \pm 0.00}$
& $\boldsymbol{100.00 \pm 0.00}$ \\

& & \texttt{9s}
& $69.79 \pm 28.36$
& $\boldsymbol{100.00 \pm 0.00}$
& --- \\

& & \texttt{5s}
& $83.67 \pm 25.70$
& $93.33 \pm 0.58$
& $\boldsymbol{96.33 \pm 1.53}$ \\

\cmidrule(l){2-6}

& \multirow{3}{*}{\texttt{8m\_vs\_9m}}
& \texttt{15s}
& $96.67 \pm 2.52$
& $\boldsymbol{97.67 \pm 1.15}$
& $71.33 \pm 3.51$ \\

& & \texttt{9s}
& $\boldsymbol{94.79 \pm 6.51}$
& $85.33 \pm 18.18$
& --- \\

& & \texttt{5s}
& $\boldsymbol{95.33 \pm 1.53}$
& $4.00 \pm 1.73$
& $1.67 \pm 1.15$ \\

\cmidrule(l){2-6}

& \multirow{3}{*}{\texttt{3s5z\_vs\_3s6z}}
& \texttt{15s}
& $21.67 \pm 36.67$
& $\boldsymbol{100.00 \pm 0.00}$
& $\boldsymbol{100.00 \pm 0.00}$ \\

& & \texttt{9s}
& $6.25 \pm 5.41$
& $\boldsymbol{100.00 \pm 0.00}$
& --- \\

& & \texttt{5s}
& $31.00 \pm 53.69$
& $5.67 \pm 1.53$
& $\boldsymbol{34.00 \pm 6.08}$ \\

\bottomrule
\end{tabular}

\endgroup
\end{table*}

\textbf{Online.}\quad
Table~\ref{tab:sec7-partial_obs} compares PS-RWM with conventional RL methods under different observation settings in MiniGrid and SMAC.
The results show that broader observations do not necessarily lead to better RL performance.
DQN struggles on several MiniGrid tasks requiring historical information, while QMIX exhibits considerable performance variations across different sight radii.
In contrast, PS-RWM achieves perfect success across all four MiniGrid tasks under the original $7\times7$ observation setting.
The generated policies explicitly incorporate memory and state inference to compensate for incomplete observations.
For example, the \texttt{Memory} policy retains previously observed cues, while the \texttt{FourRooms} policy infers unobserved map structures from accumulated information.
Directly transferring these policies to different observation dimensions leads to substantial degradation due to mismatches in observation handling.
However, a single source-guided revision without additional environment interaction restores nearly perfect success under both full and $5\times5$ observations.
This demonstrates the ability of coding agents to adapt existing policies to new observation formats through code reasoning.

The SMAC experiments further examine policy adaptation under changing local visibility.
Unlike MiniGrid, changing the sight radius preserves the observation vector format, allowing direct execution of the original policies.
With expanded visibility, PS-RWM maintains strong performance and outperforms QMIX across the evaluated scenarios.
Under restricted visibility, however, the results become more task-dependent.
Some policies remain competitive, while others suffer substantial degradation because their decision rules rely on information about nearby units.
A single interaction-free revision improves performance in certain scenarios but can also weaken previously effective strategies.
These results suggest that explicit decision rules can support adaptation to changing visibility, although their effectiveness depends on the information available during execution.

\begin{table}[t]
\centering
\caption{Offline BabyAI success rates (\%) and mean episode lengths (steps) under different view sizes.}
\label{tab:sec7-offline-partial-obs}
\begingroup
\footnotesize
\setlength{\tabcolsep}{5pt}
\renewcommand{\arraystretch}{1.0}
\begin{tabular}{ll*{3}{rr}}
\toprule
\multirow{2}{*}{\textbf{Task}} & \multirow{2}{*}{\textbf{Method}}
& \multicolumn{2}{c}{$\boldsymbol{7\times7}$}
& \multicolumn{2}{c}{$\boldsymbol{5\times5}$}
& \multicolumn{2}{c}{$\boldsymbol{3\times3}$} \\
\cmidrule(lr){3-4}\cmidrule(lr){5-6}\cmidrule(lr){7-8}
& & \textbf{Success (\%)} & \textbf{Length}
& \textbf{Success (\%)} & \textbf{Length}
& \textbf{Success (\%)} & \textbf{Length} \\
\midrule
\rowcolor{black!5}
\cellcolor{white}\multirow{7}{*}{\texttt{GoToRedBall}} & \textbf{PS-RWM} & $\mathbf{100.0}$ & $\mathbf{5.4}$ & $\mathbf{100.0}$ & $\mathbf{6.0}$ & $\mathbf{100.0}$ & $\mathbf{9.0}$ \\
 & BC & $94.3$ & $9.3$ & $94.7$ & $10.6$ & $89.7$ & $17.1$ \\
 & TD3+BC & $91.7$ & $11.0$ & $92.7$ & $11.0$ & $91.3$ & $15.9$ \\
 & IQL & $93.0$ & $10.3$ & $92.7$ & $11.8$ & $87.0$ & $19.1$ \\
 & CQL & $96.3$ & $8.5$ & $95.3$ & $9.5$ & $97.0$ & $13.7$ \\
 & BEAR & $90.3$ & $11.8$ & $89.7$ & $12.8$ & $85.3$ & $20.2$ \\
 & DT & $98.7$ & $8.4$ & $99.0$ & $9.7$ & $98.0$ & $13.7$ \\
\midrule
\rowcolor{black!5}
\cellcolor{white}\multirow{7}{*}{\texttt{GoToLocal}} & \textbf{PS-RWM} & $\mathbf{100.0}$ & $\mathbf{4.9}$ & $\mathbf{100.0}$ & $\mathbf{5.5}$ & $\mathbf{100.0}$ & $\mathbf{8.3}$ \\
 & BC & $73.7$ & $30.9$ & $74.3$ & $30.6$ & $75.0$ & $31.0$ \\
 & TD3+BC & $75.7$ & $29.5$ & $73.3$ & $29.4$ & $73.7$ & $30.8$ \\
 & IQL & $74.0$ & $30.2$ & $73.3$ & $31.1$ & $71.7$ & $32.4$ \\
 & CQL & $76.3$ & $29.4$ & $75.0$ & $29.2$ & $69.7$ & $32.1$ \\
 & BEAR & $73.0$ & $29.5$ & $70.0$ & $30.8$ & $71.0$ & $31.5$ \\
 & DT & $80.3$ & $26.2$ & $79.0$ & $26.8$ & $82.7$ & $28.3$ \\
\midrule
\rowcolor{black!5}
\cellcolor{white}\multirow{7}{*}{\texttt{GoToSeq}} & \textbf{PS-RWM} & $\mathbf{100.0}$ & $\mathbf{60.2}$ & $\mathbf{98.3}$ & $\mathbf{134.3}$ & $\mathbf{100.0}$ & $\mathbf{135.8}$ \\
 & BC & $52.0$ & $668.4$ & $50.3$ & $685.6$ & $41.7$ & $781.3$ \\
 & TD3+BC & $41.0$ & $735.0$ & $40.3$ & $764.2$ & $34.3$ & $814.1$ \\
 & IQL & $53.3$ & $662.4$ & $51.7$ & $668.7$ & $47.0$ & $735.8$ \\
 & CQL & $55.0$ & $640.2$ & $49.7$ & $687.0$ & $42.3$ & $785.2$ \\
 & BEAR & $56.7$ & $642.3$ & $34.7$ & $830.7$ & $26.0$ & $891.3$ \\
 & DT & $70.3$ & $577.1$ & $71.3$ & $577.7$ & $73.3$ & $558.2$ \\
\midrule
\rowcolor{black!5}
\cellcolor{white}\multirow{7}{*}{\texttt{PutNextLocal}} & \textbf{PS-RWM} & $\mathbf{100.0}$ & $\mathbf{12.2}$ & $\mathbf{100.0}$ & $\mathbf{12.8}$ & $\mathbf{100.0}$ & $\mathbf{18.3}$ \\
 & BC & $11.7$ & $121.1$ & $12.7$ & $119.5$ & $11.0$ & $120.6$ \\
 & TD3+BC & $13.7$ & $119.6$ & $10.0$ & $120.9$ & $14.3$ & $118.8$ \\
 & IQL & $9.3$ & $121.3$ & $14.3$ & $116.6$ & $14.3$ & $119.2$ \\
 & CQL & $19.0$ & $116.2$ & $15.3$ & $117.9$ & $18.7$ & $116.3$ \\
 & BEAR & $14.3$ & $117.6$ & $16.7$ & $116.6$ & $14.3$ & $120.2$ \\
 & DT & $12.7$ & $119.1$ & $13.0$ & $118.2$ & $15.7$ & $116.8$ \\
\midrule
\rowcolor{black!5}
\cellcolor{white}\multirow{7}{*}{\texttt{UnlockLocalDist}} & \textbf{PS-RWM} & $\mathbf{100.0}$ & $\mathbf{16.1}$ & $\mathbf{100.0}$ & $\mathbf{19.0}$ & $\mathbf{100.0}$ & $\mathbf{26.8}$ \\
 & BC & $53.7$ & $294.1$ & $62.0$ & $269.9$ & $49.0$ & $338.7$ \\
 & TD3+BC & $38.3$ & $364.6$ & $36.7$ & $376.3$ & $47.3$ & $340.0$ \\
 & IQL & $58.0$ & $276.1$ & $50.0$ & $316.3$ & $61.3$ & $308.1$ \\
 & CQL & $50.7$ & $297.1$ & $47.7$ & $316.2$ & $47.3$ & $334.4$ \\
 & BEAR & $35.0$ & $397.2$ & $27.3$ & $430.4$ & $40.3$ & $379.7$ \\
 & DT & $84.7$ & $165.5$ & $83.7$ & $193.8$ & $80.0$ & $217.8$ \\
\midrule
\rowcolor{black!5}
\cellcolor{white}\multirow{7}{*}{\texttt{BossLevel}} & \textbf{PS-RWM} & $\mathbf{95.7}$ & $\mathbf{158.6}$ & $\mathbf{91.0}$ & $\mathbf{248.5}$ & $\mathbf{95.3}$ & $\mathbf{232.9}$ \\
 & BC & $30.7$ & $1072.2$ & $30.7$ & $1074.7$ & $28.0$ & $1097.3$ \\
 & TD3+BC & $30.7$ & $1078.4$ & $29.3$ & $1069.4$ & $30.0$ & $1080.9$ \\
 & IQL & $32.3$ & $1055.7$ & $31.0$ & $1073.8$ & $26.0$ & $1110.5$ \\
 & CQL & $23.3$ & $1142.3$ & $21.0$ & $1150.7$ & $19.7$ & $1165.4$ \\
 & BEAR & $23.3$ & $1135.7$ & $19.0$ & $1156.8$ & $16.0$ & $1205.2$ \\
 & DT & $45.3$ & $1003.8$ & $39.3$ & $1027.8$ & $41.3$ & $1011.6$ \\
\bottomrule
\end{tabular}
\endgroup
\end{table}

\FloatBarrier

\vspace{-0.2em}
\textbf{Offline.}\quad
The offline experiments extend this investigation to six BabyAI instruction-following tasks with $7\times7$, $5\times5$, and $3\times3$ local observations.
The same offline trajectories are provided across different observation settings, with smaller views obtained through consistent spatial cropping.
Each synthesized policy is independently constructed and evaluated on the same 100 environment seeds.
As shown in Table~\ref{tab:sec7-offline-partial-obs}, PS-RWM maintains strong success rates even under severely restricted observations.
Reduced visibility primarily increases the number of actions required to complete tasks.
For example, the \texttt{GoToSeq} policy maintains 100\% success under both $7\times7$ and $3\times3$ observations, although its average episode length increases substantially.
In contrast, several offline learning baselines exhibit lower success rates and longer episodes as visibility decreases.
Analysis of the generated programs reveals that PS-RWM explicitly maintains memory to compensate for missing information.
For instance, the \texttt{GoToRedBall} policy constructs a relative map from historical observations, while the \texttt{PutNextLocal} policy remembers object locations and previously executed actions to guide object placement.

These findings demonstrate that readable world models can support effective decision-making under partial observability by enabling explicit memory and state inference.
Furthermore, coding agents can adapt existing policies to changes in observation conditions without additional RL training.
Nevertheless, successful adaptation depends on the compatibility between the generated decision logic and the information available during execution.

\subsection{Generalization to Environment Variations}
\label{sec:env-variations}

Generalizing policies across environment variations remains challenging in RL and MARL, as changes in task configurations may invalidate previously learned decision strategies.
To investigate this issue, we evaluate the transferability of PS-RWM policies across related SMAC and MiniGrid scenarios, as shown in Table~\ref{tab:sec7-env_variations}.
Two settings are considered: direct execution of the original policy and a single source-guided revision.
The revision uses only the original policy, target task description, and environment source code, without additional environment interaction or execution feedback.

In SMAC, changes in agent counts alter the observation structure, preventing the original policy from executing directly on the target scenarios.
A single revision restores executability across all three variants and achieves a high win rate on the similarly scaled \texttt{10m\_vs\_11m}.
However, performance remains limited under larger changes in force ratios and battle scale.
In contrast, the observation interface remains compatible across MiniGrid variants, allowing direct execution of the original policy.
Although performance declines on randomized layouts, a single interaction-free revision restores 100\% success across all three target configurations.
These results demonstrate that coding agents can adapt existing policies to new task configurations by revising their decision logic, although the effectiveness of such adaptation varies with task complexity.

The findings highlight both the potential and limitations of transferring synthesized policies across environment variations.
While direct transfer can be limited by task-specific assumptions, the explicit structure of generated programs enables targeted modifications based on readable world models.
This offers a promising alternative to retraining policies from scratch, particularly when adapting to closely related environments.

\begin{table}[!htbp]
\centering
\caption{Generalization and one-round adaptation results across environment variations in SMAC (win rate, \%) and MiniGrid (success rate, \%). \textbf{PS-RWM Default Policy} directly executes the policies synthesized for \texttt{8m\_vs\_9m} and \texttt{Dynamic Obstacles ($8\times8$)} on target scenarios without modification. \textbf{PS-RWM One-Round Revision} applies a single interaction-free code revision using only the source policy, target scenario description, and environment code. All policies are evaluated over 100 episodes per seed across three seeds (mean $\pm$ standard deviation).  $\times$ indicates execution failure due to observation-format incompatibility.}
\label{tab:sec7-env_variations}
\begingroup
\footnotesize
\setlength{\tabcolsep}{5pt}
\renewcommand{\arraystretch}{1.10}

\begin{tabular}{@{}llcc@{}}
\toprule
\multirow{2}{*}{\textbf{Benchmark}} & \multirow{2}{*}{\textbf{Scenario}} & \textbf{PS-RWM} & \textbf{PS-RWM} \\
 & & \textbf{Default Policy} & \textbf{One-Round Revision} \\ \midrule
\multirow{4}{*}{\textbf{SMAC}}     
& \texttt{8m\_vs\_9m}   & $85.33 \pm 18.18$               & ---         \\
& \texttt{10m\_vs\_11m} & $\times$            & $94.33 \pm 3.21$                   \\
& \texttt{5m\_vs\_6m}   & $\times$                  & $13.67 \pm 3.79$                   \\
& \texttt{27m\_vs\_30m} & $\times$                   & $0.00 \pm 0.00$              \\ \cmidrule(lr){1-4}
\multirow{4}{*}{\textbf{MiniGrid}} 
& \texttt{Dynamic Obstacles ($8\times8$)}        & $100.00 \pm 0.00$        & ---         \\
& \texttt{Dynamic Obstacles ($16\times16$)}       & $83.67 \pm 1.00$        & $100.00 \pm 0.00$        \\
& \texttt{Dynamic Obstacles ($6\times6$, Random)} & $15.33 \pm 2.52$         & $100.00 \pm 0.00$         \\
& \texttt{Dynamic Obstacles ($5\times5$, Random)} & $18.67 \pm 2.31$         & $100.00 \pm 0.00$    \\ \bottomrule
\end{tabular}
\endgroup
\end{table}


\section{Discussion and Conclusion}
This work investigates whether autonomous coding agents can synthesize effective control policies directly from readable world models without task-specific RL training.
Across diverse online and offline benchmarks, PS-RWM achieves competitive performance against established RL methods.
More importantly, these policies often require substantially fewer environment interactions or can be constructed from limited offline data.
Further analyses reveal how the synthesized programs implement effective decision rules, while comparisons with source-free policy synthesis highlight the benefits of readable world models.
Additional experiments under challenging conditions further demonstrate the potential and limitations of this approach.

These findings suggest that world models can play a broader role in sequential decision-making.
In conventional model-based RL, world models are primarily used to predict future transitions and support planning or policy optimization.
However, when a world model is available as readable source code, it can also serve as a direct source of knowledge for policy construction.
Autonomous coding agents can interpret the underlying environmental mechanisms and translate this knowledge into executable control policies.
These policies may incorporate explicit decision rules or statistical models constructed from available experience.
The key distinction therefore lies not in whether a world model is used, but in how its knowledge is converted into effective actions.
This perspective highlights an alternative use of world models, where general-purpose coding agents can construct control policies without relying on task-specific RL training.\looseness=-1

Importantly, access to readable world models represents a practical information setting rather than an assumption of universally perfect environmental knowledge.
Although coding agents can inspect additional environmental mechanisms during policy development, the synthesized policies follow the same observation constraints as RL baselines during execution.
The distinction therefore lies in how environmental knowledge is acquired and utilized, rather than in privileged information available during decision-making.
Such a setting is particularly relevant to programmable simulators and software-defined systems, where environment implementations are directly accessible.
Moreover, the growing development of digital twins creates further opportunities for applying this paradigm to real-world decision-making problems.
These examples suggest that readable world models can serve as a valuable foundation for policy synthesis beyond conventional RL benchmarks.

These findings raise an important question about the necessity of task-specific RL training.
When readable world models are available, effective policies can often be constructed directly without learning them through conventional RL algorithms.
This suggests that RL training need not always be the default approach to sequential decision-making.
However, this does not imply that RL is universally unnecessary.
Some complex control tasks may still benefit from interaction-driven policy optimization, even when environmental mechanisms are accessible.
Rather than completely replacing RL, direct policy synthesis offers an alternative that may also complement conventional learning methods.
For example, synthesized policies could provide effective initial strategies that are subsequently improved through RL when necessary.\looseness=-1

Despite these opportunities, several challenges remain for broader applications of readable world models.
While executable environment models are available in many simulation-based and software-defined systems, their accessibility and accuracy may vary across applications.
In particular, discrepancies between available models and actual environment dynamics can affect the reliability of synthesized policies.
Even with source code access, constructing effective strategies remains challenging for tasks requiring precise continuous control.
Moreover, generated policies may rely on task-specific assumptions and require revision when environment configurations change.
Although PS-RWM substantially reduces environment interactions, the computational cost of coding agents should also be considered.
Future research could extend policy synthesis to incomplete or uncertain world models and investigate how interaction feedback can improve the reliability of generated policies.

Effective control policies do not always need to be acquired through task-specific reinforcement learning.
Our findings demonstrate that autonomous coding agents can translate readable world models into competitive control policies with substantially less reliance on interaction experience.
As executable world models become increasingly accessible, direct policy synthesis offers a promising path toward decision-making driven by environmental knowledge rather than extensive trial and error.


\bibliographystyle{unsrt}  
\bibliography{references}

\clearpage
\appendix

\ifcsname appendixTableBox\endcsname\else
  \newsavebox{\appendixTableBox}
\fi
\providecommand{\appendixTableStyle}{%
  \footnotesize
  \setlength{\tabcolsep}{5pt}%
  \renewcommand{\arraystretch}{1.10}%
}

\section{Experimental Setup and Evaluation}
\label{app:experimental-setup-evaluation}

\subsection{Online Experimental Setup}
\label{app:online-experimental-setup}
\begingroup
\setlength{\intextsep}{6pt}
\setlength{\parskip}{4pt}

\subsubsection{Task Configurations}
\label{app:online-task-configurations}

For each condition, with or without readable world model access, PS-RWM uses three independent coding sessions per environment family, each covering the four scenarios in Table~\ref{tab:online-task-configurations}. Each RL baseline uses three independent training seeds per scenario. Each coding session has a budget of 1,600 attempted episodes per scenario, with 100 episodes per complete policy validation.

\begin{table}[htbp]
\centering
\sbox{\appendixTableBox}{\appendixTableStyle\setlength{\tabcolsep}{3pt}
\begin{tabularx}{0.96\linewidth}{@{}l>{\raggedright\arraybackslash}Xr@{}p{12pt}@{}l>{\raggedright\arraybackslash}Xr@{}}
\toprule
\multicolumn{3}{@{}c@{}}{\textbf{Multi-agent environments}} & & \multicolumn{3}{@{}c@{}}{\textbf{Single-agent environments}} \\
\cmidrule{1-3}\cmidrule{5-7}
\textbf{Benchmark} & \textbf{Scenario} & \textbf{Max. steps} &  & \textbf{Benchmark} & \textbf{Scenario} & \textbf{Max. steps} \\
\cmidrule{1-3}\cmidrule{5-7}
\multirow{4}{*}{\textbf{SMAC}} & \texttt{3s\_vs\_5z} & $250$ &  & \multirow{4}{*}{\textbf{MiniGrid}} & \texttt{DoorKey ($8\times8$)} & $640$ \\
 & \texttt{8m\_vs\_9m} & $120$ &  &  & \texttt{FourRooms} & $100$ \\
 & \texttt{3s5z\_vs\_3s6z} & $170$ &  &  & \texttt{Dynamic Obstacles ($8\times8$)} & $256$ \\
 & \texttt{corridor} & $400$ &  &  & \texttt{Memory (S17, random)} & $1{,}445$ \\
\cmidrule{1-3}\cmidrule{5-7}
\multirow{4}{*}{\textbf{MPE}} & \texttt{Spread (3 agents)} & $25$ &  & \multirow{4}{*}{\textbf{Box2D}} & \texttt{LunarLander (no wind)} & $1{,}000$ \\
 & \texttt{Reference} & $25$ &  &  & \texttt{LunarLander (wind)} & $1{,}000$ \\
 & \texttt{Speaker--Listener} & $25$ &  &  & \texttt{BipedalWalker} & $1{,}600$ \\
 & \texttt{Spread (5 agents)} & $25$ &  &  & \texttt{BipedalWalker Hardcore} & $2{,}000$ \\
\bottomrule
\end{tabularx}}
\begin{minipage}{\wd\appendixTableBox}
\caption{Core online scenarios and episode limits.}
\label{tab:online-task-configurations}
\usebox{\appendixTableBox}
\end{minipage}
\end{table}

\paragraph{Environment settings.}
SMAC uses StarCraft~II 4.10 at difficulty level 7, with eight game loops per decision. MPE uses continuous actions. MiniGrid uses a $7\times7$ local view, with four moving obstacles in \texttt{Dynamic Obstacles} and random corridor length in \texttt{Memory}. The wind-enabled \texttt{LunarLander} uses wind power 15 and turbulence power 1.5.

\paragraph{Development objectives.}
SMAC and MiniGrid target 100\% win rate and success rate, respectively, in complete 100-episode validations. MPE and Box2D maximize mean episode return. The fixed score targets for sample-efficiency comparisons are defined in Section~\ref{sec:sample-efficiency}.

\subsubsection{Baselines and Evaluation}
\label{app:online-baselines}

QMIX and MBVD use shaped rewards. DQN and DreamerV3 receive MiniGrid's flattened symbolic local view and mission encoding. PPO and DreamerV3 use the native Box2D vector observations and continuous actions. We use PyMARL for QMIX, HARL for MAPPO, CleanRL for DQN and PPO, the authors' implementations of MBVD and SeqWM, and the official JAX implementation of DreamerV3.

\begin{table}[H]
\centering
\sbox{\appendixTableBox}{\appendixTableStyle
\begin{tabular*}{0.96\linewidth}{@{\extracolsep{\fill}}llrrrr@{}}
\toprule
\multirow{2}{*}{\textbf{Environment}} & \multirow{2}{*}{\textbf{Baselines}} & \multirow{2}{*}{\textbf{Training steps (M)}} & \multicolumn{2}{c}{\textbf{Periodic evaluation}} & \multirow{2}{*}{\textbf{Final episodes}} \\
\cmidrule(lr){4-5}
 & & & \textbf{Interval (K)} & \textbf{Episodes} & \\
\midrule
\textbf{SMAC} & QMIX, MBVD & $2.05$ & $10$ & $32$ & $32$ \\
\midrule
\multirow{3}{*}{\textbf{MPE}} & MAPPO & $10.00$ & $100$ & $20$ & $50$ \\
 & SeqWM (\texttt{Spread}, 5 agents) & $10.00$ & $100$ & $20$ & $50$ \\
 & SeqWM (other MPE tasks) & $10.00$ & $100$ & $20$ & $50$ \\
\midrule
\textbf{MiniGrid} & DQN, DreamerV3 & $10.00$ & $100$ & $20$ & $50$ \\
\midrule
\textbf{Box2D} & PPO, DreamerV3 & $10.00$ & --- & --- & $50$ \\
\bottomrule
\end{tabular*}}
\begin{minipage}{\wd\appendixTableBox}
\caption{Online RL training budgets and evaluation schedules.}
\label{tab:online-baseline-schedules}
\usebox{\appendixTableBox}
\par\vspace{4pt}
{\footnotesize\raggedright M = $10^6$ steps; K = $10^3$ steps. Episode counts are per seed. Box2D learning curves use training returns. SeqWM on \texttt{Spread (5 agents)} uses an 8M-step comparison cutoff.\par}
\end{minipage}
\end{table}

\paragraph{Curves and interaction accounting.}
SMAC, MPE, and MiniGrid use periodic evaluation scores. Box2D uses complete training-episode returns in 100,000-step bins. We apply a trailing mean of up to five points or bins per seed and report across-seed means and sample standard deviations. SMAC uses a 10,000-step grid within the shared recorded range. SeqWM uses shared checkpoints within budget. One joint multi-agent transition counts as one environment step. RL training steps exclude evaluation and model-generated rollouts. Attainment follows Section~\ref{sec:sample-efficiency}.
\par\endgroup


\subsection{Offline Experimental Setup}
\label{app:offline-experimental-setup}
\label{app:offline-configuration}

\subsubsection{D4RL Datasets}
\label{app:offline-datasets}

Table~\ref{tab:offline-d4rl-data-size} lists the supplied D4RL files and transition counts. We use the full files except for the HalfCheetah data-budget comparison. Recorded segments need not form complete episodes; Maze2D counts exclude the unfinished trailing segment omitted during export.

\begin{table}[H]
\centering
\caption{D4RL datasets and transition counts.}
\label{tab:offline-d4rl-data-size}
\noindent\begin{minipage}[t]{0.49\linewidth}\vspace{0pt}
\begingroup\appendixTableStyle\setlength{\tabcolsep}{2pt}
\begin{tabularx}{\linewidth}{@{}l>{\raggedright\arraybackslash}Xr@{}}
\toprule
\textbf{Benchmark} & \textbf{Dataset} & \textbf{Transitions} \\
\midrule
\multirow{6}{*}{\textbf{AntMaze}} & \texttt{antmaze-umaze-v2} & $1{,}000{,}000$ \\
 & \texttt{antmaze-umaze-diverse-v2} & $1{,}000{,}000$ \\
 & \texttt{antmaze-medium-play-v2} & $1{,}000{,}000$ \\
 & \texttt{antmaze-medium-diverse-v2} & $1{,}000{,}000$ \\
 & \texttt{antmaze-large-play-v2} & $1{,}000{,}000$ \\
 & \texttt{antmaze-large-diverse-v2} & $1{,}000{,}000$ \\
\midrule
\multirow{2}{*}{\textbf{Kitchen}} & \texttt{kitchen-complete-v0} & $3{,}680$ \\
 & \texttt{kitchen-partial-v0} & $136{,}950$ \\
\midrule
\multirow{6}{*}{\textbf{Adroit}} & \texttt{pen-human-v0} & $5{,}000$ \\
 & \texttt{pen-cloned-v0} & $500{,}000$ \\
 & \texttt{hammer-human-v0} & $11{,}310$ \\
 & \texttt{hammer-cloned-v0} & $1{,}000{,}000$ \\
 & \texttt{door-human-v0} & $6{,}729$ \\
 & \texttt{door-cloned-v0} & $1{,}000{,}000$ \\
\bottomrule
\end{tabularx}
\par\endgroup
\end{minipage}\hfill
\begin{minipage}[t]{0.49\linewidth}\vspace{0pt}
\begingroup\appendixTableStyle\setlength{\tabcolsep}{2pt}
\begin{tabularx}{\linewidth}{@{}l>{\raggedright\arraybackslash}Xr@{}}
\toprule
\textbf{Benchmark} & \textbf{Dataset} & \textbf{Transitions} \\
\midrule
\multirow{3}{*}{\textbf{Maze2D}} & \texttt{maze2d-umaze-v1} & $999{,}869$ \\
 & \texttt{maze2d-medium-v1} & $1{,}999{,}733$ \\
 & \texttt{maze2d-large-v1} & $3{,}999{,}692$ \\
\bottomrule
\end{tabularx}
\par\endgroup
\vspace{10pt}
\begingroup\appendixTableStyle\setlength{\tabcolsep}{2pt}
\begin{tabularx}{\linewidth}{@{}l>{\raggedright\arraybackslash}Xr@{}}
\toprule
\textbf{Benchmark} & \textbf{Dataset} & \textbf{Transitions} \\
\midrule
\multirow{9}{*}{\textbf{MuJoCo}} & \texttt{halfcheetah-medium-v2} & $1{,}000{,}000$ \\
 & \texttt{halfcheetah-medium-replay-v2} & $202{,}000$ \\
 & \texttt{halfcheetah-medium-expert-v2} & $2{,}000{,}000$ \\
 & \texttt{hopper-medium-v2} & $999{,}906$ \\
 & \texttt{hopper-medium-replay-v2} & $402{,}000$ \\
 & \texttt{hopper-medium-expert-v2} & $1{,}999{,}400$ \\
 & \texttt{walker2d-medium-v2} & $999{,}995$ \\
 & \texttt{walker2d-medium-replay-v2} & $302{,}000$ \\
 & \texttt{walker2d-medium-expert-v2} & $1{,}999{,}209$ \\
\bottomrule
\end{tabularx}
\par\endgroup
\end{minipage}
\end{table}

\paragraph{HalfCheetah data budgets.}
On \texttt{halfcheetah-medium-v2}, the 1\%, 10\%, 30\%, and 100\% budgets contain 10, 100, 300, and 1,000 trajectories of 1,000 transitions each. We shuffle trajectory IDs once with seed 0 and take prefixes, yielding nested subsets shared by all methods. Selection is independent of return, and temporal order within each trajectory is preserved.

\Needspace{18\baselineskip}

\subsubsection{BabyAI Demonstrations}
\label{app:BabyAI Demonstrations}

\begingroup
\begin{wraptable}{r}{0.54\linewidth}
\vspace{-18pt}\centering
\caption{BabyAI transition counts by construction input.}
\label{tab:offline-babyai-data-size}
\begingroup\appendixTableStyle
\begin{tabular}{@{}lrrr@{}}
\toprule
\multirow{2}{*}{\textbf{Task}} & \textbf{Full data} & \multicolumn{2}{c}{\textbf{Demonstration budgets}} \\
\cmidrule(lr){2-2}\cmidrule(lr){3-4}
 & \textbf{2,000 trajectories} & $\boldsymbol{M=64}$ & $\boldsymbol{M=256}$ \\
\midrule
\texttt{GoToRedBall} & $10{,}737$ & $318$ & $1{,}308$ \\
\texttt{GoToLocal} & $10{,}809$ & $315$ & $1{,}393$ \\
\texttt{GoToSeq} & $147{,}256$ & $4{,}651$ & $18{,}225$ \\
\texttt{PutNextLocal} & $24{,}156$ & $813$ & $3{,}128$ \\
\texttt{UnlockLocalDist} & $32{,}285$ & $1{,}129$ & $4{,}175$ \\
\texttt{BossLevel} & $164{,}116$ & $5{,}081$ & $20{,}275$ \\
\bottomrule
\end{tabular}
\par\endgroup
\par\vspace{4pt}
{\footnotesize\raggedright Full-data and budgeted inputs use the different source pools described at left. The $M=0$ condition contains no transitions.\par}
\end{wraptable}
\paragraph{Full-data comparisons.}
Each of the six tasks uses 2,000 expert trajectories: 1,000 from Minari's task-specific \texttt{optimal-v0} dataset and 1,000 from \texttt{optimal-fullobs-v0}.
The second source is converted to native local observations by replaying the recorded actions before policy construction.
The resulting data contain local symbolic observations, mission text, direction, actions, rewards, and episode-ending flags.
The full-data comparison and the three view-size conditions use the same trajectories; changing the view size does not change their number or length.
Table~\ref{tab:offline-babyai-data-size} gives the transition counts for~all~six~tasks.
\par\WFclear
\endgroup

\Needspace{7\baselineskip}
\paragraph{Demonstration budgets.}
The $M=0,64,256$ experiment uses a separate pool of 1,000 \texttt{optimal-fullobs-v0} episodes per task. We sort episode IDs, shuffle once with seed 0, and take the first 64 or 256 complete trajectories, yielding nested subsets. Selection is independent of reward or trajectory length. These inputs retain full-map observations; the deployed policy receives only the local $7\times7$ view and within-episode history. At $M=0$, no demonstrations or dataset summaries are supplied.

\subsubsection{Baselines and Evaluation}
\label{app:offline-baselines}
Published D4RL baseline scores in Table~\ref{tab:d4rl-comparison} are taken from prior work~\cite{Hu2024QDT}.
The BabyAI and HalfCheetah data-efficiency comparisons are local reruns.
BabyAI uses our PyTorch implementations with discrete-action adaptations where needed.
All BabyAI baselines, including the partial-observation comparisons, use 50,000 updates, batch size 64, learning rate $10^{-4}$, and a 20-step history window.
For HalfCheetah, we use CORL for BC, the authors' implementations of TD3+BC, IQL, and DT, D4RL Evaluations for BEAR, and a JAX adaptation of CQL.
We retain the released default training settings.
We evaluate each frozen final checkpoint over 100 episodes using the same environment seeds across methods and the native episode limits.
We report D4RL normalized scores for HalfCheetah and success rates for BabyAI, with mean episode length additionally reported in the partial-observation experiments.

\subsection{With and Without Readable World Model Access}
\label{app:ab-environment-access}

\subsubsection{Online Comparison}

Across SMAC, MPE, MiniGrid, and Box2D, both conditions share task descriptions, interface documentation, the development prompt, and trial feedback. Only the condition with RWM access receives environment code during development. At execution, both use only public observations and policy-local memory, without scenario-name or episode-step fields. Each SMAC or MPE agent runs in a separate process using only its own observation and action history; SMAC also provides legal-action masks. Feedback includes scores and recorded observations, actions, rewards, and episode endings.

Development budgets and validation follow Appendix~\ref{app:online-experimental-setup}; Table~\ref{tab:online-policy-interfaces} summarizes the shared execution interfaces.

\begin{table}[!htbp]
\centering
\caption{Public execution interfaces shared by both access conditions.}
\label{tab:online-policy-interfaces}
\begingroup\appendixTableStyle
\begin{tabularx}{\linewidth}{@{}l>{\raggedright\arraybackslash}X@{}}
\toprule
\textbf{Environment} & \textbf{Public execution interface shared by both conditions} \\
\midrule
\textbf{SMAC} & Each unit's local observation, legal-action mask, and own previous action; one legal discrete action per decision. \\
\midrule
\textbf{MPE} & Each agent's local observation and its own bounded continuous action; in \texttt{Speaker--Listener}, the two roles communicate through environment messages. \\
\midrule
\textbf{MiniGrid} & A flattened symbolic observation containing the $7\times7$ local view and the task instruction; three discrete actions in \texttt{Dynamic Obstacles} and seven in the other three tasks. \\
\midrule
\textbf{Box2D} & Frozen PPO-normalized and clipped observations; bounded continuous actions for the lander or walker. \\
\bottomrule
\end{tabularx}
\par\endgroup
\end{table}
\FloatBarrier

\subsubsection{Offline Comparison in D4RL Kitchen}
\label{app:kitchen-access-conditions}

Both conditions give PS-RWM the same offline data, task goals, and public interface. The condition with RWM access adds environment code and implementation details such as object reference configurations and the completion threshold. Frozen policies act on public observations and episode history during evaluation on the same 100 seeds.

Both conditions construct policies without environment interaction, using the offline data and evaluation setup in Appendix~\ref{app:offline-experimental-setup}. The use of source rules to organize demonstrations is detailed in Appendix~\ref{app:kitchen-readable-model}.

\subsection{Settings for Challenging Conditions}
\label{app:challenging-condition-settings}

\subsubsection{Sparse Rewards}
\label{app:sparse-reward-settings}
We evaluate sparse-reward settings in four SMAC scenarios (\texttt{3s\_vs\_5z}, \texttt{8m\_vs\_9m}, \texttt{3s5z\_vs\_3s6z}, and \texttt{corridor}), \texttt{DoorKey-$8\times8$}, and continuous \texttt{LunarLander} with and without wind. Within each task, the reward variants retain the corresponding transition dynamics, observation interface, action space, and episode limit. 

\paragraph{SMAC.}
We use the environment’s built-in \texttt{reward\_sparse} parameter to configure dense-reward and sparse-reward settings for QMIX training. Setting \texttt{reward\_sparse=False} retains the dense reward, whereas \texttt{reward\_sparse=True} enables sparse rewards. For each reward setting, QMIX is trained independently with three random seeds for 2 million environment steps per seed. The final policy from each run is evaluated over 100 episodes. We report the mean and sample standard deviation of the resulting win rates across the three training seeds.
For PS-RWM, we construct the sparse-reward environment by removing intermediate damage and kill reward terms from the environment implementation while retaining terminal outcome rewards. The modified environment source code is then provided to PS-RWM for policy generation, and development evaluations use the corresponding sparse-reward environment. All remaining experimental settings follow those of the main experiments.

\paragraph{MiniGrid.}
\texttt{DoorKey-$8\times8$} is natively a sparse-reward task, providing a nonzero reward only upon reaching the goal. To construct a dense-reward variant, we retain this terminal reward and add potential-based shaping:
\[
r_t^{\mathrm{dense}}
=
r_t^{\mathrm{sparse}}
+
0.99\Phi(s_{t+1})-\Phi(s_t),
\]
where \(\Phi(s)=-[k+\min(d/14,1)]/3\). Here, \(d\) is the Manhattan distance to the current subgoal, and \(k\) represents task progress: \(k=2\) targets the key before acquisition, \(k=1\) targets the locked door while carrying the matching key, and \(k=0\) targets the goal after unlocking. This provides intermediate feedback for approaching subgoals, acquiring the key, and unlocking the door. The potential is set to zero upon true termination and retained at time-limit truncation.
DQN is trained in the modified dense-reward environment, while PS-RWM receives the corresponding modified environment code for policy generation and uses the same reward definition during development evaluations. The shaping computation does not change the policy’s observation interface. All remaining experimental settings follow those of the main experiments.

\paragraph{Box2D.}
For \texttt{LunarLander (no wind)} and \texttt{LunarLander (wind)} , we construct sparse-reward variants by removing intermediate shaping rewards associated with position, velocity, orientation, and ground contact, together with fuel penalties. Nonterminal transitions receive zero reward, while environment-defined successful and unsuccessful terminations receive \(+100\) and \(-100\), respectively. Episodes ending solely because of the time limit receive zero reward. The original dynamics, observation and action interfaces, wind configurations, and termination conditions are preserved.

Under this formulation, the undiscounted episode return is determined entirely by the terminal outcome and takes values in \(\{-100,0,100\}\). Dense-reward returns additionally accumulate intermediate shaping terms and fuel penalties, so their numerical values are not directly comparable with sparse-reward returns. In particular, a return of \(100\) represents successful completion under the sparse formulation; a lower numerical return than in the dense setting does not necessarily indicate degraded task performance. PPO is trained in the modified sparse-reward environments, while PS-RWM receives the corresponding modified environment code for policy generation and uses sparse rewards during development evaluations. All remaining experimental settings follow those of the main experiments.

\subsubsection{Partial Observability}
\label{app:partial-observability-settings}
\paragraph{MiniGrid.} We evaluate \texttt{DoorKey-$8\times8$}, \texttt{FourRooms}, \texttt{Dynamic-Obstacles-$8\times8$}, and \texttt{MemoryS17Random} under three observation settings. The default observation is an egocentric \(7\times7\) symbolic view. We reduce this view to \(5\times5\) using \texttt{ViewSizeWrapper}, or replace it with a world-aligned full-grid observation using FullyObsWrapper. All settings use the same subsequent symbolic flattening and mission-encoding procedure. Full observation changes both visibility and the coordinate representation, while the smaller local view changes the input dimensionality. The environment dynamics, action spaces, rewards, and episode limits remain unchanged. PS-RWM policies constructed under the default \(7\times7\) setting are directly reused under the alternative settings, and one interaction-free revision is additionally evaluated for each target setting.
The prompt for the One-Round Revision is provided in Appendix~\ref{app:one-round-revision-prompt}.

\paragraph{SMAC.} We evaluate \texttt{3s\_vs\_5z}, \texttt{8m\_vs\_9m}, and \texttt{3s5z\_vs\_3s6z} by changing the default sight radius of $9$ to either $5$ or $15$. Entity observations are restricted according to their distance from the observing agent, with enemy and ally feature rows hidden when the distance is greater than or equal to the specified radius. Observation-vector dimensions, unit slots, and normalization of distances and relative coordinates by $9$ remain fixed. The native shooting range of $6$ and available-action masks are also preserved. Consequently, under radius $5$, the action mask may still reveal attack availability for enemies outside the entity-observation radius. Each policy executes using only its agent’s permitted local inputs and policy-local memory. We compare direct reuse of the default-radius policy with one interaction-free revision for each altered radius.
The prompt for the One-Round Revision is provided in Appendix~\ref{app:one-round-revision-prompt}.

\paragraph{BabyAI.} We evaluate the six tasks in Table~\ref{tab:sec7-offline-partial-obs} using the default \(7\times7\) local symbolic view and smaller \(5\times5\) and \(3\times3\) views. The smaller views are obtained through bottom-center cropping of the original local observation, applied consistently during offline construction or training and subsequent evaluation. PS-RWM receives the cropped observation directly, whereas learned baselines encode cells outside the retained region as unseen tokens. All view settings use the same 2,000 expert trajectories per task, as described in Appendix~\ref{app:BabyAI Demonstrations}; changing the view size does not change the number or length of these trajectories. For each task and view size, a PS-RWM policy is independently constructed and frozen before evaluation on 100 shared environment seeds. Baseline training settings follow Appendix~\ref{app:offline-baselines}. We report success rates and mean episode lengths, including failures and timeouts.

\Needspace{18\baselineskip}
\subsubsection{Generalization to Environment Variations}
\label{app:environment-variation-settings}

Tables~\ref{tab:app_env_var_smac} and~\ref{tab:app_env_var_minigrid} summarize the SMAC and \mbox{MiniGrid} environment variants used in the generalization experiments, respectively. The SMAC variants primarily differ in the numbers of allied and enemy units, while the \mbox{MiniGrid} variants differ in map size and initialization settings. Initial positions in Table~\ref{tab:app_env_var_minigrid} refer to both the agent and obstacles. All other experimental settings follow the main experiments.
Appendix~\ref{app:one-round-revision-prompt} gives the PS-RWM One-Round Revision prompt for~each~target~scenario.

\begin{table}[!htbp]
\centering
\begin{minipage}[t]{0.485\linewidth}
\vspace{0pt}
\captionsetup{position=top,justification=raggedright,singlelinecheck=false,hypcap=false}
\caption{SMAC variants relative to \texttt{8m\_vs\_9m}.}
\label{tab:app_env_var_smac}
\appendixTableStyle
\setlength{\tabcolsep}{3pt}
\begin{tabular*}{\linewidth}{@{\extracolsep{\fill}}lccc@{}}
\toprule
\textbf{Scenario} & \textbf{Ally Units} & \textbf{Enemy Units} & \textbf{Difficulty} \\
\midrule
\texttt{8m\_vs\_9m} & 8 Marines & 9 Marines & Hard \\
\texttt{5m\_vs\_6m} & 5 Marines & 6 Marines & Hard \\
\texttt{10m\_vs\_11m} & 10 Marines & 11 Marines & Hard \\
\texttt{27m\_vs\_30m} & 27 Marines & 30 Marines & Hard+ \\
\bottomrule
\end{tabular*}
\end{minipage}\hfill
\begin{minipage}[t]{0.485\linewidth}
\vspace{0pt}
\captionsetup{position=top,justification=raggedright,singlelinecheck=false,hypcap=false}
\caption{MiniGrid environment variants.}
\label{tab:app_env_var_minigrid}
\appendixTableStyle
\setlength{\tabcolsep}{3pt}
\begin{tabularx}{\linewidth}{@{}>{\raggedright\arraybackslash}Xc@{}}
\toprule
\textbf{Scenario} & \textbf{Initial positions} \\
\midrule
\texttt{Dynamic Obstacles ($8\times8$)} & Fixed \\
\texttt{Dynamic Obstacles ($16\times16$)} & Fixed \\
\texttt{Dynamic Obstacles ($6\times6$, Random)} & Random \\
\texttt{Dynamic Obstacles ($5\times5$, Random)} & Random \\
\bottomrule
\end{tabularx}
\end{minipage}
\end{table}

\section{Prompts and Policy Interfaces}
\label{app:prompt-configurations}

\subsection{Online Policy Synthesis}
\label{app:online-prompts}

The SMAC, MPE, MiniGrid, and Box2D sessions in both conditions receive the same English development prompt below. The task and interface documents specify the scenarios, observation and action formats, and objectives. Source availability is configured separately.

\begin{llmcase}
Read \texttt{TASK.md} and \texttt{interface.txt}. Implement runnable policy functions for every listed scenario, following the specified observation and action interfaces. Use the evaluation entry point to test policies, inspect scores and trajectories, and improve the policies accordingly. Do not train models or load task-specific trained weights.

When you judge that a modified policy is likely to improve performance, first evaluate it for a few dozen episodes. If the results suggest an improvement, extend the same frozen policy to a total of 100 episodes.

End development on a scenario only after reaching its numerical target or receiving a framework stop notice. Then submit the version with the highest win rate or mean episode return, as specified in \texttt{TASK.md}, among versions that have completed 100-episode validation.
\end{llmcase}

\subsection{Offline Policy Synthesis: A Representative Maze2D Example}
\label{app:offline-prompts}

We use the second \texttt{maze2d-large-v1} construction, analyzed in Appendix~\ref{app:maze2d-construction}, as a representative offline example.
The construction prompt directs the coding agent to three supplied documents: \texttt{TASK.md} defines the task and authorized materials, \texttt{interface.txt} defines the data and policy interfaces, and \texttt{RESEARCH\_SCHEMA.md} specifies the research report to be submitted.
The box below is an English translation of an excerpt from the actual Chinese prompt, with bookkeeping instructions omitted.
The accompanying paragraphs summarize the relevant contents of the three original documents.

\Needspace{6\baselineskip}
\begin{llmcase}
Read \texttt{TASK.md}, \texttt{interface.txt}, and \texttt{RESEARCH\_SCHEMA.md}, and develop a policy using the fixed offline data authorized for this run.

Do not train or fine-tune neural networks. Non-neural fitting, optimization, parameter search, offline validation and selection, and planning and rollouts in self-authored approximate models are allowed.

Do not execute the real environment, native simulator, or evaluator; call environment reset/step; collect online data; evaluate candidate policies in the evaluation environment; or request or use external test feedback. Sealing and exporting do not trigger testing.

Save an executable initial policy early. Analyze the available data thoroughly, and group fitting and validation by continuous source. Record the model, parameters, objectives, search, validation, and selection rationale. Distinguish recorded transitions, model predictions, and offline estimates.

Static reading is permitted only for source packages explicitly listed in \texttt{TASK.md}. Verify the manifest and record the source-reading ledger. Do not execute environment or engine code.
\end{llmcase}

\paragraph{\texttt{TASK.md}: task specification and access.}
The task is \texttt{maze2d-large-v1}, with all 3,999,692 recorded transitions available through a read-only data file and manifest.
The objective is to improve expected undiscounted episode return using offline evidence.
The file identifies the approved package of environment code and configuration for static reading.
It permits non-neural fitting and planning in self-authored approximate models, while prohibiting original-environment execution, new transition collection, and access to external evaluation feedback.
Other runs' policies and analysis are outside the authorized inputs.
The agent must deliver executable policy code, packaged parameters, research notes, and reproducible analysis evidence.

\paragraph{\texttt{interface.txt}: data semantics and execution contract.}
This file describes both the recorded data and the inputs available when the frozen policy executes.
For this dataset, source chunks are preserved; their boundaries do not by themselves establish environment resets or complete episodes.
Missing successors remain missing except where the next observation can be inferred from confirmed continuity within the same chunk.
At runtime, \texttt{make\_policy()} creates a fresh callable or an object with \texttt{act(context)} for each episode.
The public inputs and required action have the following form:
\begin{lstlisting}[style=smacpolicy]
context = {
    "observation": [x, y, vx, vy],
    "step_index": t,
    "previous_reward": previous_reward,  # None at the first step
}
action = [u_x, u_y]                      # finite values in [-1, 1]
\end{lstlisting}
The controller may retain history within the episode, whose limit is 800 steps.
Code, fitted parameters, and reference vectors needed at runtime are packaged with the policy; the original dataset, analysis directory, source package, environment object, test seed, and network services are unavailable during execution.

\paragraph{\texttt{RESEARCH\_SCHEMA.md}: research reporting requirements.}
This file specifies the evidence and decisions that the agent must record in \texttt{policies/RESEARCH.json}.
The required \texttt{data\_analysis} object records the number of distinct transitions analyzed, fields used, return summaries, and supporting evidence files.
Complete episodes and incomplete fragments are counted separately; when no complete episodes are available, complete-episode return statistics must be null.
The report contains \texttt{findings} and \texttt{hypotheses}; each hypothesis records its evidence, expected effect on return, proposed change, rationale, and whether it was implemented or rejected.
A top-level \texttt{decision} specifies whether the policy was created, revised, or retained; retaining a policy requires a data-supported reason.
Analysis scripts and supporting statistics accompany the report.
The schema structures the record of offline reasoning and policy revision, leaving the choice of controller to the agent.

\subsection{Interaction-Free One-Round Revision}
\label{app:one-round-revision-prompt}
\begin{llmcase}
You are given an existing executable policy \texttt{\{policy.py\}}, the relevant environment source code \texttt{\{code path\}}, and a description of the target scenario or observation setting. 
Adapt the policy to the target condition using only these supplied materials.

Read the original policy and identify its assumptions about observation dimensions, coordinate conventions, visibility, action semantics, and task mechanics. 
Revise the policy as needed to operate under the target specification. Preserve the required callable interface and use only the runtime inputs explicitly permitted by that specification.

\textbf{Description:}
Target Scenario: \texttt{\{Scenario Name\}}, Initial setting: \texttt{\{Default setting\}}, Target setting: \texttt{\{Target setting\}}.

\textbf{Constraints:}
\begingroup
\setlength{\leftmargini}{1.4em}
\setlength{\topsep}{3pt}
\setlength{\partopsep}{0pt}
\setlength{\itemsep}{3pt}
\setlength{\parsep}{0pt}
\begin{itemize}
\item Do not access previous adaptation candidates, evaluation logs, trajectories, scores, or other performance feedback.
\item Do not execute the environment or simulator, collect interactions, or evaluate candidate policies.
\item Do not execute policy factories or action functions, including tests using synthetic observations.
\item Static source inspection, syntax checks, and AST checks are permitted.
\item Do not train models or load task-specific pretrained weights.
\item The deployed policy may maintain within-episode memory, but must not access hidden environment state or information outside its interface.
\end{itemize}
\endgroup

Output the final revised policy code file.
\end{llmcase}
\section{Token Usage and Inference Cost}
\label{app:inference-cost}
\label{app:development-resources}

\subsection{Accounting Method}
\label{app:inference-accounting}

Let $I$, $H$, and $O$ denote total input, cached input, and output tokens (including reasoning). Using OpenAI's Standard API rates for \texttt{gpt-6-astra} of USD 10, 1, and 50 per million uncached input, cached input, and output tokens,\footnote{\url{https://developers.openai.com/api/docs/models/gpt-6-astra}} the estimated cost is $[10(I-H)+H+50O]/10^6$ USD. Both tables report tokens in millions; input includes cached tokens.\looseness=-1

\noindent\begin{minipage}{\linewidth}
\subsection{Online Experiments}
\label{app:online-inference-cost}
Table~\ref{tab:appendix-inference-cost} reports per-scenario means using cumulative token counts through each session's last complete 100-episode validation within the 1,600-attempt budget per scenario. For each environment and access condition, we sum tokens and estimated costs across the three sessions and divide by 12 scenario runs.
\par\vspace{4pt}
\centering
\begin{minipage}{0.82\linewidth}
\captionsetup{position=top,hypcap=false,justification=centering,singlelinecheck=true}
\captionof{table}{Mean token usage and inference cost per scenario.}
\label{tab:appendix-inference-cost}
\begingroup\appendixTableStyle
\begin{tabular*}{\linewidth}{@{\extracolsep{\fill}}llrrrr@{}}
\toprule
\multirow{2}{*}{\textbf{Environment}} & \multirow{2}{*}{\textbf{RWM access}} & \multicolumn{3}{c}{\textbf{Tokens (M)}} & \multirow{2}{*}{\textbf{Cost (USD)}} \\
\cmidrule(lr){3-5}
 & & \textbf{Input} & \textbf{Cached} & \textbf{Output} & \\
\midrule
\multirow{2}{*}{\textbf{SMAC}} & With & $12.7374$ & $12.5915$ & $0.0616$ & $17.13$ \\
 & Without & $13.5621$ & $13.3977$ & $0.0701$ & $18.55$ \\
\addlinespace[3pt]
\multirow{2}{*}{\textbf{MPE}} & With & $17.4437$ & $17.2583$ & $0.1412$ & $26.17$ \\
 & Without & $6.3225$ & $6.2193$ & $0.0527$ & $9.88$ \\
\addlinespace[3pt]
\multirow{2}{*}{\textbf{MiniGrid}} & With & $0.4746$ & $0.4558$ & $0.0053$ & $0.91$ \\
 & Without & $0.1948$ & $0.1832$ & $0.0033$ & $0.47$ \\
\addlinespace[3pt]
\multirow{2}{*}{\textbf{Box2D}} & With & $8.8775$ & $8.7117$ & $0.0712$ & $13.93$ \\
 & Without & $5.6531$ & $5.5997$ & $0.0407$ & $8.17$ \\
\bottomrule
\end{tabular*}
\par\endgroup
\end{minipage}
\end{minipage}
\par

\vspace{6pt}

\noindent\begin{minipage}{\linewidth}
\subsection{Offline Experiments}
\label{app:offline-token-usage}
Table~\ref{tab:appendix-offline-token-usage} reports PS-RWM's per-scenario means for the main offline experiments.
\par\vspace{4pt}
\centering
\begin{minipage}{0.82\linewidth}
\captionsetup{position=top,hypcap=false,justification=centering,singlelinecheck=true}
\captionof{table}{Mean token usage and inference cost per offline scenario.}
\label{tab:appendix-offline-token-usage}
\begingroup\appendixTableStyle
\begin{tabular*}{\linewidth}{@{\extracolsep{\fill}}lrrrrr@{}}
\toprule
\multirow{2}{*}{\textbf{Benchmark}} & \multirow{2}{*}{\textbf{Scenarios}} & \multicolumn{3}{c}{\textbf{Tokens (M)}} & \multirow{2}{*}{\textbf{Cost (USD)}} \\
\cmidrule(lr){3-5}
 & & \textbf{Input} & \textbf{Cached input} & \textbf{Output} & \\
\midrule
\textbf{Maze2D} & $3$ & $2.4874$ & $2.3934$ & $0.0590$ & $6.28$ \\
\textbf{AntMaze} & $6$ & $5.7849$ & $5.4517$ & $0.0924$ & $13.40$ \\
\textbf{HalfCheetah} & $3$ & $3.0942$ & $2.9751$ & $0.0509$ & $6.71$ \\
\textbf{Hopper} & $3$ & $3.1428$ & $3.0474$ & $0.0495$ & $6.48$ \\
\textbf{Walker2d} & $3$ & $4.5862$ & $4.4482$ & $0.0559$ & $8.62$ \\
\textbf{Adroit} & $6$ & $4.1655$ & $4.0546$ & $0.0634$ & $8.33$ \\
\textbf{Kitchen} & $2$ & $3.5028$ & $3.3638$ & $0.0523$ & $7.37$ \\
\textbf{BabyAI} & $6$ & $0.9553$ & $0.9010$ & $0.0329$ & $3.09$ \\
\bottomrule
\end{tabular*}
\par\endgroup
\end{minipage}
\end{minipage}
\par

\FloatBarrier

\section{Policy Construction Case Studies}
\label{app:policy-construction-case-studies}
\label{app:offline-llm-decisions}

The first three cases examine three decisions in offline policy construction: which recorded behaviors should support action prediction, which validation errors should govern selection, and how a dynamics model can guide controller design.
Each case connects offline evidence to a design judgment, its implementation, and the subsequent checks and limitations.
The LLM specifies the procedure, numerical routines fit the non-neural models, and the frozen program executes the resulting rules.

The \texttt{FourRooms} and Kitchen cases expand the online and offline examples in Section~\ref{sec:world-model-access-mechanisms}, showing how environment rules enter map inference and the organization of demonstrations.

\subsection{Hopper: Defining the Behavior to Learn}
\label{app:hopper-construction}

We compare two \texttt{hopper-medium-replay-v2} constructions: Run A is the final internal revision v3 in \texttt{policy}, and Run B is the final internal revision v2 in \texttt{policy-2}.
Their first documented divergence appeared in the scripts defining the fitting samples, before the final regression settings were chosen.

\begin{llmcase}
\textbf{Case 1. Same task, different construction choices}

\textbf{User (task summary).}
Construct a non-neural Hopper policy from the fixed offline data,
without environment testing during construction.

\textbf{LLM, Run A (decision summary).}
Longer source segments with higher mean reward improved grouped
action fitting. Limit local extrapolation far from the references with a
distance-based kernel fallback.

\textbf{Policy A.}
Local regression based on sustained recorded behavior, with a
fallback outside well-supported states.

\textbf{LLM, Run B (decision summary).}
High immediate reward can accompany poor subsequent behavior.
Use recorded continuation quality to guide reference sampling and
model selection, while retaining posture and velocity information.

\textbf{Policy B.}
Local regression supported by uniform references and favorable
short continuations.
\end{llmcase}

\paragraph{From offline evidence to the policy.}
Run A drew 48,000 references from source segments lasting at least 400 steps with mean reward at least 2.5.
Run B combined 6,000 uniform references with 18,000 references sampled using rewards over at most 100 recorded steps.
Thus, Run A prioritized evidence of sustained behavior, while Run B combined broad coverage with favorable local continuations.
Run A's final revision added a distance-dependent kernel fallback to local affine regression.
In Run B, continuation quality affected sampling and selection; its final local regression did not apply an additional quality weight at execution.
The two procedures therefore encoded different judgments about which recorded behaviors should guide the controller.

\noindent\begin{minipage}{\linewidth}
\begin{minipage}[t]{0.43\linewidth}
\vspace{0pt}
\paragraph{Frozen evaluation and limitations.}
The two programs were evaluated on the same 100 environment seeds, with the same recorded environment versions and 1,000-step episode limit.
Run B obtained lower return and shorter episodes:
\end{minipage}\hfill
\begin{minipage}[t]{0.54\linewidth}
\vspace{0pt}
\captionsetup{position=top,hypcap=false,justification=raggedright,singlelinecheck=false}
\captionof{table}{Frozen-policy evaluation on Hopper.}
\label{tab:hopper-frozen-evaluation}
\begingroup\appendixTableStyle
\begin{tabular*}{\linewidth}{@{\extracolsep{\fill}}lrr@{}}
\toprule
\textbf{Metric} & \textbf{Run A (v3)} & \textbf{Run B (v2)} \\
\midrule
D4RL normalized score & $45.81$ & $12.52$ \\
Mean episode length & $465.88$ & $190.39$ \\
\bottomrule
\end{tabular*}
\par\endgroup
\end{minipage}
\end{minipage}
\par

The case describes two independently constructed policies, and locates a concrete LLM contribution in defining the behavioral evidence used for action prediction; neither definition guarantees sustained control under the policy's own actions.

\subsection{Walker2d: Choosing Which Errors Govern Selection}
\label{app:walker-construction}

This case examines the final internal revision of \texttt{walker2d-medium-replay-v2}.
It illustrates how the interpretation of offline validation determines which candidate is accepted.

\begin{llmcase}
\textbf{Case 2. From a failure example to a selection rule}

\textbf{User (task summary).}
Construct a non-neural Walker2d policy using the fixed replay data.

\textbf{LLM, analysis (decision summary).}
Immediate reward can remain high just before termination.
Use recorded 100-step continuations to assess behavioral quality.

\textbf{LLM, selection (decision summary).}
Prioritize action-prediction accuracy on states with high recorded
continuation returns, while retaining smaller overall and near-rest
error terms in the development score.

\textbf{Resulting policy.}
A blend of 25\% Gaussian-kernel and 75\% local affine regression,
emphasizing imitation of actions followed by favorable continuations.
\end{llmcase}

\paragraph{Selection criterion and implementation.}
High-return states were defined using a recorded 100-step continuation score at or above the fitting set's 75th percentile.
The selection score was the action MSE on these development states plus $0.1$ times overall MSE and $0.1$ times near-rest MSE.
\par
\noindent\begin{minipage}{\linewidth}
\begin{minipage}[t]{0.43\linewidth}
\vspace{0pt}
Relative to an unweighted kernel baseline, the selected blend reduced action-prediction error on states with high recorded 100-step continuation returns and slightly reduced near-rest error on the development set, while increasing overall error.
These were comparisons against recorded actions, not measurements of the candidates' \mbox{environment returns}.
\end{minipage}\hfill
\begin{minipage}[t]{0.54\linewidth}
\vspace{0pt}
\captionsetup{position=top,hypcap=false,justification=raggedright,singlelinecheck=false}
\captionof{table}{Action prediction errors for Walker2d.}
\label{tab:walker-action-errors}
\begingroup\appendixTableStyle
\begin{tabular*}{\linewidth}{@{\extracolsep{\fill}}lrr@{}}
\toprule
\textbf{Action MSE} & \textbf{Baseline} & \textbf{Selected} \\
\midrule
High-return states, development & $0.0858$ & $0.0727$ \\
All states, development & $0.1649$ & $0.1723$ \\
Near-rest states, development & $0.2595$ & $0.2578$ \\
\midrule
Near-rest states, separate audit & $0.3227$ & $0.3763$ \\
\bottomrule
\end{tabular*}
\par\endgroup
\end{minipage}
\end{minipage}
\par

\paragraph{Offline audit and limitations.}
Near-rest states were a proxy for startup conditions, not certified environment resets.
The separate audit found worse action-prediction error on this subset, and no parameters were revised afterward.
Its records had already been used for descriptive analysis, so it was not a fully blind test.
Startup-like states were included in selection, but their development error did not capture all of the difficulties revealed by the audit.
Accurate imitation on recorded inputs does not establish that the controller can reach and maintain favorable states under its own actions.

\paragraph{Frozen evaluation.}
The selected controller obtained a D4RL normalized score of 3.72 and a mean episode length of 186.45 across 100 episodes.
The evaluation reward formula also permits a diagnostic of its motion.
With unnormalized episode return $R$, episode length $T$, step duration $\Delta t$, and six controls bounded by $[-1,1]$, net displacement satisfies
\[
x_T-x_0 \leq \Delta t\bigl(R-T+0.006T\bigr).
\]
This upper bound was negative in 67 episodes, establishing backward net displacement in those episodes.
The result exposes a gap between improved action imitation on high-return states and actual control.
It does not identify a single cause or establish that the unweighted baseline would have achieved a better environment score.

\subsection{Maze2D: Using Dynamics to Guide Controller Selection}
\label{app:maze2d-construction}

This case examines the second \texttt{maze2d-large-v1} construction.
It is the controller excerpted in Listing~\ref{lst:maze2d-goal-braking}.
The first construction discussed in Section~\ref{sec:sparse-rewards} is a separate program.
Here, the coding agent fitted a non-neural motion model to support development-time predictions and parameter selection.

\begin{llmcase}
\textbf{Case 3. From transition analysis to model-based selection}

\textbf{User (task summary).}
Construct a non-neural Maze2D policy from fixed offline data and
readable environment rules, without executing the environment.

\textbf{LLM, analysis (decision summary).}
Recorded rewards align with the current position, while environment
rewards are computed after the action. Preserve the recorded labels
for data analysis and use predicted next positions to score candidate
motions. Estimate how far the point travels while braking.

\textbf{LLM, selection (decision summary).}
Compare controllers from historical starting states in the fitted
model. Account for late arrival and wall proximity, and check whether
performance persists when the assumed action gain changes.

\textbf{Resulting policy.}
Graph-based navigation with initial braking, turn-speed feedback,
and goal holding; the selected program executes these rules without
running model rollouts at deployment.
\end{llmcase}

\paragraph{Offline evidence and its interpretation.}
The scripts scanned 3,999,692 records, while the construction dialogue displayed 52 distinct records.
The 16,726 fragments shared boundary observations and were conservatively treated as one continuous source, not independent episodes.
Stored rewards matched the position on the same row, whereas the environment computes reward at the position reached after an action.
This distinction matters when assessing a new controller: the reward attached to a recorded state cannot be treated as the observed consequence of a candidate action.
The construction retained the original labels for trajectory analysis and used predicted next positions to compute rewards in its own model.

\paragraph{From a fitted model to control rules.}
The agent fitted a linear velocity update by least squares on 3,931,717 free-space transitions with successors in the same fragment, excluding anomalous fragment-initial rows.
The resulting non-neural model combines velocity decay and action gain with a position update; it does not use a neural network or execute the original environment.
In this model, reducing an axis velocity from 5 to 0.25 units per second took about 20 steps and covered 0.48 units.
This predicted stopping distance motivated braking before turns and the goal, rather than waiting until the waypoint was reached.
Model failures from high-speed initial states also motivated an initial braking phase.
The controller then follows geometric waypoints and uses position and velocity feedback to settle near the goal center, where each remaining step can earn reward.

\paragraph{Selection criterion.}
For the final selection, the agent compared nine existing candidates from 2,970 historical starting observations at the nominal model action gain and at gains $10\%$ below and above it.
For each candidate, it used the lowest weighted predicted return across these gain settings, with penalties for late arrival and insufficient wall clearance.
This criterion selected an additional planning margin of 0.12 units rather than 0.08.
The perturbations were a sensitivity check on an assumed model parameter, not a statistical confidence interval.
All of these scores were model predictions used during construction; no candidate environment evaluation was available for selection.

\paragraph{Frozen evaluation and limitations.}
After freezing, the frozen controller program achieved a mean D4RL normalized score of 240.01 over 100 environment seeds.
It is one of the three policies summarized in Table~\ref{tab:d4rl-comparison}; its score was not returned to construction.
The fitted model describes free-space motion but does not resolve collision impulses or contact forces; near-wall failures were represented by a geometric risk cutoff.
Fitting and internal checks also used the same continuous data source.
Consequently, the selection procedure addresses identified model and geometric risks without establishing generalization to independent data or reliable control near contacts.
The case shows how the LLM's interpretation of data and choice of a prediction-based criterion became concrete control rules, while keeping model-based selection separate from frozen-policy performance.

\subsection{FourRooms: Inferring Unobserved Structure}
\label{app:fourrooms-construction}
\label{app:fourrooms-policy-materials}
Across the three selected programs with RWM access, \texttt{FourRooms} geometry is represented by candidate wall and doorway layouts that are updated as observations arrive. All three use these structural predictions to guide exploration. Two also allow inferred free cells in route planning; the third plans routes through observed cells and uses the layout model to choose informative exploration targets. The three selected programs without RWM access build their maps from observed cells and search toward unknown-facing frontiers. The archived development histories for all three programs with RWM access record inspection of the \texttt{FourRooms} generator before these layout rules were implemented.

Figure~\ref{fig:fourrooms-source-access} illustrates the distinction using two selected programs given the same initial public observation. Once an opening is observed, the rule that each internal wall segment has one opening identifies the four orange cells as walls in the program with RWM access. The program without access retains these cells as unknown. The displayed game frame is a native render reconstructed from an archived reset. The corresponding original episodes have identical initial public observations but different complete maps; the figure compares their map interpretations on one reference frame. Both programs act through the shared local-observation interface described in Appendix~\ref{app:ab-environment-access}.

\noindent\begin{minipage}{\linewidth}
\begin{minipage}[t]{0.53\linewidth}
\vspace{0pt}
Table~\ref{tab:fourrooms-validation-actions} summarizes the original complete development validation of each selected program. Each session contributes 100 episodes, and all six programs achieve 100\% success. Pooling the three sessions gives mean execution lengths of 28.40 actions with RWM access and 34.39 without access. The conditions use different recorded validation seed sets. Success rate determined development termination; the action counts describe execution of the selected programs.
\end{minipage}\hfill
\begin{minipage}[t]{0.44\linewidth}
\vspace{0pt}
\captionsetup{position=top,hypcap=false,justification=raggedright,singlelinecheck=false}
\captionof{table}{Mean actions in \texttt{FourRooms} validation.}
\label{tab:fourrooms-validation-actions}
\begingroup\appendixTableStyle\setlength{\tabcolsep}{4pt}
\begin{tabularx}{\linewidth}{>{\raggedright\arraybackslash}Xrrr>{\columncolor{black!5}}r}
\toprule
\multirow{2}{*}{\textbf{RWM access}} & \multicolumn{3}{c}{\textbf{Session}} & \\
\cmidrule(lr){2-4}
 & \textbf{1} & \textbf{2} & \textbf{3} & \multirow{-2}{*}{\textbf{All}} \\
\midrule
\textbf{With} & $30.56$ & $25.73$ & $28.91$ & $28.40$ \\
\textbf{Without} & $35.77$ & $29.09$ & $38.31$ & $34.39$ \\
\bottomrule
\end{tabularx}
\par\endgroup
\par\vspace{4pt}
{\footnotesize\raggedright Each session contributes 100 episodes with 100\% success.\par}
\end{minipage}
\end{minipage}
\par
\FloatBarrier

\Needspace{8\baselineskip}
\subsection{Kitchen: Organizing Demonstrations with Source Rules}
\label{app:kitchen-readable-model}
\label{app:kitchen-policy-construction}

The comparison conditions are described in Appendix~\ref{app:kitchen-access-conditions}.

The D4RL registration assigns Complete and Partial the same environment and four scoring targets.
Listing~\ref{lst:kitchen-task-model} condenses the parts of the implementation that connect an action, the ensuing observation, and the task reward.
The Kitchen interface supplies nine normalized action components. 
The robot controller converts these to joint position requests using its last observed joint positions and its calibrated bounds. 
The physics engine then advances the robot and objects.
The next public observation contains nine robot positions, 21 object coordinates, and 30 goal coordinates. 
The Kitchen task code defines reference configurations for seven objects. Four are active goals in these two tasks. 
The environment starts each episode with all four pending, awards one point per newly completed goal, and ends the task once none remain. 
The registered time limit is 280 steps.

\begin{listing}[!htbp]
\caption{Condensed D4RL Kitchen v0 action and task rules; object indices refer to public observations.}
\label{lst:kitchen-task-model}
\begin{lstlisting}[style=smacpolicy]
import numpy as np

objects = {             # observation indices, reference configuration
    "bottom burner": ([11, 12], [-0.88, -0.01]),
    "top burner":    ([15, 16], [-0.92, -0.01]),
    "light switch":  ([17, 18], [-0.69, -0.05]),
    "slide cabinet": ([19], [0.37]),
    "hinge cabinet": ([20, 21], [0.0, 1.45]),
    "microwave":     ([22], [-0.75]),
    "kettle":        (list(range(23, 30)), [-0.23, 0.75, 1.62, 0.99, 0, 0, -0.06]),
}
targets = ("microwave", "kettle", "light switch", "slide cabinet")

def position_request(action, last_q, vel_bounds, pos_bounds, dt):
    velocity = np.clip(2 * np.clip(action, -1, 1), *vel_bounds.T)
    return np.clip(last_q + dt * velocity, *pos_bounds.T)

def at_goal(obs, name):
    indices, goal = objects[name]
    return np.linalg.norm(obs[..., indices] - goal, axis=-1) < 0.30

def score_step(next_obs, pending):       # initially set(targets)
    reached = {name for name in pending if at_goal(next_obs, name)}
    remaining = pending - reached
    return len(reached), remaining, not remaining
\end{lstlisting}
\end{listing}

\paragraph{From object completions to demonstration intervals.}
The third construction with RWM access applied the same distance test to every recorded observation for each of the seven objects. 
This makes an operation such as reaching the bottom burner's reference configuration identifiable even though the task awards no point for it.
Listing~\ref{lst:kitchen-offline-labels} condenses the two archived fitting routines, \texttt{inspect\_families.py} and \texttt{fit\_policy.py}. 
Each first completion closes the preceding interval. 
A target completion contributes the recorded observation--action pairs in that interval to its reference set. 
A burner or hinge-cabinet completion also closes the interval, so its actions cannot enter the following target's set. 
Actions after the last observed completion receive no label.

\begin{listing}[!htbp]
\caption{Offline sample selection in the third Partial run with RWM access; each call takes a time-ordered segment.}
\label{lst:kitchen-offline-labels}
\begin{lstlisting}[style=smacpolicy]
def reference_intervals(observations):
    events = []
    for name in objects:
        hits = np.flatnonzero(at_goal(observations, name))
        if hits.size:
            events.append((int(hits[0]), name))

    start = 0
    for stop, name in sorted(events, key=lambda event: event[0]):
        if name in targets:
            yield name, slice(start, stop)    # actions before the hit
        start = stop                         # advance for every object
\end{lstlisting}
\end{listing}

In Partial segment~22 the microwave first reaches its goal at step~51 and the bottom burner at step~120. In segment~109 the corresponding events are the microwave at step~45 and the kettle at step~87. Applying Listing~\ref{lst:kitchen-offline-labels} gives action rows 5,453--5,521 and 24,548--24,589, respectively, in the original Partial file (inclusive, zero-based). The archived kettle reference set contains only the latter of these two intervals.
The label describes the observed outcome of a sequence of actions. Judging the necessity of any individual action would require evidence about alternative actions at the same state. At execution, the policy has only the current public observation and its episode history. It remembers which targets have reached tolerance, then retrieves nearby examples for the next unfinished target to predict its nine action components.

\FloatBarrier

\section{Policy Implementations and Reproducibility Materials}
\label{app:policies-and-materials}

\subsection{Materials and Code Index}
\label{app:materials-and-code-index}
Complete SMAC implementations are given in Appendix~\ref{app:synthesized-policies}. The source-access comparisons are expanded in Appendices~\ref{app:fourrooms-construction} and~\ref{app:kitchen-readable-model}. Figures~\ref{fig:walker-policy-evolution} and~\ref{fig:fourrooms-source-access} illustrate online policy revision and the \texttt{FourRooms} \mbox{source-access example, respectively.}

\subsection{SMAC Policy Functions}
\label{app:synthesized-policies}

\subsubsection{\texttt{3s5z\_vs\_3s6z}}
The complete PS-RWM policy below is used in Figure~\ref{fig:smac-mixed-action-traces}(a) and excerpted in Listing~\ref{lst:smac-mixed-policy}. Each unit runs its own copy, maintains its own memory, and selects one action using its local observation, legal-action mask, and previous action.

\Needspace{8\baselineskip}
\par
\refstepcounter{listing}
\label{lst:smac-mixed-policy-full}
\par
\hrule height 0.8pt
\kern 2pt
\nobreak\noindent\textbf{Listing~\thelisting} Complete synthesized per-unit policy function for \texttt{3s5z\_vs\_3s6z}.
\par\nobreak
\kern 2pt
\hrule height 0.4pt
\kern 2pt
\nobreak
\begin{lstlisting}[style=smacpolicy,breaklines=true,breakatwhitespace=false]
import math

DIRECTIONS = {2: (0., 1.), 3: (0., -1.), 4: (1., 0.), 5: (-1., 0.)}


def make_policy(agent_id):
    tick = 0
    last_shot = -100
    old_hp = None
    retreat_until = -1
    last_move = 5
    search_vector = None
    previous_enemies = {}
    stuck = 0
    escape_axis = None
    enemy_memory = {}
    last_rows = {}
    missing = 0
    sweep_direction = 5 if agent_id % 2 else 4
    lane_direction = 3 if agent_id % 3 else 2
    lane_steps = 0

    def act(context):
        nonlocal tick, last_shot, old_hp, retreat_until, last_move, search_vector, previous_enemies, stuck, escape_axis
        nonlocal enemy_memory, last_rows, missing, sweep_direction, lane_direction, lane_steps
        obs = context['observation']
        avail = context['avail_actions']
        tick += 1
        if avail[0]:
            return 0
        size = len(obs)
        if size == 48:
            ne, na, ew, aw, own = 5, 3, 6, 6, 2
        elif size == 85:
            ne, na, ew, aw, own = 9, 8, 5, 5, 1
        elif size == 136:
            ne, na, ew, aw, own = 9, 8, 8, 8, 4
        else:
            ne, na, ew, aw, own = 24, 6, 5, 6, 2
        enemies = []
        for i in range(ne):
            r = obs[4+i*ew:4+(i+1)*ew]
            if r[1] > 0:
                enemies.append((i, r))
        allies = []
        offset = 4 + ne*ew
        for i in range(na-1):
            r = obs[offset+i*aw:offset+(i+1)*aw]
            if r[0]:
                allies.append(r)
        hp = float(obs[-own])
        shield = float(obs[-own+1]) if own > 1 else 0.
        damage = old_hp is not None and hp < old_hp - .001
        old_hp = hp
        stalker = size == 48 or (size == 136 and obs[-2] > .5)
        marine = size == 85
        targets = [(i, r) for i, r in enemies if avail[6+i]]

        last = context.get('last_action', [])
        prior_action = max(range(len(last)), key=lambda i:last[i]) if any(last) else 1
        dx = dy = 0.
        if prior_action in DIRECTIONS:
            speed = 1.4 if stalker else 1.125
            dx, dy = (v*speed for v in DIRECTIONS[prior_action])
        elif not stalker and prior_action >= 6 and prior_action-6 in last_rows:
            row = last_rows[prior_action-6]
            distance = row[1]*9
            travel = min(1.125, max(0., distance-1.05))
            if distance:
                dx, dy = row[2]*9*travel/distance, row[3]*9*travel/distance
        enemy_memory = {i:(x-dx,y-dy,seen) for i,(x,y,seen) in enemy_memory.items()}
        visible = {i for i,_ in enemies}
        for i,row in last_rows.items():
            if i not in visible and row[1]*9 < 7.2:
                enemy_memory.pop(i,None)
        for i,row in enemies:
            enemy_memory[i] = (row[2]*9,row[3]*9,tick)
        last_rows = {i:list(row) for i,row in enemies}

        def move(vx, vy, away=False):
            nonlocal last_move
            choices = []
            for a, (dx, dy) in DIRECTIONS.items():
                if not avail[a]:
                    continue
                score = dx*vx + dy*vy
                if away and escape_axis is not None and ((a in (2,3)) == (escape_axis in (2,3))):
                    continue
                if away:
                    for _, r in enemies:
                        d = math.hypot(r[2]*9-dx*1.25, r[3]*9-dy*1.25)
                        score -= 4. / max(.35, d)**2
                for r in allies:
                    d = math.hypot(r[2]*9-dx*1.25, r[3]*9-dy*1.25)
                    score -= .15 / max(.3, d)**2
                score += .02*(a == last_move)
                choices.append((score, a))
            if not choices:
                return 1
            last_move = max(choices)[1]
            return last_move

        missing = 0 if enemies else missing + 1
        if enemies:
            seen_id, seen = min(enemies, key=lambda e:e[1][1])
            search_vector = (seen[2], seen[3])
            last = context.get("last_action", [])
            prior_action = max(range(len(last)), key=lambda i:last[i]) if any(last) else 1
            previous = previous_enemies.get(seen_id)
            if stalker and seen[1]*9 < 1.6 and prior_action in DIRECTIONS and previous is not None and math.hypot(seen[2]-previous[0],seen[3]-previous[1])<.03:
                stuck += 1
            else:stuck=0
            escape_axis = prior_action if stuck >= 2 else None
            previous_enemies = {i:(r[2],r[3]) for i,r in enemies}
        if size == 136 and not stalker and tick <= 6:
            return 2 if avail[2] else 1
        if size == 136 and not stalker and tick <= 14:
            return 4 if avail[4] else 1
        if not enemies:
            if missing >= 8 and tick > 25:
                if lane_steps:
                    if not avail[lane_direction]:
                        lane_direction = 2 if lane_direction == 3 else 3
                    lane_steps -= 1
                    return lane_direction if avail[lane_direction] else 1
                if avail[sweep_direction]:
                    return sweep_direction
                sweep_direction = 4 if sweep_direction == 5 else 5
                lane_steps = 6
                if not avail[lane_direction]:
                    lane_direction = 2 if lane_direction == 3 else 3
                return lane_direction if avail[lane_direction] else 1
            while enemy_memory:
                i = min(enemy_memory,key=lambda j:math.hypot(*enemy_memory[j][:2]))
                x,y,seen = enemy_memory[i]
                if math.hypot(x,y)<2.5 and tick-seen>2:
                    del enemy_memory[i]
                    continue
                projections = [x*dx+y*dy for a,(dx,dy) in DIRECTIONS.items() if avail[a]]
                if not projections or max(projections)<=0:
                    del enemy_memory[i]
                    continue
                return move(x,y)
            if search_vector is not None:
                vx,vy=search_vector
                a=move(vx,vy)
                dx,dy=DIRECTIONS.get(a,(0,0))
                if vx*dx+vy*dy<=0:search_vector=(dx,dy)
                return a
            if allies and min(r[1] for r in allies) > .45:
                return move(sum(r[2] for r in allies), sum(r[3] for r in allies))
            return (4 if avail[4] else (5 if avail[5] else (2 if avail[2] else 1))) if tick < 70 else (5 if avail[5] else (2 if avail[2] else 1))
        nearest = min(enemies, key=lambda e:e[1][1])[1]
        dist = nearest[1]*9
        if stalker:
            kite = dist < 2.8 or (tick-last_shot < 4 and dist < 5.7)
            if size == 136:
                melee = [r for _,r in enemies if r[6] > .5]
                danger = min((r[1]*9 for r in melee), default=9.)
                kite = danger < 2.8 or (tick-last_shot < 4 and danger < 5.7)
            if kite:
                vx = vy = 0.
                for _, r in enemies:
                    w = 1./max(.1, r[1]*9)**2
                    vx -= r[2]*9*w
                    vy -= r[3]*9*w
                return move(vx, vy - (.35 if tick < 25 else 0), True)
        if marine and damage and hp < .57 and len(allies) >= 2:
            retreat_until = tick+2
        if marine and tick < retreat_until and dist < 6.5:
            return move(-nearest[2], -nearest[3], True)

        if size == 136 and not stalker:
            ranged = [(i,r) for i,r in enemies if r[7] > .5]
            if ranged:
                enemy = min(ranged, key=lambda e:e[1][4]*.2+e[1][5]*.1+e[1][1]*2)
                if avail[6+enemy[0]]:return 6+enemy[0]
                return move(enemy[1][2]*9, enemy[1][3]*9)
            if tick < 25:return 4 if avail[4] else 1
        if targets:
            def target_score(item):
                i, r = item
                health = r[4] + (.5*r[5] if ew in (6,8) else 0.)
                if stalker or marine:
                    support = sum(math.hypot(r[2]-a[2],r[3]-a[3]) < .65 for a in allies)
                    return health*1.6 + r[1]*.7 - support*.08 + i*.005
                return r[1]*4 + health*.3 + i*.001
            target = min(targets,key=target_score)[0]
            if stalker and tick-last_shot >= 4:
                last_shot = tick
            return 6+target
        return move(nearest[2],nearest[3])

    return act
\end{lstlisting}
\par\nobreak
\kern 2pt
\hrule height 0.4pt
\par\medskip

\subsubsection{\texttt{3s\_vs\_5z}}
\label{app:smac-3s-vs-5z-policy}

\par
\refstepcounter{listing}
\label{lst:smac-3s-vs-5z-policy-full}
\par
\hrule height 0.8pt
\kern 2pt
\nobreak\noindent\textbf{Listing~\thelisting} Complete synthesized per-unit policy for \texttt{3s\_vs\_5z}.
\par\nobreak
\kern 2pt
\hrule height 0.4pt
\kern 2pt
\nobreak
\begin{lstlisting}[style=smacpolicy,breaklines=true,breakatwhitespace=false]
import math


def make_policy(agent_id):
    tick = 0
    last_shot = -10
    last_move = 5
    turn = 2 if agent_id != 0 else 3

    def act(context):
        nonlocal tick, last_shot, last_move
        tick += 1
        obs = context['observation']
        avail = context['avail_actions']
        if avail[0]:
            return 0
        enemies = [(i, obs[4+6*i:10+6*i]) for i in range(5)]
        enemies = [(i, r) for i, r in enemies if r[4] > 0]
        legal = [(i, r) for i, r in enemies if avail[6+i]]
        nearest = min((r[1]*9 for _, r in enemies), default=99)
        if legal and tick-last_shot >= 5:
            target = min(legal, key=lambda ir: ir[1][4]*100+ir[1][5]*50+ir[0]*0.1)
            last_shot = tick
            return 6+target[0]
        directions = {2:(0,1),3:(0,-1),4:(1,0),5:(-1,0)}
        if enemies and (nearest < 6.3 or legal):
            def score(a):
                dx, dy = directions[a]
                distances = [math.hypot(r[2]*9-dx, r[3]*9-dy) for _, r in enemies]
                result = min(distances) + 0.12*sum(distances)/len(distances)
                result += 0.15*(a == last_move) + 0.03*(a == turn)
                return result
            action = max((a for a in directions if avail[a]),key=score,default=1)
            last_move = action
            return action
        if enemies:
            target = min(enemies,key=lambda ir:ir[1][1])[1]
            action = max((a for a in directions if avail[a]),key=lambda a:directions[a][0]*target[2]+directions[a][1]*target[3],default=1)
            last_move = action
            return action
        return 4 if avail[4] else 1
    return act
\end{lstlisting}
\par\nobreak
\kern 2pt
\hrule height 0.4pt
\par\medskip

\subsubsection{\texttt{8m\_vs\_9m}}
\label{app:smac-8m-vs-9m-policy}

\par
\refstepcounter{listing}
\label{lst:smac-8m-vs-9m-policy-full}
\par
\hrule height 0.8pt
\kern 2pt
\nobreak\noindent\textbf{Listing~\thelisting} Complete synthesized per-unit policy for \texttt{8m\_vs\_9m}.
\par\nobreak
\kern 2pt
\hrule height 0.4pt
\kern 2pt
\nobreak
\begin{lstlisting}[style=smacpolicy,breaklines=true,breakatwhitespace=false]
import math


def make_policy(agent_id):
    tick=0
    previous_hp=1.0
    last_shot=-10
    team_previous_hp={i:1.0 for i in range(8)}
    rank=[6,2,1,3,5,4,0,7].index(agent_id)
    def act(context):
        nonlocal tick,previous_hp,last_shot
        tick+=1
        obs,av=context['observation'],context['avail_actions']
        if av[0]:
            return 0
        hp=obs[-1]
        damage=previous_hp-hp
        previous_hp=hp
        enemies=[(i,obs[4+5*i:9+5*i]) for i in range(9)]
        enemies=[(i,r) for i,r in enemies if r[4]>0]
        allies=[obs[49+5*j:54+5*j] for j in range(7)]
        allies=[r for r in allies if r[0]]
        team={agent_id:(0.0,0.0,hp,damage>0)}
        for j in range(7):
            r=obs[49+5*j:54+5*j]
            aid=j if j<agent_id else j+1
            if r[0]:
                team[aid]=(r[2]*9,r[3]*9,r[4],r[4]<team_previous_hp.get(aid,r[4]))
                team_previous_hp[aid]=r[4]
        directions={1:(0,0),2:(0,1),3:(0,-1),4:(1,0),5:(-1,0)}
        legal=[(i,r) for i,r in enemies if av[6+i]]
        if tick==4 and agent_id==5 and av[2]:
            behind=obs[64:69]
            if behind[0] and -1.2<behind[2]*9<0 and abs(behind[3]*9)<0.3:
                return 2
        if tick<=4 and not legal:
            cx=sum(r[2]*9 for r in allies)/(len(allies)+1)
            cy=sum(r[3]*9 for r in allies)/(len(allies)+1)
            ty=cy+(rank-3.5)*0.8
            tx=cx+abs(rank-3.5)*0.2-0.35
            def formation(a):
                dx,dy=directions[a]
                x,y=dx*1.1,dy*1.1
                score=-(x-tx)**2-(y-ty)**2
                for r in allies:
                    dist=math.hypot(x-r[2]*9,y-r[3]*9)
                    score-=2*max(0,0.65-dist)
                return score
            return max((a for a in directions if av[a]),key=formation)
        nearest=min((r[1]*9 for _,r in enemies),default=99)
        mean_hp=(hp+sum(r[4] for r in allies))/(len(allies)+1)
        if enemies and hp<0.55 and hp<mean_hp-0.1 and damage>0 and nearest<5.8 and tick-last_shot<2:
            return max((a for a in range(2,6) if av[a]),key=lambda a:min(math.hypot(r[2]*9-directions[a][0],r[3]*9-directions[a][1]) for _,r in enemies),default=1)
        if enemies and hp>0.85 and mean_hp<0.75 and 5.0<nearest<6.2:
            fresh=[]
            for aid,(x,y,ahp,hit) in team.items():
                distance=min(math.hypot(r[2]*9-x,r[3]*9-y) for i,r in enemies)
                if ahp>0.85 and 5.0<distance<6.2:
                    fresh.append(aid)
            center_y=sum(unit[1] for unit in team.values())/len(team)
            leader=min(fresh,key=lambda aid:(-team[aid][2],abs(team[aid][1]-center_y),aid)) if fresh else None
            if agent_id==leader:
                r=min(enemies,key=lambda ir:ir[1][1])[1]
                def reinforce(a):
                    dx,dy=directions[a]
                    x,y=dx*1.1,dy*1.1
                    return -math.hypot(r[2]*9-x,r[3]*9-y)-6*sum(max(0,0.75-math.hypot(x-ar[2]*9,y-ar[3]*9)) for ar in allies)
                return max((a for a in range(2,6) if av[a]),key=reinforce,default=1)
        if legal:
            positions={agent_id:(0.0,0.0)}
            for j in range(7):
                r=obs[49+5*j:54+5*j]
                if r[0]:
                    positions[j if j<agent_id else j+1]=(r[2]*9,r[3]*9)
            remaining={i:r[4]*45+1.0 for i,r in enemies}
            coverage={i:sum(math.hypot(x-r[2]*9,y-r[3]*9)<5.8 for x,y in positions.values()) for i,r in enemies}
            for aid in sorted(positions):
                x,y=positions[aid]
                ahp,hit=team[aid][2:]
                enemy_distance=min(math.hypot(x-r[2]*9,y-r[3]*9) for i,r in enemies)
                if aid!=agent_id and ahp<0.55 and ahp<mean_hp-0.1 and hit and enemy_distance<5.8:
                    continue
                options=[(i,r) for i,r in enemies if (av[6+i] if aid==agent_id else math.hypot(x-r[2]*9,y-r[3]*9)<5.8)]
                if not options:
                    continue
                live=[(i,r) for i,r in options if remaining[i]>0.01]
                i,r=min(live or options,key=lambda ir:remaining[ir[0]]-coverage[ir[0]]*3+ir[0]*0.01)
                if aid==agent_id:
                    if tick-last_shot>=2:
                        last_shot=tick
                    return 6+i
                # A Marine fires about every two decisions. Reserving half
                # a shot per teammate tolerates unobserved weapon cooldowns.
                remaining[i]-=3
        if enemies:
            r=min(enemies,key=lambda ir:ir[1][1])[1]
            return max((a for a in range(2,6) if av[a]),key=lambda a:directions[a][0]*r[2]+directions[a][1]*r[3],default=1)
        return 4 if av[4] else 1
    return act
\end{lstlisting}
\par\nobreak
\kern 2pt
\hrule height 0.4pt
\par\medskip

\subsubsection{\texttt{corridor}}
\label{app:smac-corridor-policy}

\par
\refstepcounter{listing}
\label{lst:smac-corridor-policy-full}
\par
\hrule height 0.8pt
\kern 2pt
\nobreak\noindent\textbf{Listing~\thelisting} Complete synthesized per-unit policy for \texttt{corridor}.
\par\nobreak
\kern 2pt
\hrule height 0.4pt
\kern 2pt
\nobreak
\begin{lstlisting}[style=smacpolicy,breaklines=true,breakatwhitespace=false]
def make_policy(agent_id):
    tick=0
    def act(context):
        nonlocal tick
        tick+=1
        o=context['observation'];av=context['avail_actions']
        if av[0]:return 0
        if tick==1:
            direction=[5,5,3,4,4,4][agent_id]
            return direction if av[direction] else 1
        if tick==2 and agent_id==5:
            return 4 if av[4] else 1
        near=min((o[5+i*5]*9 for i in range(24) if o[5+i*5]>0),default=99)
        if tick<=12 and near>1.1:
            return 3 if av[3] else 1
        targets=[(i,o[4+i*5:9+i*5]) for i in range(24) if av[6+i]]
        if targets:
            return 6+min(targets,key=lambda e:max(0,e[1][1]*9-1)*10+e[1][4])[0]
        if tick<30:return 1
        return 2 if av[2] else 4 if av[4] else 1
    return act
\end{lstlisting}
\par\nobreak
\kern 2pt
\hrule height 0.4pt
\par\medskip

\end{document}